%% file: main.tex
\documentclass{article} 
\usepackage[svgnames]{xcolor}

\usepackage{iclr2025_conference,times}
\usepackage[inline]{enumitem}
\input{commands/math_commands}

\usepackage{hyperref}
\usepackage{url}

\usepackage{tabularx}

\title{Ctrl-V: High Fidelity Video Generation with Bounding-Box Controlled Object Motion}

\author{Ge Ya Luo\\
Mila, Université de Montréal
\And
Zhi Hao Luo\\
Mila, Polytechnique Montreal
\AND
Anthony Gosselin\\
Mila, Polytechnique Montreal
\And
Alexia Jolicoeur-Martineau\\
Samsung - SAIT AI Lab, Montreal
\AND
Christopher Pal\\
Mila, Polytechnique Montreal\\
Canada CIFAR AI Chair
}

\begin{document}

\maketitle


\input{sections/abstract}

\begin{figure}[h!]
    \centering
\includegraphics[width=\textwidth]{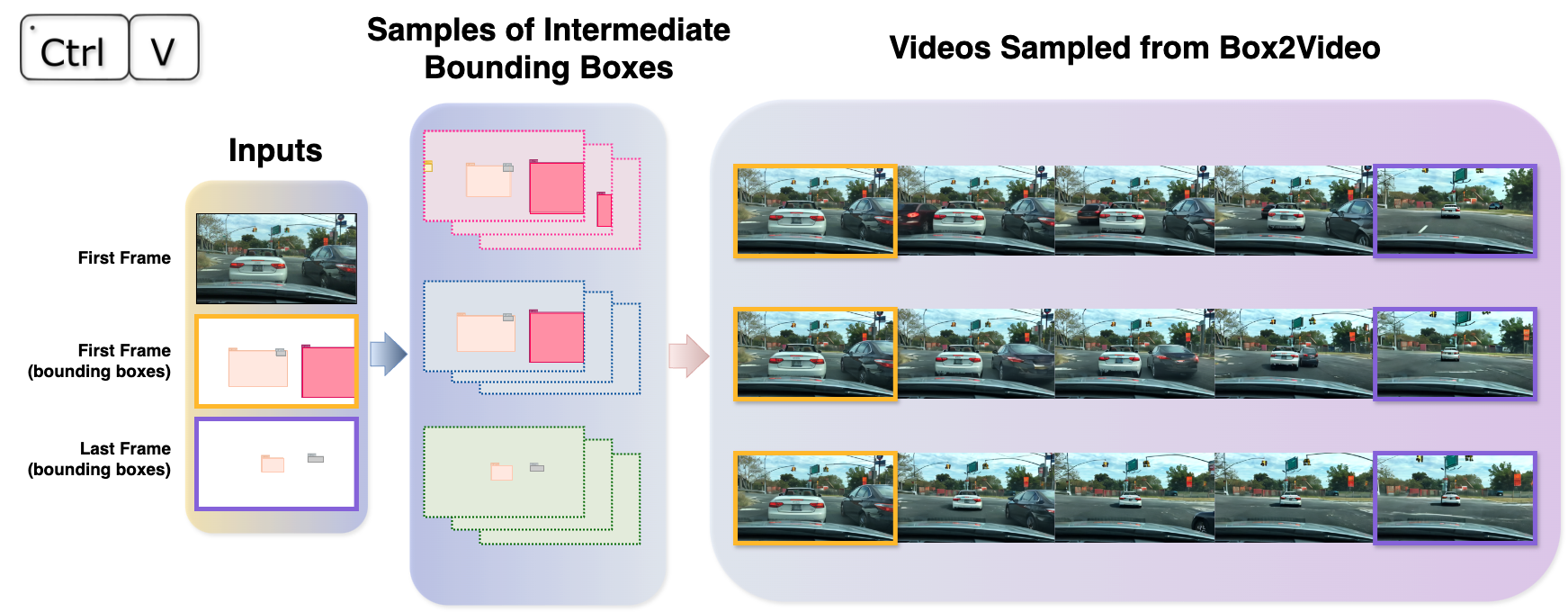}
    \caption{Overview of Ctrl-V's generation pipeline: \textbf{(Left) inputs: }Our inputs include an initial frame, its corresponding bounding box image and the final frame's bounding box image. \textbf{(Middle) generated bounding box trajectories:} We demonstrate three distinct possible trajectory sequences produced by our diffusion-based bounding box motion generation model -- \modelbbox. \textbf{(Right) generated video clips: }Our \modelvid~model conditions on the generated bounding box trajectory frames to produce the final video clips.}
    \label{fig:enter-label}
\end{figure}
\section{Introduction}

\label{sec:intro}
\input{sections/introduction}

\section{Related Work}
\label{sec:related}
\input{sections/related_work}

\section{Our Method: Ctrl-V}
\label{sec:method}
\input{sections/method}

\section{Experimental Analysis and Ablation Studies}
\label{sec:experiments}
\input{sections/experiments}

\section{Conclusions}\label{sec:conclusion}
\input{sections/conclusion}


\bibliography{references}  
\bibliographystyle{iclr2025_conference}

\appendix
\input{sections/appendix}

\end{document}

%% file: commands/math_commands.tex
\usepackage{amsmath,amsfonts,bm,amssymb}
\usepackage{graphicx}
\usepackage{subfig}
\usepackage{array,booktabs}
\newcolumntype{M}[1]{>{\centering\arraybackslash}m{#1}} 
\usepackage{subcaption} 

\newcommand{\modelbbox}{BBox Generator}
\newcommand{\modelvid}{Box2Video}

\definecolor{svdunet}{cmyk}{0, 0.7808, 0.4429, 0.1412}

\def\eqref#1{equation~\ref{#1}}

\def\1{\bm{1}}

\def\vb{{\bm{b}}}
\def\vc{{\bm{c}}}

\def\vf{{\bm{f}}}

\def\vz{{\bm{z}}}

\DeclareMathAlphabet{\mathsfit}{\encodingdefault}{\sfdefault}{m}{sl}
\SetMathAlphabet{\mathsfit}{bold}{\encodingdefault}{\sfdefault}{bx}{n}

\def\sD{{\mathbb{D}}}

\def\sU{{\mathbb{U}}}

\newcommand{\En}{\mathcal{E}}
\newcommand{\De}{\mathcal{D}}

\newcommand{\R}{\mathbb{R}}

\newcommand{\specialcell}[2][c]{%
\begin{tabular}[#1]{@{}c@{}}#2\end{tabular}}

\usepackage{arydshln}
\usepackage{array,multirow,graphicx}
\newcommand{\STAB}[1]{\begin{tabular}{@{}c@{}}#1\end{tabular}}

\makeatletter
\def\adl@drawiv#1#2#3{%
        \hskip.5\tabcolsep
        \xleaders#3{#2.5\@tempdimb #1{1}#2.5\@tempdimb}%
                #2\z@ plus1fil minus1fil\relax
        \hskip.5\tabcolsep}
\newcommand{\cdashlinelr}[1]{%
  \noalign{\vskip\aboverulesep
           \global\let\@dashdrawstore\adl@draw
           \global\let\adl@draw\adl@drawiv}
  \cdashline{#1}
  \noalign{\global\let\adl@draw\@dashdrawstore
           \vskip\belowrulesep}}
\makeatother

\usepackage{sidecap}
\sidecaptionvpos{figure}{t}
\usepackage{booktabs,multirow,array}
\newcolumntype{N}{@{}m{0pt}@{}}

%% file: sections/abstract.tex
\begin{abstract}

Controllable video generation has attracted significant attention, largely due to advances in video diffusion models. In domains such as autonomous driving, it is essential to develop highly accurate predictions for object motions. This paper tackles a crucial challenge of how to exert precise control over object motion for realistic video synthesis. To accomplish this, we 1) control object movements using bounding boxes and extend this control to the renderings of 2D or 3D boxes in pixel space, 2) employ a distinct, specialized model to forecast the trajectories of object bounding boxes based on their previous and, if desired, future positions, and 3) adapt and enhance a separate video diffusion network to create video content based on these high quality trajectory forecasts. Our method, \textbf{Ctrl-V}, leverages modified and fine-tuned Stable Video Diffusion (SVD) models to solve both trajectory and video generation. Extensive experiments conducted on the KITTI, Virtual-KITTI 2, BDD100k, and nuScenes datasets validate the effectiveness of our approach in producing realistic and controllable video generation.

\end{abstract}


%% file: sections/introduction.tex
Recent advances in controllable \emph{image} generation have enabled the creation of highly realistic images from various conditioning inputs, including points, bounding boxes, scribbles, segmentation maps, and skeleton poses. Yet, translating this control to \emph{video} generation is markedly more challenging due to the added temporal dimension. Incorporating time dynamics into diffusion models significantly complicates controllable video generation, as it requires accounting for object interactions, physical consistency, and coherent motion across frames.

Numerous recent studies have examined different forms of controllability for video generation. Researchers have used an array of methods for control, including conditioning on information such as canny edge and depth maps (\cite{zhang2023controlvideo}), similar visual information (\cite{chen2023controlavideo}), optical flow (\cite{hu2023videocontrolnet}), and pose sequences (\cite{karras2023dreampose}). These control inputs are often expensive to produce, especially when sequences of them are required in order to condition a video. Models that use accessible conditioning such as bounding boxes require additional input such as text to help with the generation process (\cite{wang2024boximator}). A controllable video generation model with an accessible and simple mode of control is greatly desired.

In this work, we focus on creating such a model. Specifically, we aim to generate higher fidelity videos controlled by, at the minimum, the beginning and ending positions of 2D and 3D bounding boxes without the help of other modes of control. Our two-part method includes a diffusion-based model that generates the motions and dynamics of objects in the form of bounding box videos (2D images of the bounding boxes evolving over time), and a generative model of videos according to those bounding box videos. To this end, we choose to train and test our model on driving datasets as they contain challenging scenes rich with different types of bounding boxes as well as complex movement and irregular appearing and disappearing objects. In our experiments, we show that our model generates videos that adhere tightly to the desired bounding box motion conditioning, accurately depicting desired object movements. Additionally, through our novel pixel-level bounding box generator and conditioning, our method robustly handles the appearance and disappearance of different objects in a scene, including cars, pedestrians, bikers, and others.

%
%
In this paper, we present Ctrl-V, a diffusion-based bounding box conditional video generation method that addresses multiple challenges and makes the following contributions to generate higher-fidelity videos using diffusion techniques. Our contributions can be enumerated as follows:
\begin{enumerate}[leftmargin=*,noitemsep,nolistsep]
\setlength{\itemsep}{1pt}
\setlist[enumerate]{leftmargin=0mm}
\item 
    \textbf{A Novel Diffusion Approach for Predicting Bounding Box Trajectories:} Our approach generates video frames with 2D or 3D bounding box \emph{trajectories} at the pixel-level based on their initial and final states, and the first frame of RGB video. Our results show that our proposed approach yields higher quality bounding box trajectory predictions than a recently proposed state of the art method (see in Table \ref{tab:predictor_scores}).
    \item 
    \textbf{2D and 3D Bounding Box Conditioned Video Generation with Diffusion:}
    We present a method that allows video generation by conditioning on 2D or 3D bounding box trajectories which allows fine-grained control over the generated videos. Our results (in Table \ref{tab:generation_quality}) indicate that this approach can generate video that is dramatically better than a state-of-the-art Stable Video Diffusion baseline fine-tuned without this conditioning mechanism across four commonly used quality metrics. Our approach further improves upon prior work by enabling the following capabilities:
    a. \textbf{Multi-subject Generation:}
    Synthesizing multiple subjects in videos poses significant challenges, requiring coherent object placement across frames, particularly during interactions. Existing models typically demonstrate capabilities with up to three subjects in online demo clips. Recent advances include: Boximator~\citep{wang2024boximator} (which can synthesize up to 8 subjects) and FACTOR~\citep{li2024trackdiffusion} (up to 12 subjects). Our method enables synthesis of scenes with any number of objects, limited only by the number of clearly renderable bounding boxes per frame;
    b. \textbf{Uninitialized Object Generation:} 
    Most bounding box-based generation methods focus solely on objects that either remain present throughout the entire clip or appear in the first frame and persist until at least the middle of the clip. They typically overlook bounding boxes that appear only after the first frame \citep{wang2024boximator}. In this work, we train our model to be sensitive to all bounding boxes, whether they are present from the first frame or appear from the middle of the clip. 
    \item
    \textbf{A New Benchmark for a New Problem Formulation:} Given the novelty of our problem formulation, there is no existing standard way to evaluate models that seek to predict vehicle video with high fidelity. We therefore present a new benchmark consisting of a particular way of evaluating video generation models using the KITTI~\citep{geiger2013KITTI}, Virtual KITTI 2 (vKITTI)~\citep{cabon2020vKITTI2}, the Berkeley Driving Dataset (BDD 100k)~\citep{yu2020bdd100k} and nuScenes~\citep{caesar2019nuscenes}.
\end{enumerate}

%% file: sections/related_work.tex

\textbf{Video latent diffusion models (VLDMs)} extend latent image diffusion techniques \citep{rombach2022highresolution} to video generation. Early VLDMs \citep{blattmann2023align, blattmann2023stable, he2023latent, zeng2023make, wu2023tuneavideo} shows temporally consistent frame generation and are tailored for text-prompted or image-prompted video generation. However, these models often struggle with complex scenes and lack the capability for precise local control.

\textbf{Conditional Video Diffusion} techniques providing a certain degree of control. Methods like VideoCompose \citep{wang2023videocomposer}, Dreamix \citep{molad2023dreamix}, Pix2Video \citep{ceylan2023pix2video}, and DreamPose \citep{karras2023dreampose} propose various designs of novel adapters on top of VLDMs in order to incorporate different conditioning to achieve frame-level control. \textbf{ControlNet Adapted Video Diffusion}, on the other hand, achieve precise regional or pixel-level control in video generation by utilizing ControlNet \citep{zhang2023controlnet} adapters within VLDM frameworks. Models such as Control-A-Video \citep{chen2023controlavideo}, Video ControlNet \citep{hu2023videocontrolnet, chu2023video}, ControlVideo \citep{zhang2023controlvideo}, and ReVideo \citep{mou2024revideo} show that these adapters are highly adaptable to various types of conditioning, easy to train, and allow for more precise manipulation and enhanced accuracy in editing and creating video content.

\textbf{Motion Control with bounding box Conditioning} There are many strategies of control that have been explored in controllable video generation research. 
Notably, ControlVideo \citep{zhang2023controlvideo} utilizes a training-free strategy that employs pre-trained image LDMs and ControlNets to generate videos based on \emph{canny edge and depth maps}. Control-A-Video \citep{chen2023controlavideo} leverages a controllable video LDM that combines a pre-trained text-to-video model with ControlNet to manipulate videos using \emph{similar visual information}.  Video ControlNets \citep{hu2023videocontrolnet, chu2023video} uses \emph{optical flow} information to enhance video generation, while ReVideo \citep{mou2024revideo} depends on extracted \emph{video trajectories}. DreamPose \citep{karras2023dreampose} injects \emph{pose sequence} information into the initial noise. VideoComposer \citep{wang2023videocomposer} uses an array of \emph{sketch, depth, mask,} and \emph{motion vectors} as conditioning. 


Many of these conditions, such as edge, depth, and optical flow maps, are costly to produce and lack the flexibility needed for customization. Bounding boxes emerge as a conditioning that are easily customizable and can be edited into different shape, size, locations and classes efficiently.
To the best of our knowledge, six other research projects are currently exploring the use of bounding boxes for motion control in video generation. However, it is important to note that our work is distinct from these in several critical respects.

\textbf{Direct-A-Video}, \textbf{TrailBlazer} \citep{ma2024trailblazer} and \textbf{Peekaboo} \citep{jain2024peekaboo} are different training-free approaches that employ attention map adjustments to direct the model in generating a particular object within a defined region. Direct-A-Video, in particular, is a text-to-video model that learns to control camera motion during training and then adopts a training-free approach to manipulate object movements using bounding boxes. \textbf{FACTOR} \citep{huang2023finegrained} augmented the transformer-based generation model, Phenaki \citep{villegas2022phenaki}, by integrating a box control module. TrailBlazer, Peekaboo and FACTOR necessitate textual descriptions for individual boxes, thus lacking direct visual grounding.

Our task setup shares mild similarities with \textbf{Boximator}\citep{wang2024boximator} and \textbf{TrackDiffusion}\citep{fischer2023qdtrack} because we also utilize bounding box conditioning during training without relying on text descriptions for individual boxes. However, our approach diverges from these text-to-video models, as our primary focus is on generating realistic videos conditioned only on a couple frames of bounding boxes, whereas Boximator and TrackDiffusion are designed to be conditioned on text information as they both are \textbf{text-to-video} models.
%
%
Boximator and TrackDiffusion enhance their models by introducing new self-attention layers to 3D U-Net blocks. These layers incorporate additional conditional information, such as box coordinates and object IDs, into the pretrained VLDM model. Their bounding box information is processed using a Fourier embedder \citep{mildenhall2020nerf}, which is then passed through multi-layer perceptron layers to encode. In contrast, our approach uses ControlNet and does not involve training additional encoding layers or utilizing Fourier embedder to handle the bounding box information.
Moreover, Boximator introduces a \emph{self-tracking technique} to ensure adherence to the bounding boxes in generated outputs, a technique also adopted by TrackDiffusion.  
This enables the network to learn the object tracking task alongside video generation, but requires a two-stage training process: one with target bounding boxes in frames, and another with the boxes removed. 
They demonstrate that without this technique, the model's performance markedly declines. Conversely, our model achieves alignment with the bounding box conditions without additional training. 

\textbf{Vehicle Oriented Generative Models} 
DriveDreamer \citep{wang2023drivedreamer} presents noteworthy contribution from autonomous driving domain. It takes an action-based approach to video simulation. It also makes use of bounding boxes and generate actions along  with a video rendering.
Within the DriveDreamer framework, Fourier embeddings \citep{mildenhall2020nerf} are also employed to encode bounding box information, and CLIP embeddings \citep{radford2021clip} are used for box categorization.
They focus on generating multiple camera views and do not condition on bounding box sequences, so cannot be directly compared with our problem setting.
In contrast, the DriveGAN work of \cite{kim2021drivegan} aims to learn a GAN based driving environment in pixel-space, complete with actions and an implicit model of dynamics encoded using the latent space of a VAE. While driving oriented, the approach does not focus on controlling the generation of vehicle video that respects well-defined object trajectories with high fidelity.

%% file: sections/method.tex
\subsection{Preliminaries}\label{sec:svd_intro}
We begin here with an overview of the Stable Video Diffusion (SVD) model \citep{blattmann2023stable}, due to its importance in our approach.
%
SVD is a diffusion based image-to-video (I2V) model performed in latent embedding space \cite{blattmann2023align}. Using an image \(\vf^{(0)}\) as initial condition, SVD is able to extend that single frame into a video \(\vf = [\vf^{(0)}, \ldots, \vf^{(N)}]\) where \(N\) is the length of the sequence.
Notably, SVD operates in latent space, where the diffusion and denoising process act upon the latents \(\vz\) of the video \(\vf\). Here, SVD employs an image encoder (\(\En\)) and an image decoder (\(\De\)) to translate each frame into and out of latent space: \( \De\big(\En(\vf^{(i)})\big) = \De(\vz^{(i)})\approx \vf^{(i)} \).
At each diffusion step, SVD progressively introduces noise into the latent representations. In this work, the amount of noise is dictated by Euler discrete noise scheduling method (EDM) introduced in \citet{karras2022elucidating}.
%
%
A UNet based denoiser network within the SVD is used to predict this noise in order to recover the original latent representations. The UNet, \(\sU_\theta\), is parameterized as:
\begin{equation}
    \sU_\theta\big(\hat{\vz}_t, \vz_{\text{pad}}^{(0)}, \vc^{(0)}, t\big),
    \label{eq:unet_input}
\end{equation}
\begin{itemize}
[leftmargin=*,noitemsep,nolistsep]
\setlength{\itemsep}{1pt}
\setlist[enumerate]{leftmargin=0mm}
    \item 
    \(\hat{\vz}_t\in\R^{N\times C'\times H'\times W'}\): latent representation of frames corrupted by noise at noise level \(t\).
    \item
    \(\vz^{(0)}\in\R^{1\times C'\times H'\times W'}\):  latent representation of the initial frame.
    \item 
    \(\vz_{\text{pad}}^{(0)} \in\R^{N\times C'\times H'\times W'}\): Padded \(\vz^{(0)}\) by repeating itself along the first dimension N times.
    \item
    \(\vc^{(0)}\): CLIP encoding \citep{radford2021clip} of the initial frame.
\end{itemize}

The full denoiser network, \(\sD_\theta\), with an EDM noise scheduler, is formulated as
\begin{equation}
    \sD_\theta(\vz; \vc^{(0)}, \sigma_t)=\lambda_{\text{skip}}(\sigma_t)\vz + \lambda_\text{out}(\sigma_t)\sU_\theta\big(\lambda_\text{in}(\sigma_t)\vz, \vz^{(0)}_{\text{pad}}, \vc^{(0)}; \lambda_\text{noise}(\sigma_t)\big)
    \label{eq:denoiser}
\end{equation}
Here \(\lambda_{\text{skip}}\), \(\lambda_{\text{out}}\), \(\lambda_{\text{in}}\) and \(\lambda_{\text{noise}}\) denote scaling functions, while \(\sigma_t\) represents the computed noise at level $t$. The precise mathematical definitions of these terms are detailed in Appendix~\ref{sec:svd_notation}.
Note that 3D UNet \(\sU_\theta\) in Equation \ref{eq:unet_input}, is a re-parameterized version of the one in 
Equation~\ref{eq:denoiser}~\citep{ronneberger2015unet}. The scaling terms are absorbed and the inputs are simplified for clarity. In the following sections, we follow the re-parameterized version in Equation \ref{eq:unet_input} when refering to the UNets in our model.

\subsection{Overview of our Method: Ctrl-V}

\begin{figure}[h]
    \centering
\includegraphics[width=\textwidth]{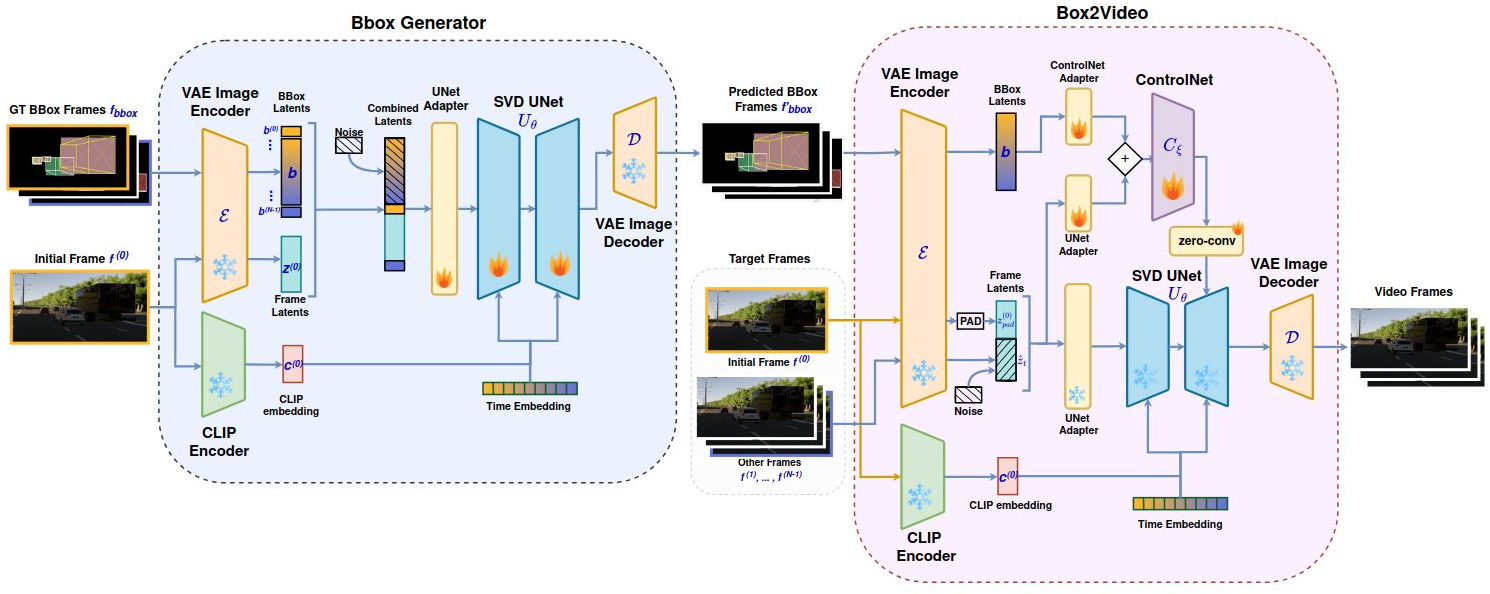}
    \caption{The diagram illustrates two components of \textbf{Ctrl-V}: (left) the{\color{SteelBlue}\textbf{~\modelbbox~}}and (right){\color{Orchid}\textbf{~\modelvid}}. For both models, we use a {\color{DarkOrange}\textbf{frozen, off-the-shelf VAE}} to encode images into latent space (\(\En\)) and decode them back into pixel space (\(\De\)). During training, (1) the {\color{SteelBlue}\textbf{\modelbbox~}}(Sec.~\ref{sec:bbox_predictor}) learns to denoise the noisy \gradientRGB{bounding box frame latents}{235, 183, 52}{57, 43, 186} \(\hat{\vb}_t\), conditioned on the {\color{Orange}first} (\(\vb^{(0)}\)) and {\color{Navy}last} (\(\vb^{(N-1)}\)) bounding box frame latents and the {\textcolor[HTML]{66CC00}{padded initial frame latent}} \(\vz_{pad}^{(0)}\) and (2) the{\color{Orchid}\textbf{~\modelvid~}}(Sec.~\ref{sec:controlnet}) denoises the \gradientRGB{target frame latents}{102, 204, 0}{247, 20, 213} \(\hat{\vz}_t\) by conditioning on the {\textcolor[HTML]{66CC00}{initial frame's latent}} \(\vz_{pad}^{(0)}\) (input to the \textcolor{svdunet}{SVD UNet}) and the  \gradientRGB{bounding box frame latents}{235, 183, 52}{57, 43, 186} \(\vb\) (input to the {\color{MediumPurple}ControlNet}). }
    \label{fig:model_diagram}
\end{figure}

Our controllable video generation method is illustrated in Figure~\ref{fig:model_diagram}. It consists of two sequential steps: 
\begin{enumerate}[leftmargin=*,noitemsep,nolistsep]
\setlength{\itemsep}{1pt}
\setlist[enumerate]{leftmargin=0mm}
    \item First, we generate bounding box frames using our diffusion based bounding box predictor network, the  \textbf{\modelbbox}, which is shown on the left side of  Figure~\ref{fig:model_diagram}. These frames contain only bounding boxes. They make up a video of moving (or stationary) bounding boxes and it serve as the ``skeleton" for the generated video.
    \item 
    Then, we generate a video using our video generator network, \textbf{\modelvid}, shown on the right side of Figure~\ref{fig:model_diagram}, where the bounding boxes frames act as the control signal. The bounding boxes in each frame determine the objects generated in the corresponding frames of the video.
\end{enumerate}

\modelbbox~and \modelvid~each utilizes a modified SVD backbone -- illustrated by the SVD backbone in Figure~\ref{fig:model_diagram}. These backbones are adapted to their respective generation tasks. Details of each model are presented in their individual sections: \modelbbox~-- Section ~\ref{sec:bbox_predictor} and \modelvid~-- Section ~\ref{sec:controlnet}.

\subsection{Ctrl-V: \modelbbox}\label{sec:bbox_predictor}
The \modelbbox~shown on the left in Figure~\ref{fig:model_diagram} aims to predict object bounding boxes across all video frames using an SVD backbone. The four inputs to the model are \(\hat{\vb}_t\), \(\vb^{(0)}\), \(\vb^{(N-1)}\), \(\vz^{(0)}\), 
where: \(\hat{\vb}_t\) is the encoded ``video" of bounding boxes with $t$ levels of noise added; \(\vb^{(0)}\) is the encoded initial bounding box frame(s); \(\vb^{(N-1)}\) is the encoded final bounding box frame; \(\vz^{(0)}\) is the encoded initial video frame. During training, the model learns to predict the noise added in \(\hat{\vb}_t\) according to the EDM noise scheduler. The model recovers the original \(\vb\) from its noisy version \(\hat{\vb}_t\) by calculating the noise with UNet outputs and eliminating the noise through scaling functions. We opt to abstract this detail in the model diagram for readability.


In practice, the four inputs are transformed and concatenated into a vector format accepted by the UNet adapter within the SVD backbone. Specifically, as shown in Figure \ref{fig:model_diagram}, \(\vz^{(0)} \in \mathbb{R}^{1 \times C' \times H' \times W'}\) is replicated along the first dimension, and its front and end (in the first dimension) are replaced by \(\vb^{(0)}\), \(\vb^{(N-1)}\) respectively. This forms \(\vz_\text{pad}^{(0)} = \text{concat}(\vb^{(0)}, \vz^{(0)},... , \vz^{(0)}, \vb^{(N-1)}) \in \mathbb{R}^{N \times C' \times H' \times W'}\). The noise-added encoding of bounding box video \(\hat{\vb}_t\) is then concatenated with \(\vz_\text{pad}^{(0)}\) to form the final input to the UNet adapter.
%
The network incorporates additional conditioning inputs, including a CLIP-encoded embedding of the initial frame \(\vc^{(0)}\) and a noise-level embedding \(t\). These embeddings are individually integrated into every sub-block of the U-Net through a self-attention mechanism.

\subsubsection*{Representing Bounding Boxes in Pixel Space}\label{sec:bbox_preprocess}

An important element of Ctrl-V is our design choice of rendering bounding boxes in pixel space. The manner in which bounding box information is provided as a control signal to the video generator is important. For example, prior work such as Boximator \citep{wang2024boximator} represents bounding boxes as a Fourier transformed concatenated vector of their raw coordinates, ID and other information. 
In contrast, in our work we choose to render bounding boxes into frames while maintaining minimal loss of meta information. Importantly, we also encode information such as track ID, object type, and orientation for each bounding box using a combination of visual attributes, including border color, fill color, and markings. Specifically, the \textit{track ID} represents a unique identifier for each tracked object across frames, the \textit{object type} specifies the category of the object (e.g., car, pedestrian), and the \textit{orientation} indicates the direction the object is facing. Further details about how these bounding box frames are rendered can be found in Appendix~\ref{sec:bbox_plot_procedure}. 
Crucially, our approach allows us to leverage the highly effective ControlNet approach to provide pixel-level guidance to influence diffusion generated imagery.


\subsection{Ctrl-V: \modelvid}\label{sec:controlnet}
\modelvid~is shown on the right in Figure \ref{fig:model_diagram} and it aims to generate high-fidelity videos controlled by bounding box frames, such as those generated by the \modelbbox~network. \modelvid~consists of an SVD backbone for video generation, and an adapted ControlNet module to process the bounding box control signal. ControlNet is a widely used network for controlling image generation. In this work, we modify ControlNet and adapt it to the video diffusion framework (as shown on the right in Figure~\ref{fig:model_diagram}). This architecture allows us to train \modelvid~in a single stage without the need for additional optimization criteria, in contrast to previous work such as Boximator and TrackDiffusion \citep{wang2024boximator, li2024trackdiffusion}, which require multi-stage learning with extra criteria to train their models. 

The SVD component takes two inputs: \(\vz^{(0)}\) and \(\hat{\vz}_t\). Here, \(\vz^{(0)}\) is the encoded initial video frame and \(\hat{\vz}_t\) is the encoded full video with $t$ levels of noise added to it. As shown in Figure \ref{fig:model_diagram}, we process these inputs by padding \(\vz^{(0)}\) by repeating it along the first dimension before concatenating it with \(\hat{\vz}_t\) to create the final input to the UNet adapter of the SVD. 
The same input is also sent to the ControlNet module through its own UNet adapter layers. Additionally, ControlNet also receives the encoded bounding box frames, $\vb$, as input, through ControlNet adapter layers. Both of these transformed input is then added together before processed by the ControlNet module. The output signal of the ControlNet module then goes through a zero-convolution before being sent to the SVD UNet decoder layers through residual paths as control signal.
%
%
%
%
During training, the weights of the SVD model (\(\theta\)) are frozen, while only the weights in the ControlNet (\(\xi\)) are updated.

%% file: sections/experiments.tex

\begin{figure}[h]
    \setlength\tabcolsep{3pt} 
    \centering
    \small
    \begin{tabular}{@{} r M{0.16\linewidth} M{0.16\linewidth} M{0.16\linewidth} M{0.16\linewidth} M{0.16\linewidth} @{}}
    & \textbf{Frame 1} & \textbf{Frame 7} & \textbf{Frame 13} & \textbf{Frame 19}  & \textbf{Frame 25}\\
    \begin{tabular}{@{}r@{}}\modelbbox's\\Output: Generated\\BBox Trajectory \end{tabular}  & \includegraphics[width=\hsize]{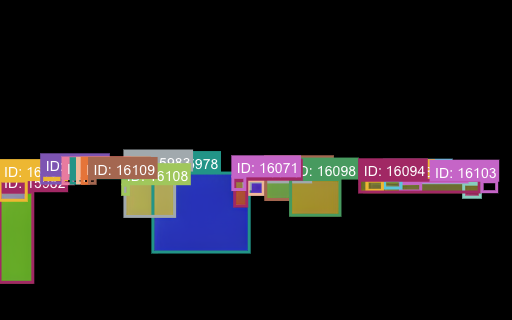}
      & \includegraphics[width=\hsize]{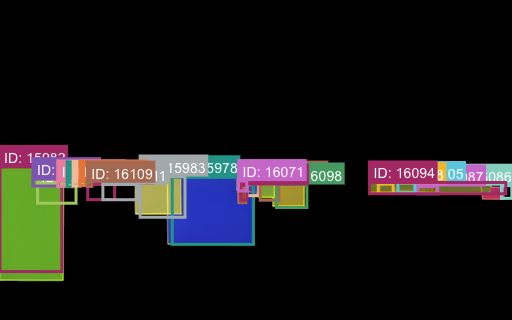}
      & \includegraphics[width=\hsize]{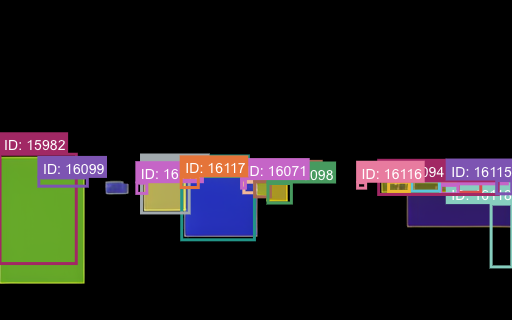}
      & \includegraphics[width=\hsize]{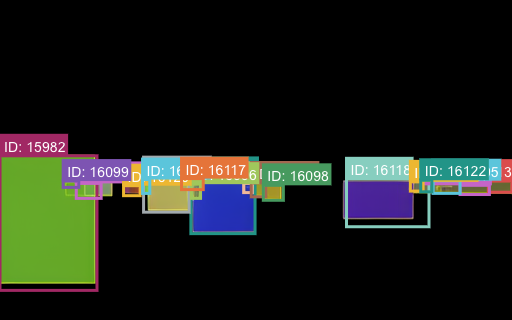}    
      & \includegraphics[width=\hsize]{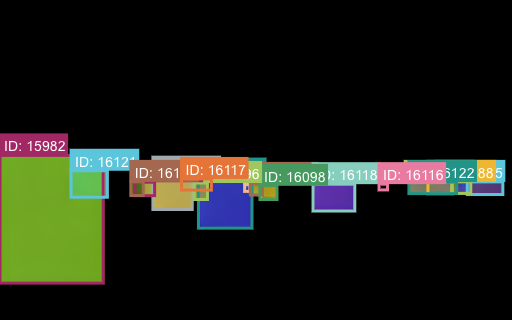}\\
      \begin{tabular}{@{}r@{}}\modelvid's\\Output: \\ Generated Frames \end{tabular} & \includegraphics[width=\hsize]{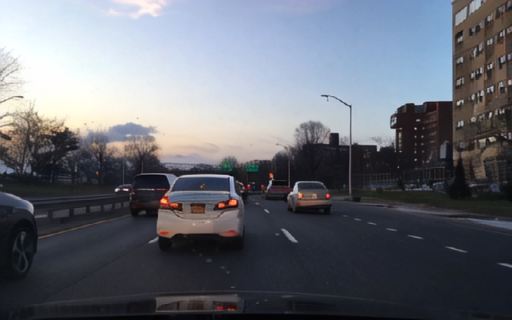}    
      & \includegraphics[width=\hsize]{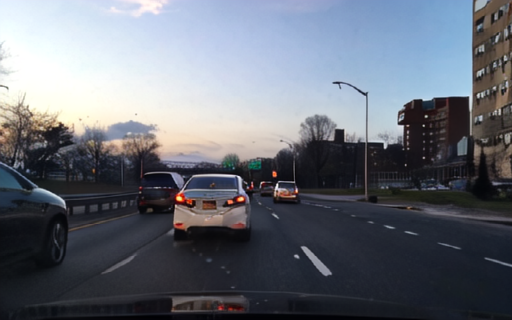}   
      & \includegraphics[width=\hsize]{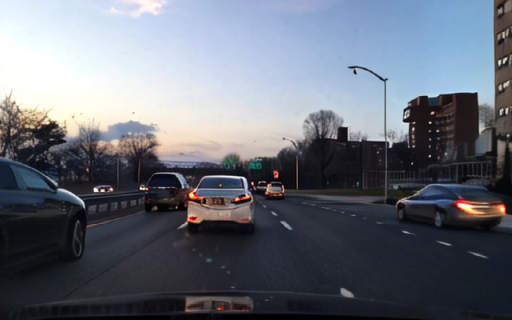}
      & \includegraphics[width=\hsize]{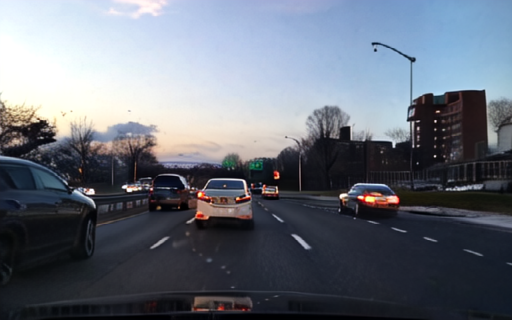}    
      & \includegraphics[width=\hsize]{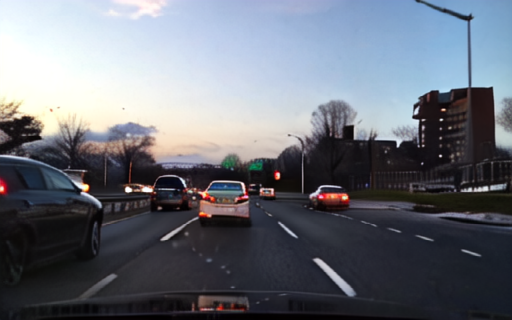}\\
      \begin{tabular}{@{}r@{}}Ground Truth\\Frames \end{tabular} & \includegraphics[width=\hsize]{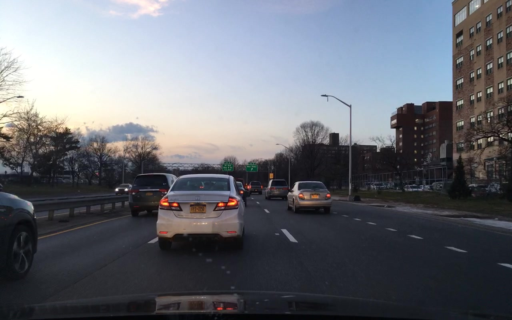}
      & \includegraphics[width=\hsize]{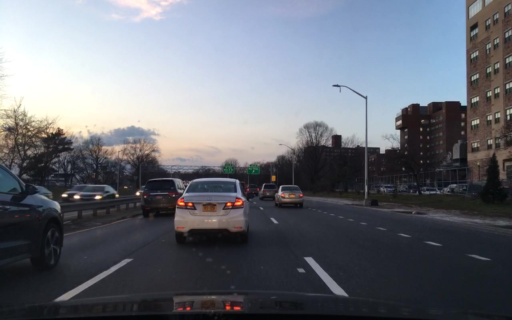}   
      & \includegraphics[width=\hsize]{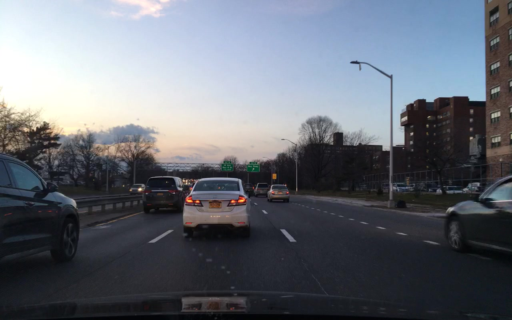}
      & \includegraphics[width=\hsize]{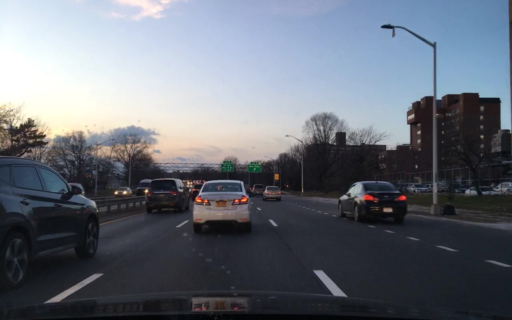}
      & \includegraphics[width=\hsize]{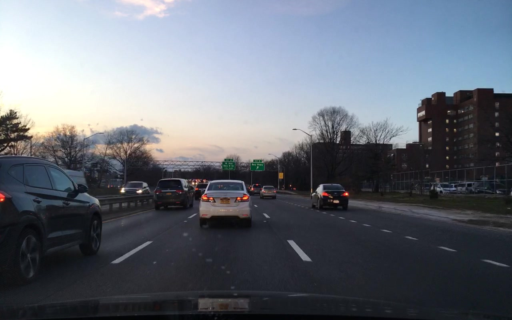} \\
\begin{tabular}{@{}r@{}}SVD Baseline\\Generated Frames \end{tabular} & \includegraphics[width=\hsize]{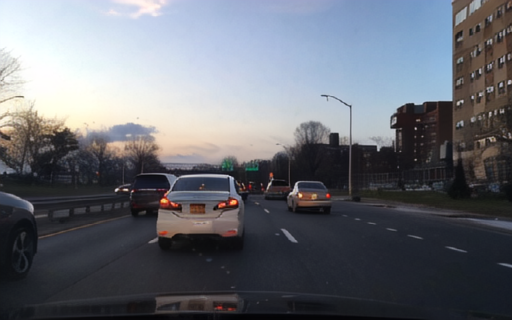}    
      & \includegraphics[width=\hsize]{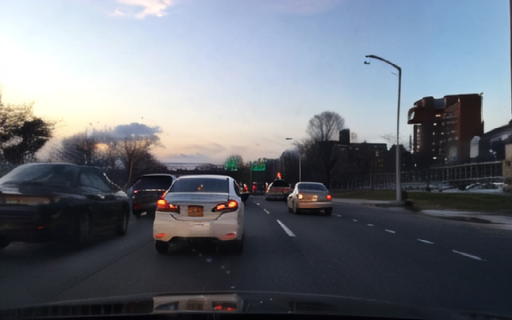}   
      & \includegraphics[width=\hsize]{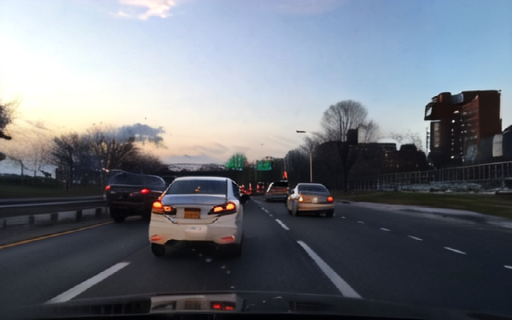}
      & \includegraphics[width=\hsize]{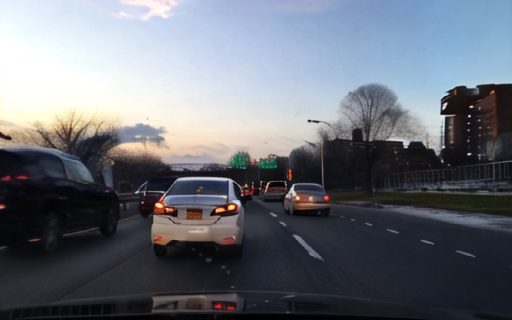}    
      & \includegraphics[width=\hsize]{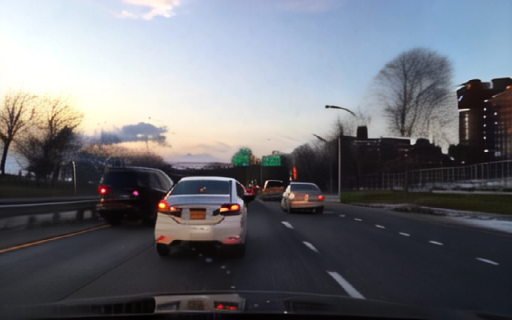}\\
      
    \end{tabular}
    \caption{The first two rows illustrate video samples generated using the Ctrl-V pipeline, with one initial frame, three initial bounding box frames, and one final bounding box frame as input. The first row shows bounding box trajectories from the BBox-generator in pixel space (solid rectangles for predictions, wireframe rectangles for ground truth). The second row presents frames generated by the Box2Video model, conditioned on the BBox-generator's output. The third row displays ground-truth frames, while the fourth row shows frames generated by the Stable Video Diffusion (SVD) baseline. \emph{In the Ctrl-V video, the car with the bright-green bounding box, which initially pokes out its nose in the lane to the left of the ego car, stays beside the ego car in the final frame. Meanwhile, the silver car with the olive bounding box, which starts in the lane to the right of the ego car, speeds off and is replaced by a new car (purple bounding box) entering the frame. These generated frames closely match the car positions seen in the conditioned inputs. In contrast, the SVD-generated video shows the black car on the left accelerating and moving ahead of the ego car, while the silver car remains in the same relative position to the ego car throughout.}}
    \label{fig:bbox_prediction_demos}%
\end{figure}

For quantitative evaluation, we assess the model's performance across four driving datasets on \textbf{three key aspects}:
 \begin{enumerate}[leftmargin=*,noitemsep,nolistsep]
\setlength{\itemsep}{1pt}
\setlist[enumerate]{leftmargin=0mm}
     \item 
     \emph{The overall visual quality of the generated results} (Section ~\ref{sec:generation_quality})
     \item 
     \emph{The alignment of the predicted bounding box trajectories with the ground truth} (Section ~\ref{sec:bbox_evaluation})
     \item 
     \emph{The fidelity of the generated objects in the video to the bounding box control signal} (Section ~\ref{sec:motion_evaluation})
 \end{enumerate}
For visual assessment, Figure~\ref{fig:bbox_prediction_demos} and Appendix~\ref{sec:generation_results} showcase sample demonstrations generated by our model.
To assess video quality, we randomly select 200 initial frames from each dataset's testing set and generate videos. The results in this section are based on analyses of these 200 generated videos per dataset.
Furthermore, we explored different bounding box conditioning options: one or three initial bounding box frames, followed by a single final bounding box. Additional variations are discussed in Appendix~\ref{sec:additional_bdd}.

\subsection{Datasets}\label{sec:datasets}
We evaluate the performance of our models across four autonomous-vehicle datasets: KITTI~\citep{geiger2013KITTI}, Virtual KITTI 2 (vKITTI)~\citep{cabon2020vKITTI2}, Berkeley Driving Dataset (BDD)~\citep{yu2020bdd100k} with Multi-object Tracking labels (MOT2020), and the nuScenes Dataset~\citep{caesar2019nuscenes}.
KITTI comprises 22 real-world driving clips with 3D object labelling. vKITTI consists of 5 virtual simulated driving scenes, each offering 6 weather variants, all including 3D object labelling. BDD is a large-scale real-world driving dataset, featuring 1603 2D-labeled sequences of driving videos. The nuScenes dataset is a large-scale driving dataset that includes 1000 scenes 20-second scenes annotated with 3D bounding boxes, multiple sensor data (lidar, radar and cameras) and map information. Further details on dataset configurations are provided in Appendix~\ref{sec:dataset_configuration}.

\subsection{\modelbbox: Quantitative Evaluation~\label{sec:bbox_evaluation}}
\input{sections/experiments/bbox_predictor}

\subsection{Ctrl-V: Generation Quality~\label{sec:generation_quality}}
\input{sections/experiments/generation_quality}

\subsection{\modelvid: Motion Control Evaluation~\label{sec:motion_evaluation}}
\input{sections/experiments/motion_control}

%% file: sections/experiments/bbox_predictor.tex
\begin{figure}[h]
    \setlength\tabcolsep{3pt} 
    \centering
    \small
    \begin{tabular}{@{} r M{0.16\linewidth} M{0.16\linewidth} M{0.16\linewidth} M{0.16\linewidth} M{0.16\linewidth} @{}}
    & \textbf{Frame 1} & \textbf{Frame 7} & \textbf{Frame 13} & \textbf{Frame 19}  & \textbf{Frame 25}\\
    \begin{tabular}{@{}r@{}}Trajectory\\Generation \#1: \end{tabular}  & \includegraphics[width=\hsize]{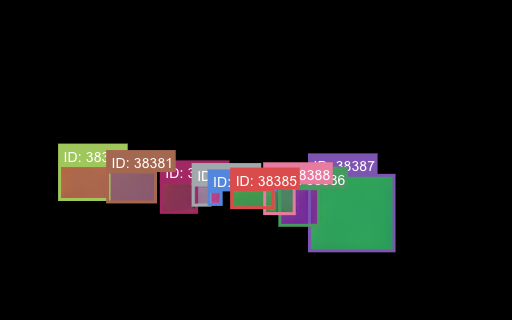}
      & \includegraphics[width=\hsize]{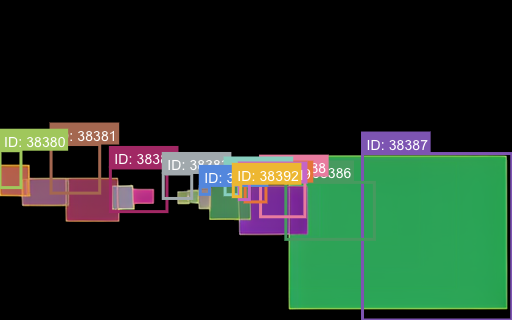}
      & \includegraphics[width=\hsize]{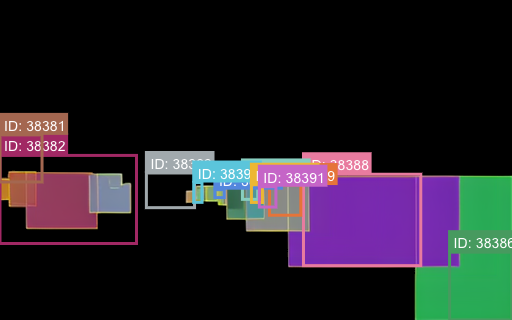}
      & \includegraphics[width=\hsize]{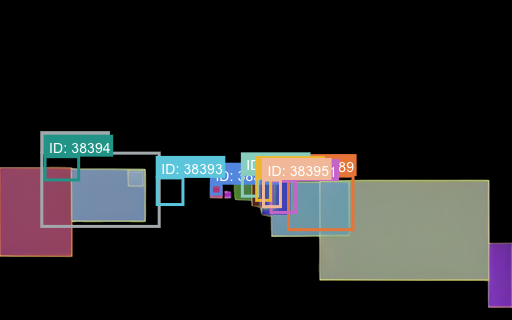}    
      & \includegraphics[width=\hsize]{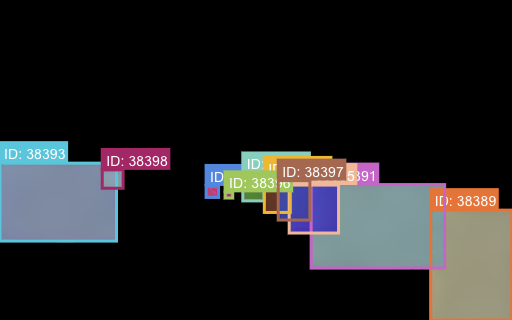}\\
      \begin{tabular}{@{}r@{}}Trajectory\\Generation \#2: \end{tabular} & \includegraphics[width=\hsize]{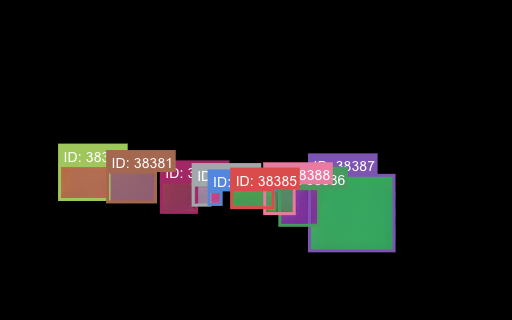}    
      & \includegraphics[width=\hsize]{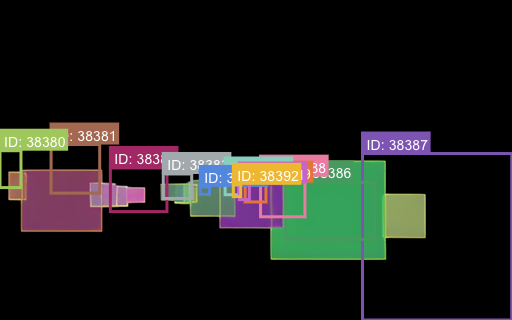}   
      & \includegraphics[width=\hsize]{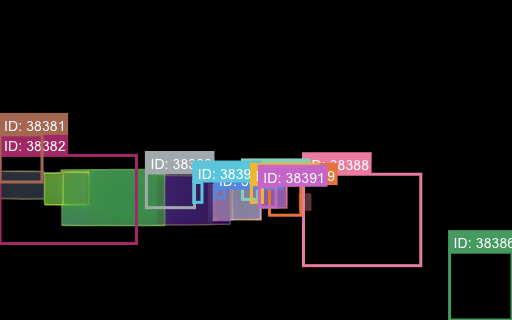}
      & \includegraphics[width=\hsize]{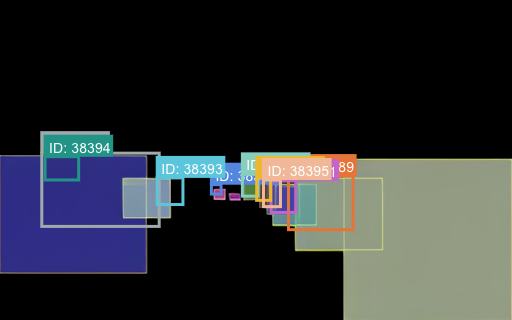}    
      & \includegraphics[width=\hsize]{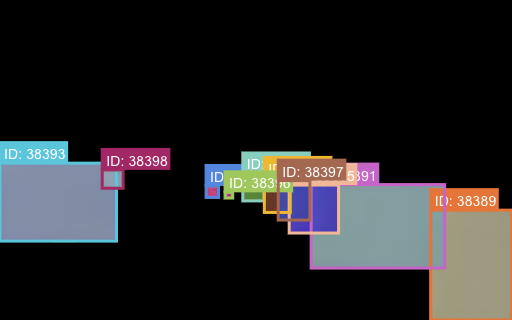}\\
    \end{tabular}
    \caption{This figure visualizes two samples of bounding box trajectories generated by the \modelbbox, conditioned on the same set of three initial bounding box frames and one final bounding box frame (solid rectangles represent predictions, and wireframe rectangles represent ground truth). Although the intermediate frames show notable differences, the initial and final frames align closely with the ground-truth bounding boxes.\label{fig:bbox_trajectory}}
\end{figure}
    
\begin{table}[h]
\centering
\resizebox{\textwidth}{!}{\begin{tabular}{lcccccccc}
\toprule
 & \textbf{Method} & \specialcell[]{\textbf{\# Cond.}\\\textbf{BBox}} & \textbf{maskIoU\(\uparrow\)} & \textbf{maskP\(\uparrow\)} & \textbf{maskR\(\uparrow\)} & \specialcell[]{\textbf{maskIoU\(\uparrow\)}\\(first+last)} & \specialcell[]{\textbf{maskP\(\uparrow\)}\\(first+last)} &  \specialcell[]{\textbf{maskR\(\uparrow\)}\\(first+last)} \\
 \midrule
\multirow{4}{*}{\STAB{\rotatebox[origin=c]{90}{KITTI}}} & BBox Generator (ours) & \multirow{2}{*}{1-to-1} & \(\textbf{.629}\pm.212\) & \(\textbf{.758}\pm.176\) & \(\textbf{.763}\pm .188 \) & \(\textbf{.986}\pm.012\) & \(\textbf{.994}\pm.008\) &\(\textbf{.992}\pm.009\)\\
 & Trajeglish-Style & & \(.447\pm .154\) & \(.568\pm .172\) & \(.679\pm .177\) & \(.561\pm .151\) & \(.663\pm .150\) & \(.789\pm .165\) \\
 \cmidrule(lr){2-9}
 & BBox Generator (ours) & \multirow{2}{*}{3-to-1} & \(\textbf{.795}\pm.112\) & \(\textbf{.881}\pm.082\) & \(\textbf{.884}\pm.078\) & \(\textbf{.986}\pm.010\)& \(\textbf{.992}\pm.007\) & \(\textbf{.994}\pm.005\)\\
 & Trajeglish-Style & & \(.491\pm .164\) & \(.622\pm .173\) & \(.691\pm .175\) & \(.576\pm .154\) & \(.684\pm .149\) & \(.784\pm .163\) \\
 \midrule 
 \midrule
 \multirow{4}{*}{\STAB{\rotatebox[origin=c]{90}{vKITTI}}} & BBox Generator (ours) & \multirow{2}{*}{1-to-1} & \(\textbf{.710}\pm.205\) & \(\textbf{.828}\pm .178\) & \(\textbf{.809}\pm.171\) & \(\textbf{.943}\pm.048\) & \(\textbf{.946}\pm.046\) & \(\textbf{.997}\pm.006\)\\
 & Trajeglish-Style & & \(.471\pm .171\) & \(.578\pm .200\) & \(.700\pm .187\) & \(.557\pm .171\) & \(.628\pm .194\) & \(.835\pm .135\) \\
 \cmidrule(lr){2-9}
  & BBox Generator (ours) & \multirow{2}{*}{3-to-1} & \(\textbf{.767}\pm.131\) & \(\textbf{.881}\pm.126\) & \(\textbf{.853}\pm.078\)  & \(\textbf{.944}\pm.039\)  &   \(\textbf{.948}\pm.036\)   &  \(\textbf{.996}\pm .006\)\\
  & Trajeglish-Style & & \(.520\pm .162\) & \(.630\pm .186\) & \(.741\pm .176\) & \(.575\pm .154\) & \(.657\pm .182\) & \(.836\pm .143\) \\
 \midrule
 \midrule
 \multirow{4}{*}{\STAB{\rotatebox[origin=c]{90}{BDD}}} & BBox Generator (ours) & \multirow{2}{*}{1-to-1} & \(\textbf{.587}\pm .214\)& \(\textbf{.747}\pm.187\)& \(\textbf{.712}\pm.194\) & \(\textbf{.954}\pm.047\) & \(\textbf{.955}\pm.047\)&\(\textbf{.999}\pm.002\)\\
 & Trajeglish-Style & & \(.305\pm .183\) & \(.372\pm .213\) & \(.658\pm .207\) & \(.432\pm .171\) & \(.483\pm .192\) & \(.840\pm .166\) \\
 \cmidrule(lr){2-9}
 \cmidrule(lr){2-9}
 & BBox Generator (ours) & \multirow{2}{*}{3-to-1} & \(\textbf{.647}\pm.176\)   & \(\textbf{.784}\pm.150\) &  \(\textbf{.783}\pm.156\) &  \(\textbf{.955}\pm.043\) & \(\textbf{.955}\pm.042\) &  \(\textbf{.997}\pm.001\) \\
 & Trajeglish-Style & & \(.373\pm .185\) & \(.454\pm .206\) & \(.686\pm .193\) & \(.492\pm .190\) & \(.553\pm .208\) & \(.842\pm .154\) \\
 \midrule 
 \midrule
 \multirow{4}{*}{\STAB{\rotatebox[origin=c]{90}{nuScenes}}} & BBox Generator (ours) & \multirow{2}{*}{1-to-1} & \(.364\pm.242\) & \(.433\pm .278\) & \(\textbf{.740}\pm.186\) & \(\textbf{.983}\pm.013\) & \(\textbf{.985}\pm.0112\) & \(\textbf{.997}\pm.003\)\\
 & Trajeglish-Style & & \(\textbf{.405}\pm .202\) & \(\textbf{.506}\pm .220\) & \(.661\pm .216\) & \(.511\pm .168\) & \(.603\pm .172\) & \(.789\pm .195\) \\
 \cmidrule(lr){2-9}
  & BBox Generator (ours) & \multirow{2}{*}{3-to-1} & \(\textbf{.827}\pm.150\) & \(\textbf{.892}\pm.120\) & \(\textbf{.906}\pm.099\)  & \(\textbf{.983}\pm.013\)  &   \(\textbf{.985}\pm.012\)   &  \(\textbf{.998}\pm .003\)\\
  & Trajeglish-Style & & \(.448\pm .194\) & \(.554\pm .213\) & \(.695\pm .196\) & \(.529\pm .172\) & \(.623\pm .177\) & \(.791\pm .192\) \\
\bottomrule
\end{tabular}}
\caption{Comparing real and generated bounding boxes. We condition on 1 or 3 initial bounding box frame(s) and 1 final bounding box or trajectory frame. The first three columns show evaluations on the entire generated bounding box sequence, measuring the alignment scores between our generated bounding box generations and ground-truth labels. The last three columns focus on testing the auto-encoding capability of the network, evaluating only the first and last frames of the generated sequence. ``BBox Generator" is our method and ``Trajeglish-Style" is a baseline inspired from \cite{philion2023trajeglish} (see Appendix~\ref{sec:trajeglish} for implementation details on this baseline).}
\label{tab:predictor_scores}
\end{table}

Figure \ref{fig:bbox_trajectory} showcases two bounding box trajectory samples generated by the ~\modelbbox, conditioned on the same initial and final bounding box frames. To evaluate the quality of our bounding box generations, we create mask images for both the ground-truth and generated bounding box sequences. The mask images are generated by converting the bounding box frames into binary masks (details can be found in Appendix~\ref{sec:frame2bimask}). We then calculate the generated averaged mask Intersection over Union (maskIoU) scores, averaged mask Precision (maskP) scores, and averaged mask Recall (maskR) scores against the ground-truth bounding box masks. To assess our bounding box trajectories, we applied the ``best-out-of-K" method, selecting the model with the highest maskIoU score for evaluation. In this instance, K equals 5. We compare our results with a baseline referred to as the ``Trajeglish-Style" model, an autoregressive GPT-like encoder-decoder that models the bounding box trajectories as a sequence of discrete motion tokens. This baseline is inspired by the work of~\cite{philion2023trajeglish} with 
implementation details provided in Appendix~\ref{sec:trajeglish}.
We present our findings in Table~\ref{tab:predictor_scores}, and demonstrate examples of our bounding box generations on each dataset in Appendix~\ref{sec:generation_results}.

In the bounding box generation figures, our generator model achieves the closest alignment with the ground-truth in the first and last frames. This near-perfect alignment is primarily attributed to conditioning the model on the bounding boxes of these key frames. When considering all generated frames, the alignment scores decrease, as shown by the plotted demonstrations and metric results in Table~\ref{tab:predictor_scores}. This is because objects in frames do not move deterministically. \emph{The role of the bounding box generator is to generate a plausible trajectory for moving objects from the initial bounding box frame to the last.}

Despite the disparity between the ground-truth trajectory and the generated trajectory, our \modelvid ~consistently generates high-fidelity videos based on either trajectory provided. Further analysis of this aspect is provided in the subsequent sections.

%% file: sections/experiments/generation_quality.tex
\begin{table}[h]
    \centering
    \resizebox{\textwidth}{!}{\begin{tabular}{cclccccc}
    \toprule
      & & \textbf{Pipeline}& \textbf{\# Cond. BBox} & \textbf{FVD\(\downarrow\)} & \textbf{LPIPS\(\downarrow\)} & \textbf{SSIM\(\uparrow\)} & \textbf{PSNR\(\uparrow\)}\\
    \midrule
    & \multirow{5}{*}{\STAB{\rotatebox[origin=c]{90}{KITTI}}} & Stable Video Diffusion Baseline~\citep{blattmann2023stable} & None & 1118.4 & 0.4575 & 0.2919 & 10.63 \\
    & & Stable Video Diffusion Fine-tuned~\citep{blattmann2023stable} & None & 552.7 & 0.3504 & 0.4030 & 13.01 \\
    & & Ctrl-V: \modelbbox~+ \modelvid (Ours) & 1-to-1 & 467.7 & 0.3416 &  0.3241 & 13.21 \\
    & & Ctrl-V: \modelbbox~+ \modelvid (Ours) & 3-to-1 & {\bfseries 422.2} & {\bfseries 0.2981} & 0.4277 & 13.85   \\
    & & Ctrl-V: Teacher-forced \modelvid (Ours) & All & 435.6 & 0.2963 & {\bfseries 0.4394} & {\bfseries 14.10} \\
    \midrule
    & \multirow{5}{*}{\STAB{\rotatebox[origin=c]{90}{vKITTI}}} & Stable Video Diffusion Baseline~\citep{blattmann2023stable} & None & 922.7 & 0.3636 & 0.4740 & 14.61 \\
    & & Stable Video Diffusion Fine-tuned~\citep{blattmann2023stable} & None & 331.0 & 0.2852 & 0.5540 & 16.60 \\
    & & Ctrl-V: \modelbbox~+ \modelvid (Ours) & 1-to-1 & 400.2 & 0.3179 & 0.4714 & 15.78\\
    & & Ctrl-V: \modelbbox~+ \modelvid (Ours) & 3-to-1 & 341.4 & 0.2645 & 0.5841 & 17.60\\
    & & Ctrl-V: Teacher-forced \modelvid (Ours)  & All & {\bfseries 313.3} & {\bfseries 0.2372} & {\bfseries 0.6203} & {\bfseries 18.41} \\
    \midrule
    & \multirow{5}{*}{\STAB{\rotatebox[origin=c]{90}{BDD}}} & Stable Video Diffusion Baseline~\citep{blattmann2023stable} & None & 933.6 &  0.4880 & 0.3349 & 12.70 \\
    & & Stable Video Diffusion Fine-tuned~\citep{blattmann2023stable} & None & 409.0 &  0.3454 & 0.5379 & 16.99 \\
    & & Ctrl-V: \modelbbox~+ \modelvid (Ours) & 1-to-1 & 412.8 & 0.2967 &  0.5470 & 17.52\\
    & & Ctrl-V: \modelbbox~+ \modelvid (Ours) & 3-to-1 & 373.1 &  0.3071 & 0.5407 & 17.37\\
    & & Ctrl-V: Teacher-forced \modelvid (Ours) & All & {\bfseries 348.9} & {\bfseries 0.2926} & {\bfseries 0.5836} & {\bfseries 18.39} \\
    \midrule
    \multirow{14}{*}{\STAB{\rotatebox[origin=c]{90}{nuScenes}}} & \multirow{7}{*}{\STAB{\rotatebox[origin=c]{90}{Single-View}}} & Stable Video Diffusion Baseline~\citep{blattmann2023stable} & None & 1179.4 &  0.5004 & 0.2877 & 13.31 \\
    & & Stable Video Diffusion Fine-tuned~\citep{blattmann2023stable} & None & 316.6 &  0.2730 & 0.4787 & 18.58 \\
    & & Ctrl-V: \modelbbox~+ \modelvid (Ours) & 1-to-1 & 285.3 & 0.2647 &  0.5050 & 18.93\\
    & & Ctrl-V: \modelbbox~+ \modelvid (Ours) & 3-to-1 & {\bfseries 235.0} &  0.2235 & 0.5500 & 20.33\\
    & & Ctrl-V: Teacher-forced \modelvid (Ours) & All & 235.5 & {\bfseries 0.2104} & {\bfseries 0.5705} & {\bfseries 23.36} \\
    & & DriveGAN \citep{kim2021drivegan} & None & 390.8 & - & - & - \\
    & & DriveDreamer \citep{wang2023drivedreamer} & All & 340.8 & - & - & - \\
    \cmidrule{2-8}
    & \multirow{7}{*}{\STAB{\rotatebox[origin=c]{90}{Multi-view}}} & WoVoGen \citep{lu2023wovogen}  & All & 417.7 & - & - & - \\
    & & Drivingdiffusion \citep{li2023drivingdiffusion}  & All & 332.0 & - & - & - \\
    & & Drive-WM \citep{lu2023wovogen}  & None & 212.5 & - & - & - \\
    & & BEVWorld \citep{zhang2024bevworld}  & None & 154.0 & - & - & - \\
    & & Panacea  \citep{wen2024panacea}  & All & 139.0 & - & - & - \\
    & & Drive-WM \citep{lu2023wovogen}  & All & 122.7 & - & - & - \\
    & & DriveDreamer-2 \citep{zhao2024drivedreamer}  & None & {\bfseries 105.1} & - & - & - \\
    \bottomrule
    \end{tabular}}
    \caption{Comparing the quality and diversity of the generated video models. The generated videos consist of 25 frames (except for our nuScenes models which consist of 11 frames videos at 4 Hz) at a resolution of \(312\times 520\), while the reported metrics from this table are evaluated at a resolution of \(256\times 410\). The ``\# Cond. BBox" column reports the number of ground-truth input bounding box frames used by the generation pipelines. ``None" indicates that no ground-truth frames are used, while ``All" indicates that all ground-truth bounding box frames are utilized. If ``\# Cond. BBox" is \(n\text{-to-}m\), then it represents the number of initial bounding box frames used by the pipeline is \(n\) and the number of final bounding box frames used by the pipeline is \(m\). 
    }
    \label{tab:generation_quality}
\end{table}

To assess the quality of video generation, we compare videos generated through 4 distinct pipelines: 
\begin{enumerate}[leftmargin=*,noitemsep,nolistsep]
\setlength{\itemsep}{1pt}
\setlist[enumerate]{leftmargin=0mm}
    \item \textbf{Pre-trained Stable Video Diffusion (SVD) baselines\footnote{Stable Video Diffusion (SVD) baseline is an image-to-video (I2V) model that generates a video sequence conditioned on a single video frame.} without fine-tuning} (initial frame $\rightarrow$ video) 
    \item \textbf{Fine-tuned Stable Video Diffusion (SVD) baselines on the provided dataset} (initial frame $\rightarrow$ video)
    \item \textbf{Teacher-forced \modelvid ~generation} (initial frame and all bounding box frames $\rightarrow$ video)
    \item \textbf{bounding box generation with \modelbbox ~and \modelvid} ~(initial frame, one or three initial and one last bounding box frames $\rightarrow$ in-between bounding box frames and video).
\end{enumerate}

We evaluate our generation across four metrics: Fr\'echet Video Distance (FVD)~\citep{unterthiner2019fvd}, Learned Perceptual Image Patch Similarity (LPIPS)~\citep{zhang2018lpips}, Structural Similarity Index Measure (SSIM)~\citep{wang2004ssim} and Peak Signal-to-Noise Ratio (PSNR). These metrics either measure the consistency of frame pixels with the ground truth or the consistency of the frame latents extracted by another network. FVD\footnote{FVD is highly sensitive to video configuration parameters—such as frame rate, clip duration, and spatial resolution—making direct comparisons of FVD values across studies challenging. Additionally, the metric's sensitivity to sample sizes raises concerns, as some datasets may lack sufficient samples for convergence, leading to unreliable estimates.} is an exception; it evaluates the generation distribution against the ground truth's distribution. 
It is important to note that while many papers report their best-out-of-K results on these metrics, due to computational constraints, we evaluate our model on a single sample for each input.

The evaluated results are reported in Table~\ref{tab:generation_quality} and visualizations are available in Appendix~\ref{sec:gq_compared_svd_vs_ctrl}. These results indicate that the generation quality improves as we condition on more ground-truth bounding box frames. Details regarding the metrics and their limitations are discussed in Appendix~\ref{sec:quality_metrics}.

%% file: sections/experiments/motion_control.tex
\begin{figure}[h]
    \setlength\tabcolsep{3pt} 
    \centering
    \small
    \begin{tabular}{@{} r M{0.16\linewidth} M{0.16\linewidth} M{0.16\linewidth} M{0.16\linewidth} M{0.16\linewidth} @{}}
    & \textbf{Frame 1} & \textbf{Frame 7} & \textbf{Frame 13} & \textbf{Frame 19}  & \textbf{Frame 25}\\
    KITTI & \includegraphics[width=\hsize]{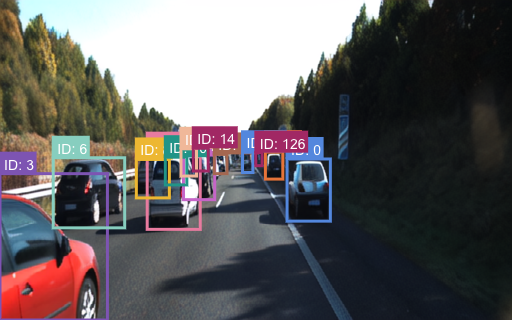}
      & \includegraphics[width=\hsize]{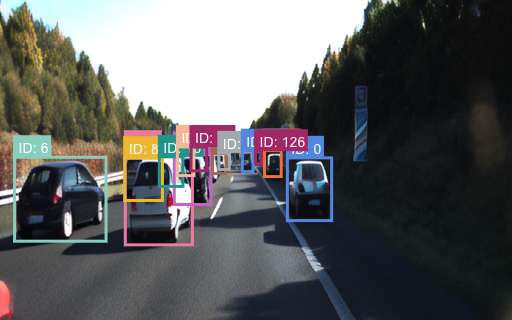}
      & \includegraphics[width=\hsize]{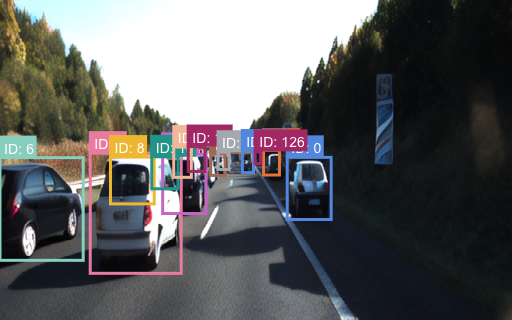}  
      & \includegraphics[width=\hsize]{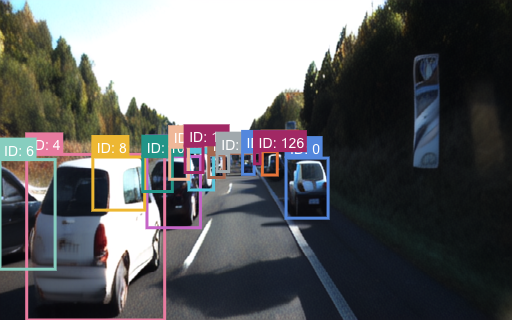}
      & \includegraphics[width=\hsize]{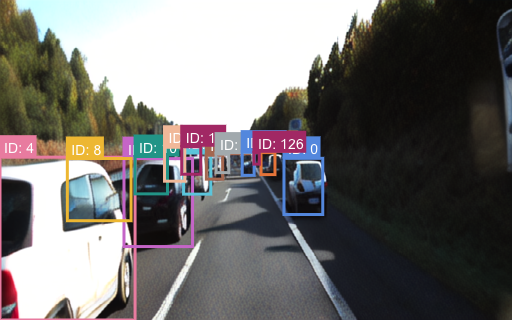}\\ 
      vKITTI & \includegraphics[width=\hsize]{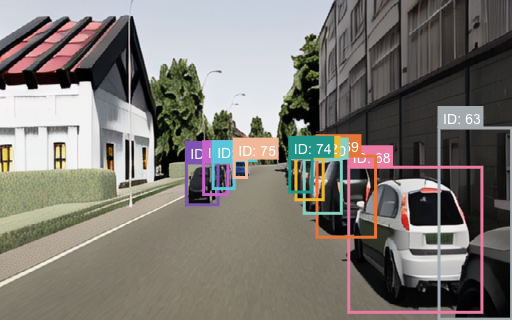}
      & \includegraphics[width=\hsize]{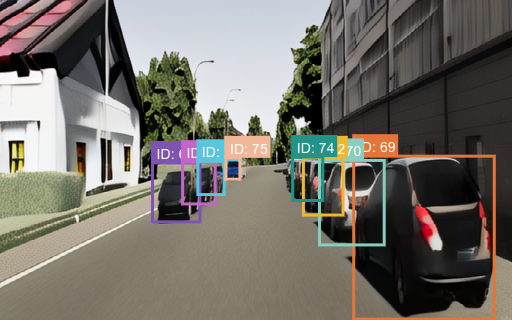}
      & \includegraphics[width=\hsize]{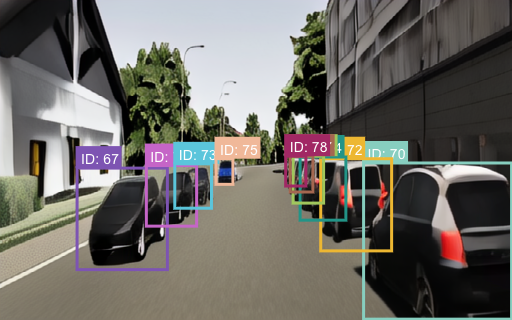}  
      & \includegraphics[width=\hsize]{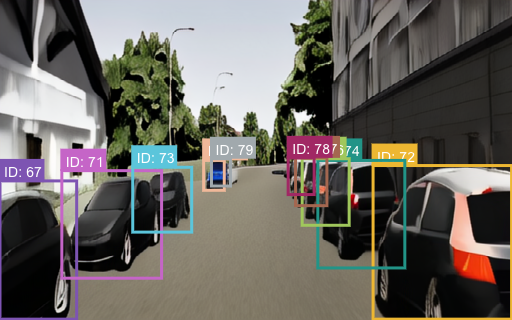}
      & \includegraphics[width=\hsize]{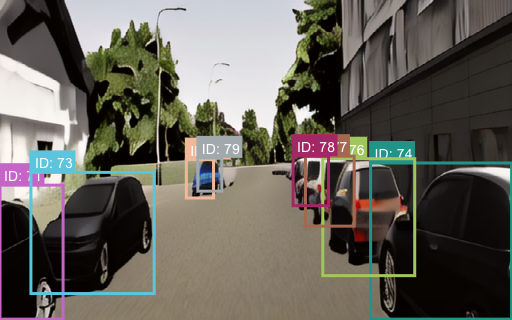}\\ 
      BDD & \includegraphics[width=\hsize]{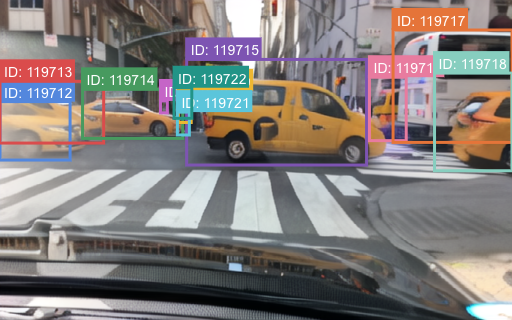}
      & \includegraphics[width=\hsize]{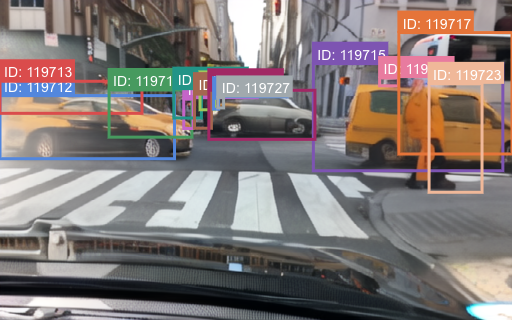}
      & \includegraphics[width=\hsize]{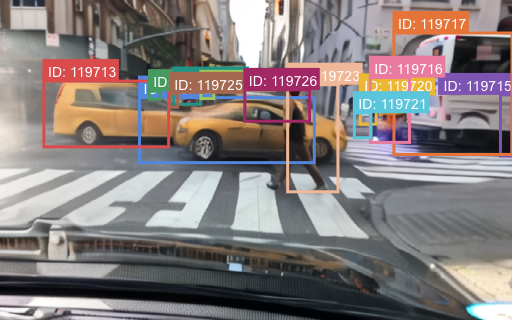}  
      & \includegraphics[width=\hsize]{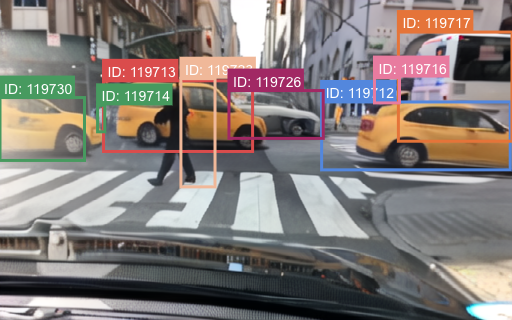}
      & \includegraphics[width=\hsize]{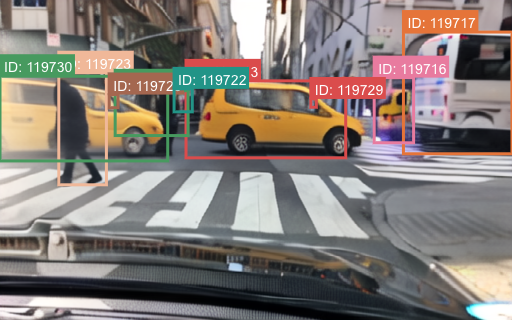}\\ 
      \end{tabular}
    \caption{Illustrations of the \modelvid~generations conditioned on ground truth 3D bounding box trajectories (2D for BDD) across various datasets. The 2D outlines of the ground-truth bounding boxes are overlaid on top.}
    \label{fig:motion_control_demos}%
\end{figure}

\begin{table}[h!]
    \centering
    \resizebox{\textwidth}{!}{\begin{tabular}{m{3cm}lcccccc}
    \toprule
       \textbf{Method} & \textbf{Dataset} &  \textbf{Dataset Type} &  \textbf{\# Frames} &\textbf{mAP}$\uparrow$ & \textbf{AP\(_{50}\)}$\uparrow$ & \textbf{AP\(_{75}\)}$\uparrow$ & \textbf{AP\(_{90}\)}$\uparrow$\\
    \midrule
      \multirow{4}{3cm}{\textbf{Ctrl-V}} & KITTI & Driving & 25 & 0.547 & 0.712 & 0.601 & 0.327\\
       & vKITTI & Driving-sim & 25& 0.599 & 0.776 & 0.667 & 0.356 \\
       & BDD & Driving & 25 & 0.685 & 0.855 &  0.781 & 0.401 \\
       & nuScenes & Driving & 25 & 0.661 & 0.833 & 0.734 & 0.381 \\

    \midrule \midrule
    \multirow{3}{*}{\textbf{ \parbox{3cm}{Boximator\footnote{Boximator only used the last generated frame's bounding boxes to compute average percision scores.} \\ \citep{wang2024boximator}}}} & MSR-VTT\citep{xu2016msr} & Web videos & 16& 0.365 & 0.521 & 0.384 & -\\
    & ActivityNet~\citep{heilbron2015activitynet} & Human-action & 16& 0.394 & 0.607 & 0.409 & -\\
    & UCF-101~\citep{soomro2012ucf}& Human-action & 16& 0.212 & 0.343 & 0.205 & -\\
    \cdashlinelr{1-8}
    \multirow{2}{*}{\textbf{\parbox{3cm}{TrackDiffusion \\ \citep{li2024trackdiffusion}}}} & YTVIS~\citep{Yang2019vis} & YouTube videos & 16& 0.467 & 0.656 & - & -\\
    & UCF-101~\cite{soomro2012ucf} & Human-action & 16& 0.205 & 0.326 & - & -\\
    \bottomrule
    \end{tabular}}
    \caption{Average Precision scores obtained by comparing the YOLOv8 bounding box estimations of real and generated samples. Prior works~\citep{wang2024boximator, li2024trackdiffusion} do not report results on driving datasets; thus, we draw upon their reported performances on alternative datasets to provide a comparative context. 
    The backbone model of Ctrl-V produces videos with 25 frames, while Boximator and TrackDiffusion create videos with 16 frames. Longer videos tend to have reduced quality and lower detection rates, which presents an extra challenge for our model (as it generates 56.25\% more frames);  yet it achieves greater precision compared to the other baseline models.}
    \label{tab:controlnet_evaluation_detailed}
\end{table}

Our \modelvid ~is trained to control object motions through bounding boxes using a teacher-forcing approach, where only ground-truth bounding box frames are provided during the training phase. In this section, we analyze the fidelity of our \modelvid ~generations to the ground-truth bounding box conditions. To access the consistency of objects' locations  between our generated content and ground-truth, we compute the average precision of the bounding boxes in the generated frames and the ground-truth frames. 

Average precision (AP) scores gauge the alignment of predicted/generated bounding boxes with the ground-truth labeling. In all related prior studies, average precision (AP) scores have been consistently reported. However, it is important to acknowledge that AP scores can vary across studies, depending on the specifics of the task setup.
Boximator~\citep{wang2024boximator}'s motion control model predicts object locations in the scene, focusing solely on objects with consistent appearances across all frames. Their AP implementation disregards the object locations in the intermediate frames, comparing the objects' locations only in the final frame. In contrast, TrackDiffusion~\citep{li2024trackdiffusion} uses TrackAP for evaluation, employing a QDTrack model~\citep{fischer2023qdtrack} to track instances in generated videos and comparing them to ground-truth labels. However, these evaluated datasets have limited instances, and TrackAP requires consistent tracking across frames, making it unsuitable for our project without modifications. Therefore, our AP score differs slightly from those in previous works.

Autonomous driving datasets often contain numerous object instances within a scene, with objects continuously entering, exiting, and interacting with each other. In line with this complexity, we have introduced our own version of the AP metric in this work.
Our AP metric is designed to comprehensively compare all objects across every scene: encompassing those that newly enter, those that exist during the intermediate frames, and those that overlap with others.

First, we utilize the state-of-the-art object detection tool, YOLOv8~\citep{reis2024yolov8}, to obtain the objects' trackings from the generated and ground-truth scenes. Detailed information about the tool and our configurations is reported in Appendix~\ref{sec:ms_coco_config}.
Next, we match objects in each generated-vs-ground-truth frame pair based on \emph{spatial similarity} -- calculating the intersection over union (IoU) score to determine the similarity in location between objects' bounding boxes. Our metric disregards object type and tracking IDs equivalence -- assuming that objects close in location should naturally have the same type and IDs.
Finally, we compute the average precision score following MS COCO protocol~\citep{lin2015mscoco}. Details are provided in Appendix~\ref{sec:average_precision} and results are listed in Table~\ref{tab:controlnet_evaluation_detailed}. These results indicate that our \modelvid ~model is particularly adept at adhering to the specified conditions, especially when evaluated with a more lenient metric (i.e., a lower IoU threshold for the AP computation).

%% file: sections/conclusion.tex
We have presented \textbf{Ctrl-V}, a novel model capable of generating controllable autonomous vehicle videos via bounding box trajectory conditioning. Our approach demonstrates that our \textbf{\modelbbox ~} technique can closely follow generation requirements for the first and last frames and produce a coherent bounding box track for intermediate frames. Moreover, our \textbf{\modelvid} network generates high-fidelity videos that strictly conform to the provided bounding boxes. Furthermore, our model accommodates both 2D and 3D bounding boxes and handles uninitialized objects appearing in the middle of the videos. 
Ctrl-V provides future researchers with an efficient way to simulate driving video data with flexible controllability in the form of bounding boxes. In addition, we further define an improved metric to evaluate bounding box conditioned video generation to account for objects that are not present in the first frame, and those that do not remain until the last frame. In Appendix~\ref{sec:future}, we discuss potential future work for this project. With Ctrl-V and an improved metric for more accurate evaluation, we aim to establish a solid foundation for future research in controllable video generation.

%% file: sections/appendix.tex
\newpage
\appendix
\section{Video Demonstrations}
For a more extensive visual presentation of our findings, including generated videos, please follow this link: \url{https://anongreenelephant.github.io/ctrlv.github.io/}.
\section{Notations}\label{sec:svd_notation}
\input{sections/appendix/svd_notations}
\clearpage
\section{Implementation Details}
\input{sections/appendix/implementation_details}
\section{Evaluation Details}
\input{sections/appendix/evaluations}

\clearpage
\section{Generation Results\label{sec:generation_results}}
\input{sections/appendix/svd_vs_ctrlnet}
\input{sections/appendix/generation_results}
\input{sections/appendix/frame-by-frame_control}

\input{sections/appendix/additional_bdd_results}
\clearpage
\section{Trajeglish-Style Bounding Box Generator}\label{sec:trajeglish}
\input{sections/appendix/trajeglish_baseline}

\clearpage

\section{Key Insights and Future Directions\label{sec:future}}
\input{sections/appendix/future_work}

%% file: sections/appendix/svd_notations.tex
\subsection{SVD Scaling Function Definitions}
As presented in Equation~\ref{eq:denoiser}, \(\lambda_{\text{skip}}\), \(\lambda_{\text{out}}\), \(\lambda_{\text{in}}\) and \(\lambda_{\text{noise}}\) are scaling functions where \(\lambda_{\text{in}}\) and \(\lambda_{\text{out}}\) scale the input and output magnitudes of the neural network \(\sU_\theta\) being trained, \(\lambda_{\text{skip}}\) modulates the skip connection, and \(\lambda_{\text{noise}}\) maps noise level \(\sigma_t\) into a conditioning input for \(\sU_\theta\). More detailed information on these functions can be found in the work of \citet{karras2022elucidating}.

Unless otherwise specified, we use the SVD definitions as presented in \citep{blattmann2023stable}. The mathematical formulations for the scaling functions are thus:

\begin{equation}
\begin{split}
    \lambda_{\text{out}}(\sigma)=\frac{-\sigma}{\sqrt{\sigma^2 + 1}}&\quad\quad \lambda_{\text{in}}(\sigma)=\frac{1}{\sqrt{\sigma^2 + 1}}\\
    \lambda_{\text{skip}}(\sigma)=(\sigma^2 + 1)^{-1}&\quad\quad\lambda_{\text{noise}}(\sigma)=\frac{\log\sigma}{4}
\end{split}
\end{equation}

\subsection{Ctrl-V Notations}
In this section, we provide a table of definition for all of the Symbols notations used in our work.
\begin{table}[h]
    \centering
    \begin{tabularx}{\linewidth}{l c X}
    \toprule
      \textbf{Symbol}& \textbf{Appearance} & \textbf{Definition}\\
    \midrule
    \(\En\) & Section \ref{sec:svd_intro}, Figure \ref{fig:model_diagram}& the off-the-shelf VAE Image Encoder used by SVD\\
    \(\De\) & Section \ref{sec:svd_intro}, Figure \ref{fig:model_diagram}& the off-the-shelf VAE Image Decoder used by SVD\\
    \(\mathcal{C}_{\xi}\) & Figure \ref{fig:model_diagram} & modified ControlNet module in \modelvid\\
    \(\sD_\theta\) & Equation \ref{eq:denoiser} & denoiser network of the SVD backbone\\
    \(\sU_\theta\) & Equation \ref{eq:unet_input}, Equation \ref{eq:denoiser} & the main UNet component of \(\sD\)\\
    \(\vf\) & Section \ref{sec:svd_intro} & a sequence of frames\\
    \(\vf^{(i)}\) & Section \ref{sec:svd_intro} & the $i^{th}$ frame in \(\vf\)\\
    \(\vz\)  & Section \ref{sec:svd_intro} & \(\vz = \En(\vf)\). a sequence of frames encoded by the VAE encoder\\
    \(\vz^{(i)}\)  & Section \ref{sec:svd_intro} &  \(\vz^{(i)} = \En(\vf^{(i)})\). the encoded $i^{th}$ frame.\\
    \(\vz^{(0)}\)  & Section \ref{sec:svd_intro}, Equation \ref{eq:unet_input} &  encoded first frame \(\vf^{(0)}\)\\
    \(\vz_{\text{pad}}^{(0)}\) & Equation \ref{eq:unet_input}, Figure \ref{fig:model_diagram} & padded \(\vz^{(0)}\) by repeating itself along the first dimension N times.\\
    $t$ &  Section \ref{sec:svd_intro}, Equation \ref{eq:unet_input} & level used by the EDM noise scheduler do determine the noise level\\
    \(\hat{\vz}_t\) & Section \ref{sec:svd_intro}, Equation \ref{eq:unet_input} &  latent representation of frames corrupted by noise level $t$\\
    \(\vc^{(0)}\)  & Section \ref{sec:svd_intro}, Equation \ref{eq:unet_input} & CLIP encoding of the first frame \(\vf^{(0)}\) \\
    \(\vf_{bbox}\) & Figure \ref{fig:model_diagram} & bounding box frames. Each frame contain pixel renderings of bounding boxes on a blank background. Used as conditioning for \modelvid\\
    \(\vb\) & Figure \ref{fig:model_diagram} & latent representation of \(\vf_{bbox}\) (encoded by \(\En\))\\
    \(\vb^{(0)}\) & Figure \ref{fig:model_diagram} & latent representation of the first bounding box frame\\
    \(\vb^{(N-1)}\) & Figure \ref{fig:model_diagram} & latent representation of the last bounding box frame\\
    \(\hat{\vb}_t\) & Figure \ref{fig:model_diagram} & \(\vb\) corrupted by noise level $t$\\
    \bottomrule
    \end{tabularx}
    \caption{Glossary of terms.\label{tab:all_notations}}
\end{table}

%% file: sections/appendix/implementation_details.tex
\subsection{Step-by-Step Guide to Plotting bounding box Frames~\label{sec:bbox_plot_procedure}}
We create bounding box plots for each frame according to the following steps:
\begin{enumerate}
\setlength{\itemsep}{1pt}
\setlist[enumerate]{leftmargin=0mm}
    \item 
    Each unique track ID is randomly assigned a specific color, and the random color selected has values exceeding 50 across all RGB channels.
    \item 
    Each unique object class label is assigned a distinct color.
    \item 
    For each object, we fill the 2D bounding box region with its track ID's color at 75\% transparency. This transparency allows the overlapping regions of the boxes to remain visible.
    \item 
    If 3D bounding box information is available for an object, we draw the 3D bounding boxes and outline them with the color corresponding to the object's class label.
    \item 
    If 3D bounding box information is unavailable, we outline the 2D bounding box with the color corresponding to the object's class label. Moreover, we mark an X at the rear face of each 3D bounding box for enhanced contextualization.
    \item 
    In certain experiments, we substitute bounding box plots with trajectory plots. To create a trajectory plot, we first compute the mid-points of its 2D bounding boxes. Subsequently, at each bounding box midpoint, we draw a circle with a diameter of 10 pixels, filled with the color corresponding to its track ID. Within this circle, we draw a smaller circle with a diameter of 5 pixels, filled with the color representing the object's class label.
\end{enumerate}

\subsection{bounding box Visualization and Reconstruction Analysis\label{sec:bbox_plots}}
In this work, we refrain from introducing new modules dedicated to encoding the bounding box frames. Instead, we leverage the encoder (\(\En\)) within the SVD's autoencoder framework. Our experiments demonstrate the efficacy of the SVD autoencoder in accurately encoding and decoding bounding box frames.
\begin{figure}[h]%
    \centering
    \subfloat[original frame plot]{{\includegraphics[width=0.45\linewidth]{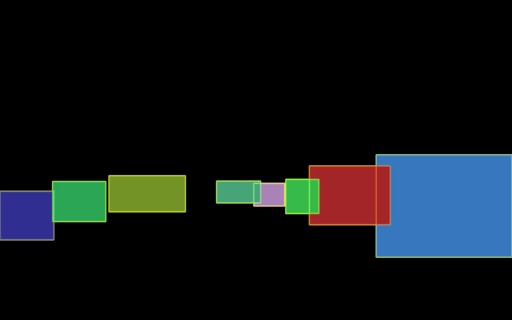} }}%
    \qquad
    \subfloat[reconstructed via SVD's VAE]{{\includegraphics[width=0.45\linewidth]{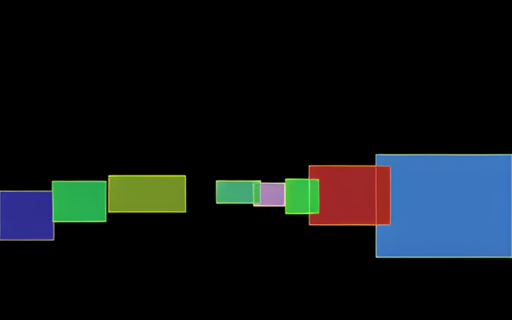} }}%
    \caption{Visualizing a BDD100K 2D-bbox frame created using the method described in \ref{sec:bbox_plot_procedure}}%
    \label{fig:bdd100k_2dbbox_demo}%
\end{figure}
\begin{figure}[h]%
    \centering
    \subfloat[original frame plot]{{\includegraphics[width=0.45\linewidth]{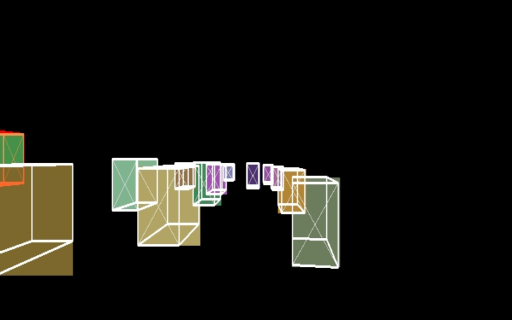} }}%
    \qquad
    \subfloat[reconstructed via SVD's VAE]{{\includegraphics[width=0.45\linewidth]{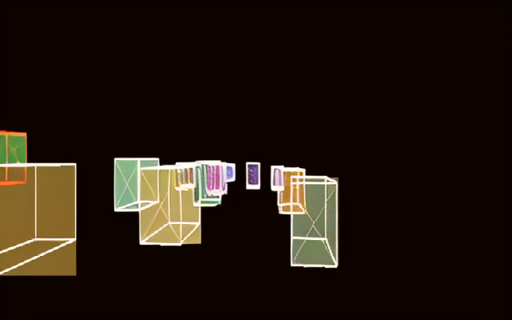} }}%
    \caption{Visualizing a VKITTI 3D-bbox frame created using the method described in \ref{sec:bbox_plot_procedure}}%
    \label{fig:vKITTI_3dbbox_demo}%
\end{figure}

\begin{SCfigure}
    \includegraphics[width=0.45\linewidth]{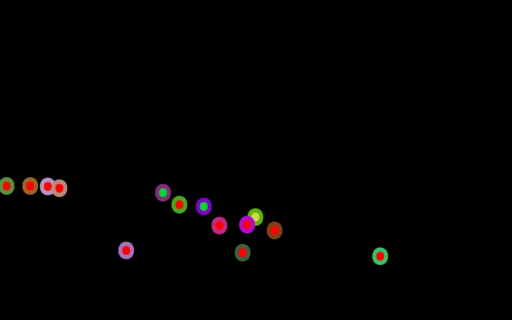}
    \qquad
    \caption{\protect\rule{0ex}{5ex}Visualizing a BDD100K trajectory frame created using the method described in \ref{sec:bbox_plot_procedure}.}\label{fig:example_traj_frame}
\end{SCfigure}

\subsection{Dataset Configuration\label{sec:dataset_configuration}}
\textbf{KITTI:} The downloadable content of the official KITTI dataset does not include bounding box label files for the testing set. As a result, we manually partition the dataset's training folder into two segments for our experiments. The training split encompasses samples `0000' to `0018', while the testing split consists of samples `0019' and `0020'.

\textbf{Virtual-KITTI 2 (vKITTI): }We've not found specification of the train/test split on the official website. During training, we exclude `Scene20', and the models were trained on the remaining four scenes.

\textbf{BDD100k}: We adhere to the official train/validation split for our training and testing phases.

\textbf{nuScenes}: We adhere to the official train/validation split (700/150 scenes) for our training and testing phases.

In our experiments, we use a standardized frame size of \(312 \times 520\) pixels and a clip length of 25 frames at a frame rate of 7 frames per second. All experiments are trained using the training set and assessed using the validation or test sets. Unless otherwise stated, the results and the visualizations provided are derived from the test sets.

\subsection{Training Configuration\label{sec:training_configuration}}

Our pretrained SVD models are adaptations from HuggingFace's diffuser library~\citep{von2022diffusers}. Specifically, we utilize the `stable-video-diffusion-img2vid-xt' SVD variant in this project.

We have trained various types of bounding box generators in this work: \begin{enumerate*}
\item \emph{3D bounding box generator (3-to-1)}
\item \emph{2D bounding box generator (3-to-1)}
\item \emph{2D bounding box-trajectory generator (3-to-1\(^\text{Traj}\))}
\end{enumerate*}.

The 3D bounding box generator anticipates frames using 3D bounding boxes, while the 2D generator does the same for 2D bounding boxes. A 1-to-1 model generates frames between the first and last bounding box frames, while a 3-to-1 model generates frames between the first 3 and last bounding box frames. Additionally, there's a bounding box-trajectory model where the last conditional frame is a trajectory frame instead of a bounding box frame.

For the bounding box generators trained on KITTI or vKITTI datasets, each model undergoes 5 epochs of training. The 2D bounding box generator (1-to-1) is trained on BDD100K for 33,000 iterations. Additionally, the 2D bounding box (3-to-1) and 2D bounding box-trajectory (3-to-1) models start their fine-tuning process for an additional 15,000 iterations from where the 2D bounding box generator (1-to-1)'s training ends. All of these generation models are trained on a single A100-80G GPU with a batch size of 1. We also set gradient accumulation steps to 5 and employed AdamW~\citep{loshchilov2019adamw} as our optimizer.

The fine-tuning and training procedures for the baseline Stable Diffusion and the ControlNet models are almost the same as of the bounding box generator model. The only difference is that we trained our baseline and ControlNet for 48,000 iterations on BDD100K.

\subsection{Model Capacity}
The number of parameters for each model, along with its submodule parameter counts, is provided in Table \ref{tab:num_parameters}.

\begin{table}[]
    \centering
    \begin{tabular}{clcc}
    \toprule
    \textbf{Model}     &  \textbf{Submodule} & \textbf{Status} & \textbf{umber of Parameters}\\
    \midrule
    \multirow{4}{*}{\modelbbox}     &  VAE-Encoder & Frozen & 34,163,592 \\
    & VAE-Decoder & Frozen & 63,579,183\\
    & CLIP-Image Encoder & Frozen & 632,076,800 \\
    & ST Condition UNet & Trainable & 1,524,623,082\\
    \midrule
    \multirow{5}{*}{\modelvid}     &  VAE-Encoder & Frozen & 34,163,592 \\
    & VAE-Decoder & Frozen & 63,579,183\\
    & CLIP-Image Encoder & Frozen & 632,076,800 \\
    & ST Condition UNet & Frozen & 1,524,623,082\\
    & ControlNet & Trainable & 680,946,897\\
    \bottomrule
    \end{tabular}
    \caption{This table presents the parameter counts for the \modelbbox~(2.25 billion) and \modelvid~(2.94 billion) models, with a detailed submodule parameter breakdown.}
    \label{tab:num_parameters}
\end{table}

%% file: sections/appendix/evaluations.tex
\subsection{Generation Quality Metrics: PSNR, SSIM, LPIPS and FVD\label{sec:quality_metrics}}
\textbf{PSNR}, or Peak signal-to-noise ratio, calculates the ratio between the maximum value of a signal and the noise that affects the fidelity of its representation.
It is closely related to Mean Squared Error (MSE) loss. It can be in fact calculated by subtracting the log of MSE from a constant. It is long been used as a metric for image reconstruction quality for its ease of use and mathematical properties. However, PSNR suffers multiple issues as a metric for image generation tasks. Its insensitiveness to various distortion such as blurring, pepper noise, and mean-shifting means that image pairs that look very different to human can have very similar PSNR scores \citep{mse1,psnr1}.

\textbf{SSIM}, or Structural similarity index measure, is a metric that aims to evaluate similarity in "structures" of image pairs -- as opposed to absolute errors as in the cases of MSE and PSNR. Because the human visual system is more sensitive to structural distortions, this metric is more closely aligned with human evaluation compared to PSNR. However, some visually apparent distortions are not captured by the SSIM. Things like changing hue and brightness do not seem to affect SSIM much \citep{ssim1} and \citet{ssim2} presents more examples of unpredictable ssim behaviours when presented with different distortions.

\textbf{LPIPS}, or Learned Perceptual Image Patch Similarity \citep{zhang2018lpips}, is a neural network based approach to accessing image similiarity. It does so by computing distance between some reprensentation of image patches. These representations are obtained by running image patches through some pre-defined neural network. LPIPS has been shown to math human perception well.

\textbf{FVD}, or Frechet Video Distance \citep{unterthiner2019fvd}, is a metric for generative video models. It is based on the Fréchet inception distance (FID) for images. Unlike the 3 other metrics that focuses on image frames and do not consider their temporal relationships, FVD is specifically designed for videos. It takes into consideration the visual quality, temporal coherence, and diversity of samples of videos generated by the model under evaluation. The large scale human study carried out by the authors suggest FVD consistently agrees with human evaluation of videos.

\subsection{Prediction Scores: Converting bounding box Frames to Binary Masks\label{sec:frame2bimask}}
\emph{To convert ground-truth bounding box frames into binary masks,} we combine all color channels and assign a value of 1 to all non-black pixels.

\emph{To transform generated bounding box frames into binary masks,} we first need to perform post-processing on the frames. These frames originate from a network output, and although background pixels that seem black, they may not all be precisely zero. To prevent these pixels being detected as bounding box masks, we aggregate the pixel values across color channels, and for those channels whose sum is below 50, we convert them to black pixels. Subsequently, we convert the frames into binary masks by aggregating all channels and changing the colored pixel values to 1.

The maskIoU score measures the averaged ratio of the intersection area to the union area of the masks. The maskP score evaluates the averaged ratio of the intersection area to the predicted mask, while the maskR score quantifies the averaged ratio of the intersection area to the ground-truth mask. 

Accurately evaluating bounding box generations remains challenging. We recognize that our current metrics (maskIoU, maskR and maskP), which rely on binary masks to evaluate the performance of bounding box generators, have limitations as they do not consider object tracking IDs (the colors of the bounding boxes). However, we still believe that the metric evaluations using binary masks can offer valuable insights into our model's performance.

\subsection{Motion Control Assessment: YOLOv8 Object Detector Configurations\label{sec:ms_coco_config}}
\begin{figure}[h]%
    \centering
    \subfloat[Detection on ground-truth BDD frame.]{{\includegraphics[width=0.32\linewidth]{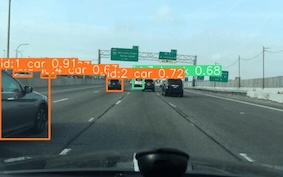} }}%
    \subfloat[Detection on generated BDD frame.]{{\includegraphics[width=0.32\linewidth]{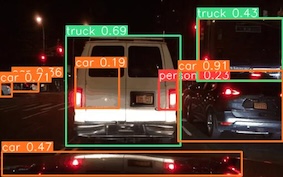} }}%
    \subfloat[Detection on generated vKITTI frame.]{{\includegraphics[width=0.32\linewidth]{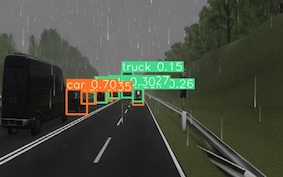} }}%
    \caption{
YOLOv8 detections on BDD and vKITTI with confidence thresholds set to 0.1. The detection confidence scores are labeled on the detected bounding boxes.}%
\label{fig:yolo_detections}
\end{figure}
We utilize the ``yolov8x" variant of the YOLOv8 model, implemented by Ultralytics~\citep{yolo2023ultralytics}, for object detection. We set the detector's Non-Maximum Suppression Intersection Over Union (NMS-IoU) threshold to be 0.35 across all experiments.

Since YOLOv8 was trained to detect objects from the MS COCO dataset~\citep{lin2015mscoco}, we adjust the class labels of our dataset to match the MS COCO labels using the following mappings:
\begin{table}[h]
    \centering
    \begin{tabular}{lp{10cm}}
    \toprule
        \textbf{KITTI}& \textsc{car--car; van--car; truck--truck; pedestrian--person; person--person; cyclist--person; tram--train}\\
        \midrule
        \textbf{vKITTI} & \textsc{car--car; van--car; truck--truck; tram--train}\\
        \midrule
        \textbf{BDD}& \textsc{pedestrian--person; rider--person; car--car; truck--truck; bus-- bus; train--train}\\
        \midrule
        \textbf{nuScenes}& \textsc{human.\{adult, child, construction\_worker, personal\_mobility, police\_officer, wheelchair\}--person; vehicle.\{bicycle, motorcycle\}--person; vehicle.\{bus, construction, ambulance, police, trailer, truck\}--truck; car--car}\\

    \bottomrule
    \end{tabular}
    \caption{Mappings of various dataset labels to MS COCO labels.}
    \label{tab:coco_label_mapping}
\end{table}

The detection resolution is maintained at the same level as the generation/training resolution -- \(312\times 520\).

YOLOv8 is a powerful tool, especially for object detection tasks. Its detection results are very impressive but not always perfect, which leaves rooms for errors when relying on its output to evaluate our model's performance. Two examples of YOLOv8's detection on generated and ground-truth BDD frames are illustrated in Figure~\ref{fig:yolo_detections}.

In the ground-truth BDD frame detections, the network has missed detecting the two black SUVs on the right, even though the confidence threshold is set as low as 0.1 during detection. 

In the generated BDD frame detections, the network has falsely detected the reflections on the engine hood of the driving vehicle as a car. This is a common mislabeling we have observed across the experiments.

In the last example, we demonstrate that YOLOv8 can generate detections on virtual datasets such as vKITTI.

We have attempted to enhance our detection results across frames by using a tracker in conjunction with the YOLOv8 detector. Specifically, we use the BoT-SORT~\citep{aharon2022bot} tracker. However, we've observed an increase in the precision but a significant drop in the recall of the detection scores. 
Thus, we've reverted to using only the YOLOv8 detection model to compute the AP metrics.

\subsection{Motion Control Assessment: Average Precision Calculation~\label{sec:average_precision}}
Average precision (AP) is the area under the recall-precision curve; it takes into account the trade-off between precision and recall at different confidence thresholds. It serves as a measure to evaluate how well the predicted or generated bounding boxes correspond to the actual ground-truth labels.

Prior to developing our custom AP metrics, we conduct experiments to assess YOLOv8's performance on our datasets. We evaluate YOLOv8 and YOLOv8+BoT-SORT on our three datasets using detection and tracking experiments. We employ Track-mAP to evaluate the alignment score between the detections on the ground-truth frames and actual ground-truth labels. The results are not satisfactory; the detection recall is low, suggesting a significant number of missed detections. This issue was exacerbated when using BoT-SORT for tracking. These results lead us to deviate from utilizing the ground-truth labeling and instead use the detections from the ground-truth frames as the labels for our mAP calculations.

We begin by employing a YOLOv8 detection model on both the generated and ground-truth videos, yielding a series of bounding box detections for each. When we generate bounding box detections on ground-truth videos, we keep the confidence cutoff fixed at 0.6. 

 Next, we calculate the AP scores at different IoU thresholds following the MS COCO protocol~\citep{lin2015mscoco}: compute 10 AP scores at IoU thresholds ranging from 0.5 to 0.95, in increments of 0.05. The AP score computed at threshold $\tau$ is referred to as \(\text{AP}_{\tau\times 100}\). Mean average precision (mAP) is the average of all the AP scores. The computed results are reported in Table~\ref{tab:controlnet_evaluation_detailed}.

\begin{figure}[h]%
    \centering
    \subfloat[KITTI]{{\includegraphics[width=0.45\linewidth]{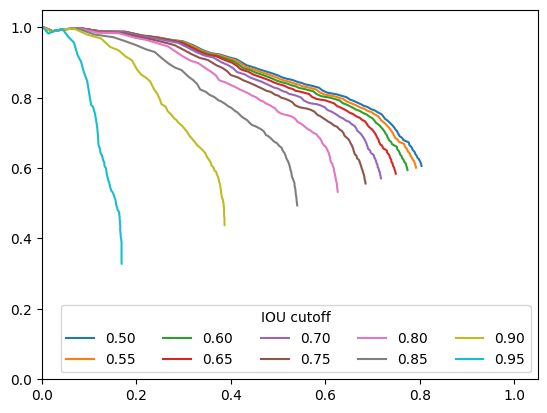} }}%
    \subfloat[vKITTI]{{\includegraphics[width=0.45\linewidth]{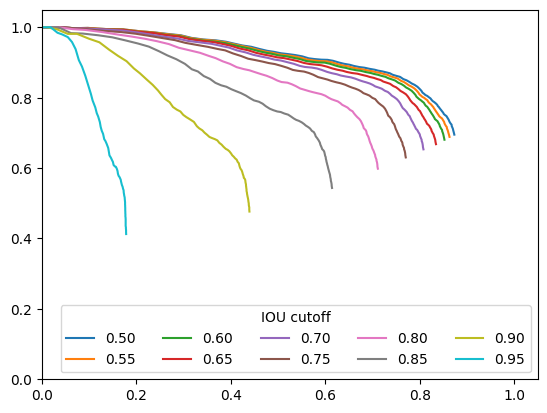} }}%
    
    \subfloat[BDD]{{\includegraphics[width=0.45\linewidth]{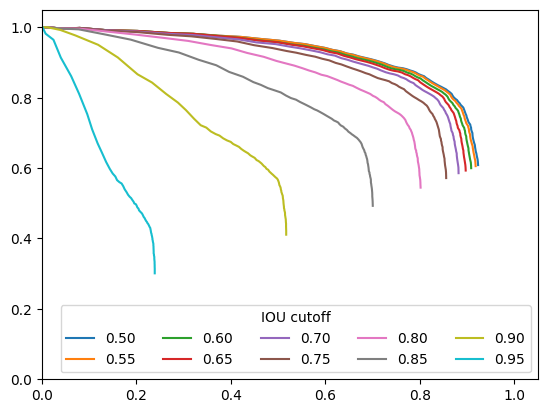} }}%
    \subfloat[nuScenes]{{\includegraphics[width=0.45\linewidth]{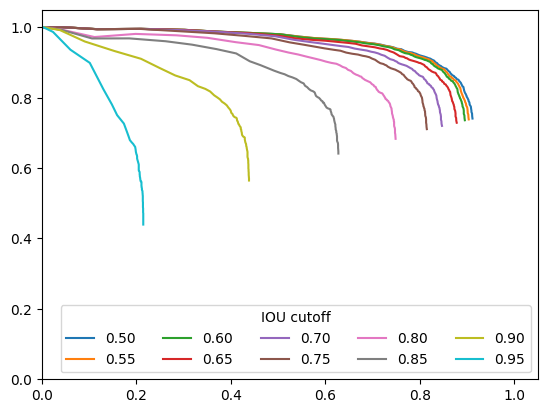} }}%
    \caption{Precision-recall curves: the x-axis represents the recall rate, and the y-axis represents the precision rate.}%
    \label{fig:recall_precision_curves}%
\end{figure}

Figure~\ref{fig:recall_precision_curves} contains the precision-recall curves of the three datasets. The precision-recall curves are computed by dividing confidence cut-off into 100 intervals ranging from 0.01 to 1.00. Each curve presents the precision-recall trade-off by adjusting the confidence cutoff while keeping the IoU cut-off constant.

Our precision-recall curves typically lie above the diagonal line, indicating that our precision is generally higher than our recall. This further implies that if the detector identifies an object in our generated frame, there is a high likelihood that an object is detected in the same location in the ground-truth frame. On the other hand, the lower recall rate implies that an object may not always be present (or detected) in the same location as identified by the detector in the ground-truth frame.

Several factors could have affected the recall rate:

\begin{itemize}
    \item 
    \emph{Generation quality: }The quality of the generated images can influence the detectability of objects, thereby affecting the recall rate.
    \item
    \emph{Detection error: }The performance of the detector has a direct impact on the recall rate. If there are many missed detections, the recall rate will also be high.
    \item 
    \emph{Data error: }This includes mislabeled samples from the dataset and our data preprocessing method. We limit the number of bounding boxes in a frame to a maximum of 15. Consequently, there may be missing bounding boxes for objects in the scene, which could affect our recall rate.
\end{itemize}

On another note, an ideal precision-recall curve would be a horizontal line at \(y=1\). This would indicate perfect precision and recall across all thresholds. Among the 3 experiments, our precision-recall curve for BDD is closest to this ideal line. One reason for this is that the BDD dataset contains significantly more data compared to the others, which can significantly boost our generation results.

%% file: sections/appendix/svd_vs_ctrlnet.tex
\subsection{Stable Video Diffusion vs \modelvid: Generation Quality Comparison\label{sec:gq_compared_svd_vs_ctrl}}

The following are generation comparisons between the fine-tuned Stable Video Diffusion baseline model and the \modelvid~conditioned on the ground-truth bounding box frames. All visualizations are generated using the BDD100k dataset.

\begin{figure}[h]%
    \centering
    \subfloat[Stable Video Diffusion's generation at frame 13]{{\includegraphics[width=0.42\linewidth]{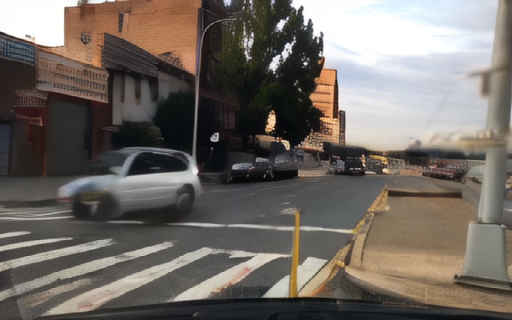} }}%
    \qquad
    \subfloat[\modelvid's generation at frame 13]{{\includegraphics[width=0.42\linewidth]{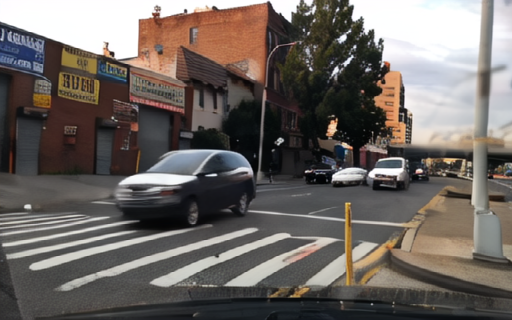} }}%
    \caption{
Comparison of an urban scene generation between the baseline Stable Video Diffusion (SVD) model and the teacher-forced \modelvid~model.}%
    \label{fig:svd_vs_ctrl}%
\end{figure}
\vspace{-.5cm}
\begin{figure}[h]%
    \centering
    \subfloat[Stable Video Diffusion's generation at frame 13]{{\includegraphics[width=0.42\linewidth]{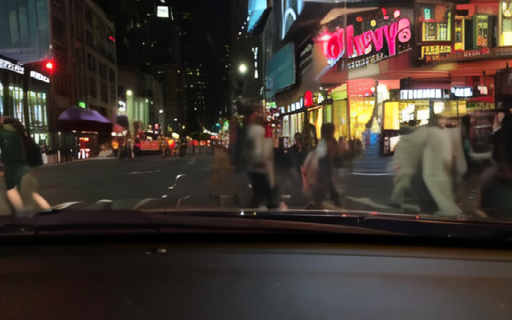} }}%
    \qquad
    \subfloat[\modelvid's generation at frame 13]{{\includegraphics[width=0.42\linewidth]{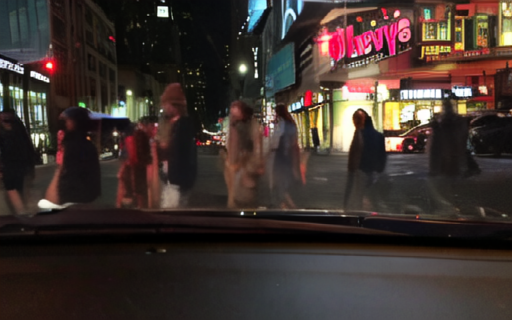} }}%
    \caption{
Comparison of a crowded intersection scene generation between the baseline Stable Video Diffusion (SVD) model and the teacher-forced \modelvid~model.}%
    \label{fig:svd_vs_ctrl_c}%
\end{figure}
\vspace{-.5cm}
\begin{figure}[h]%
    \centering
    \subfloat[Stable Video Diffusion's generation at frame 13]{{\includegraphics[width=0.42\linewidth]{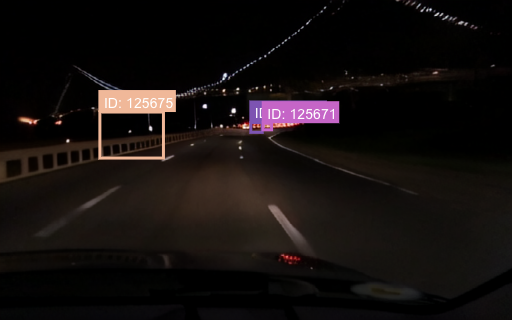} }}%
    \qquad
    \subfloat[\modelvid's generation at frame 13]{{\includegraphics[width=0.42\linewidth]{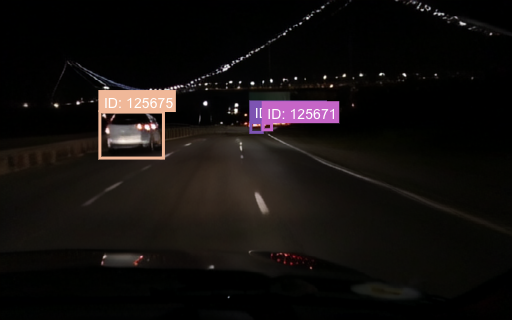} }}%
    \caption{
Comparison of a night scene generation between the baseline Stable Video Diffusion (SVD) model and the teacher-forced \modelvid~model. Ground-truth bboxes are outlined.}%
    \label{fig:svd_vs_ctrl_b}%
\end{figure}

%% file: sections/appendix/generation_results.tex
\clearpage
\subsection{BDD Result Visualizations: bounding box Generations and Motion-Controlled Video Generations}
In the following section, we showcase a range of BDD generation results produced by our 3-to-1 generation pipeline in different scene scenarios, such as city, urban, highways, busy intersections and at night. Each visualization displays every 6th frame from a 25-frame clip, with the actual video generated at a frame rate of 5 fps. In the leftmost column labels, GT represents ground truth, GB represents generated bounding box frames, and GF represents generated frames.
\begin{figure}[ht]
    \setlength\tabcolsep{3pt} 
    \centering
    \begin{tabular}{@{} r M{0.18\linewidth} M{0.18\linewidth} M{0.18\linewidth} M{0.18\linewidth} M{0.18\linewidth} @{}}
    & Frame 1 & Frame 7 & Frame 13 & Frame 19  & Frame 25\\
    GT & \includegraphics[width=\hsize]{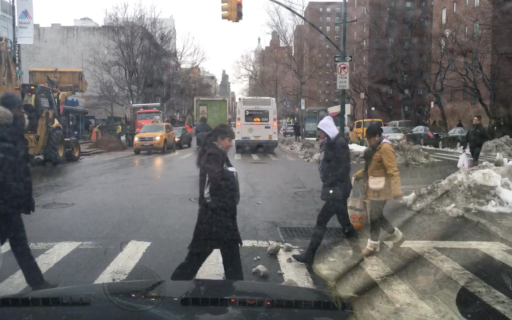}   
      & \includegraphics[width=\hsize]{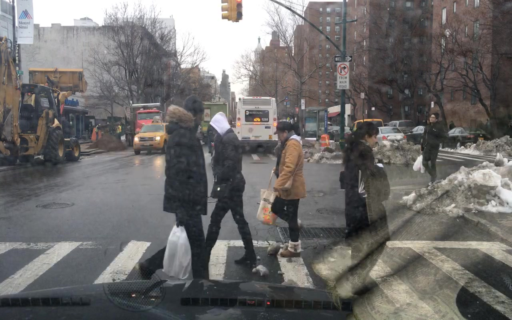} 
      & \includegraphics[width=\hsize]{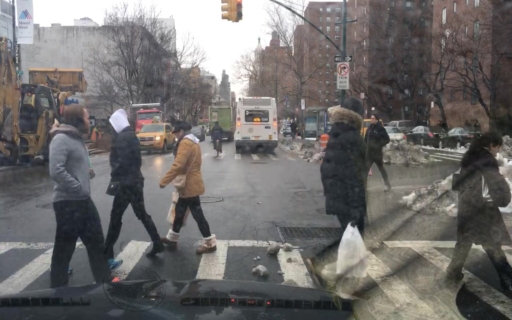}
      & \includegraphics[width=\hsize]{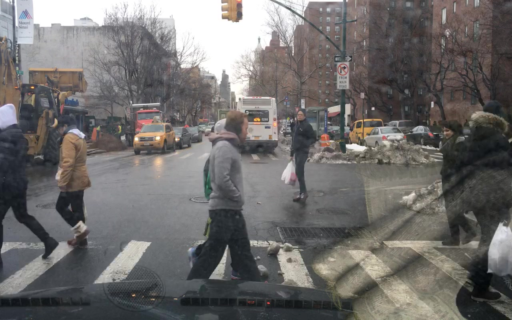}
      & \includegraphics[width=\hsize]{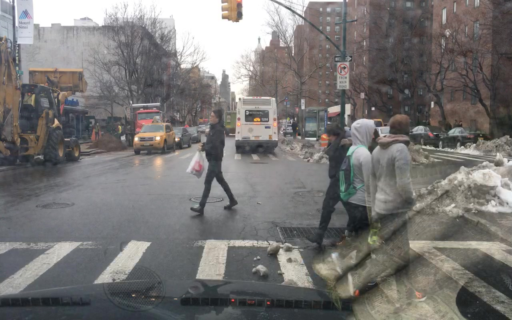} \\
      GF & \includegraphics[width=\hsize]{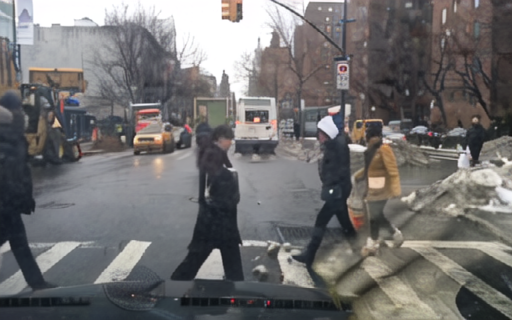}
      & \includegraphics[width=\hsize]{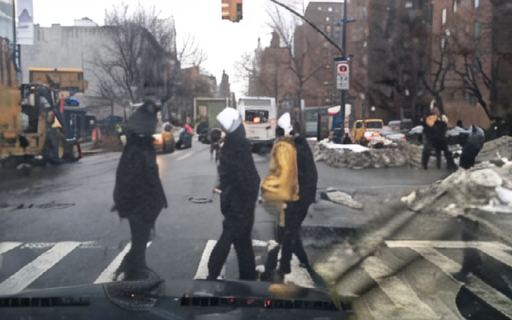}   
      & \includegraphics[width=\hsize]{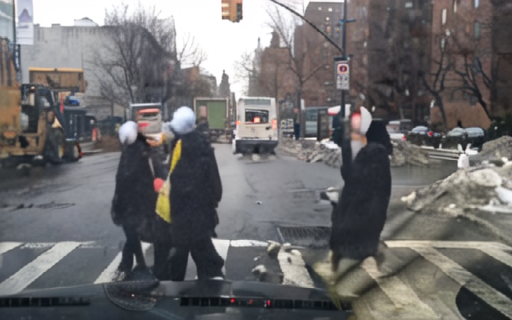}
      & \includegraphics[width=\hsize]{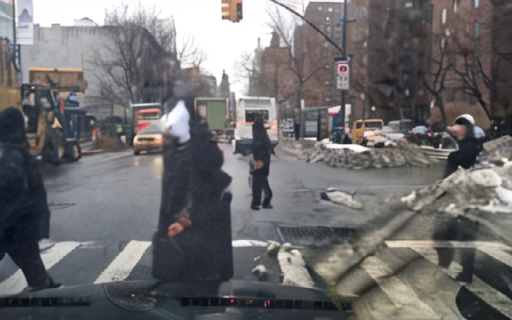} 
      & \includegraphics[width=\hsize]{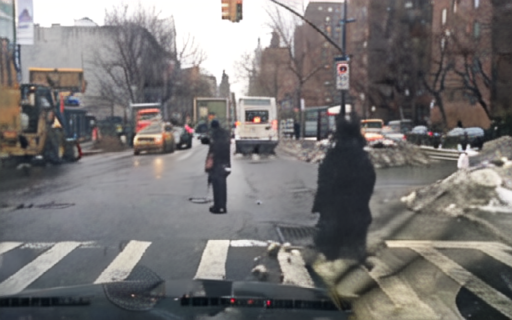}\\
      GB & \includegraphics[width=\hsize]{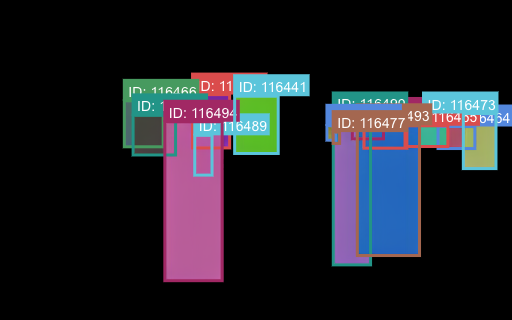}
      & \includegraphics[width=\hsize]{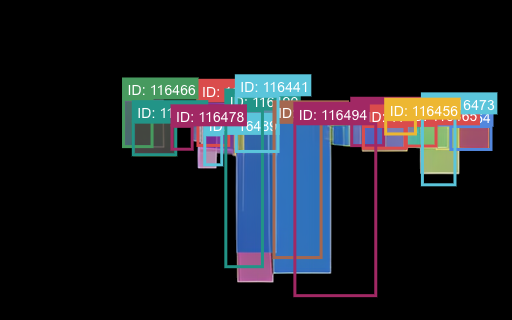}
      & \includegraphics[width=\hsize]{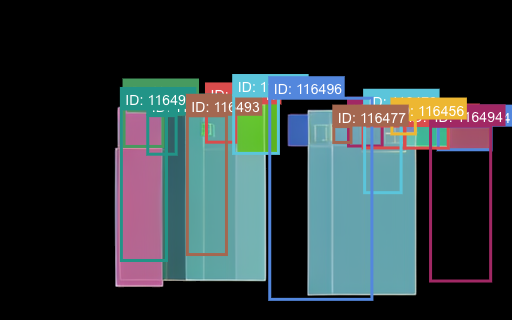}
      & \includegraphics[width=\hsize]{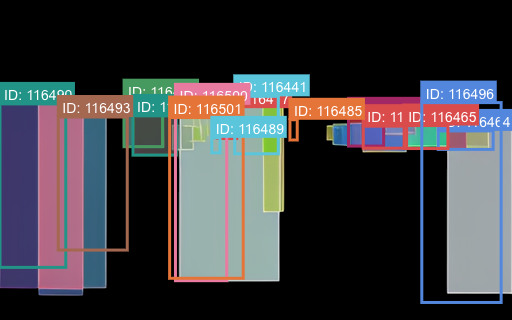}
      & \includegraphics[width=\hsize]{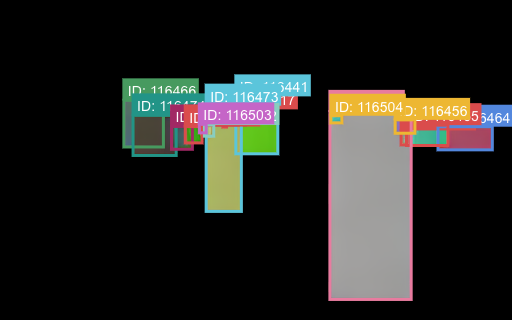}\\ 
      
      \hdashline
      GT & \includegraphics[width=\hsize]{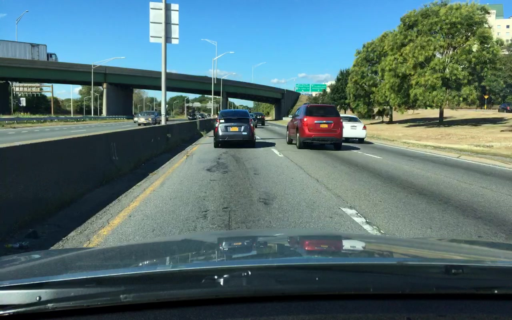}
      & \includegraphics[width=\hsize]{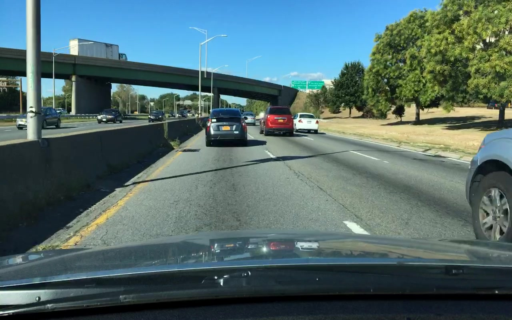}   
      & \includegraphics[width=\hsize]{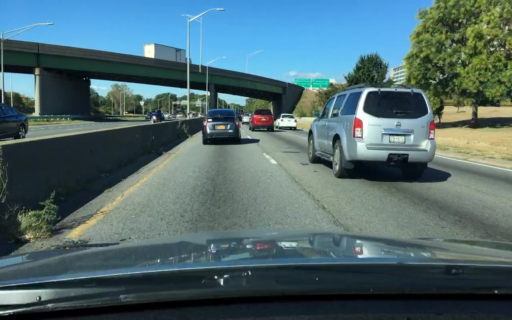}
      & \includegraphics[width=\hsize]{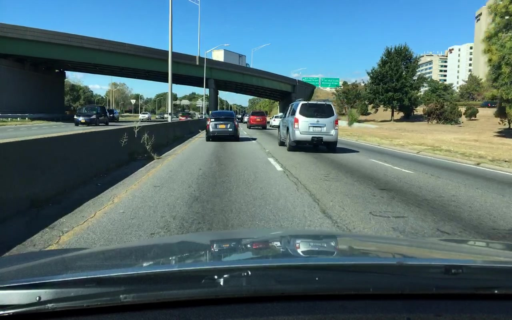}
      & \includegraphics[width=\hsize]{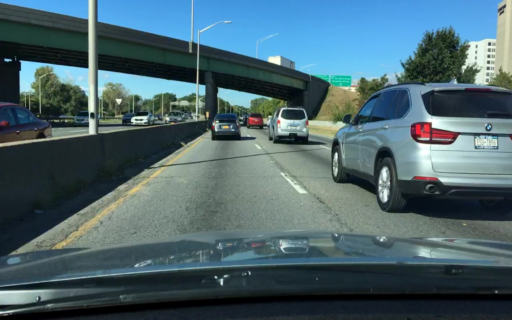} \\
      GF & \includegraphics[width=\hsize]{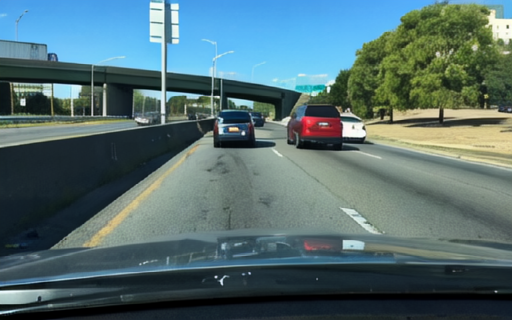}    
      & \includegraphics[width=\hsize]{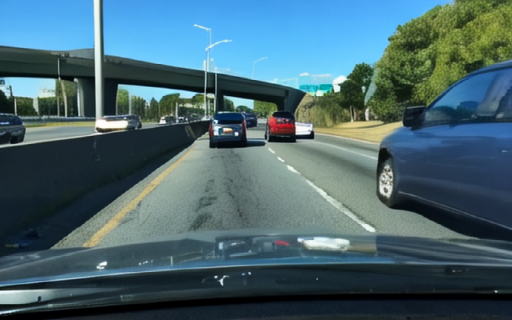}   
      & \includegraphics[width=\hsize]{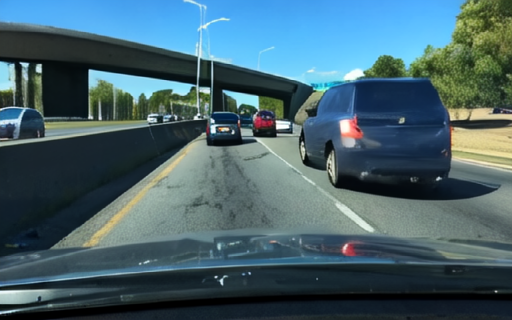}
      & \includegraphics[width=\hsize]{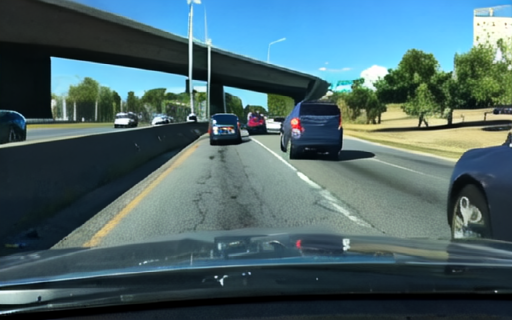}    
      & \includegraphics[width=\hsize]{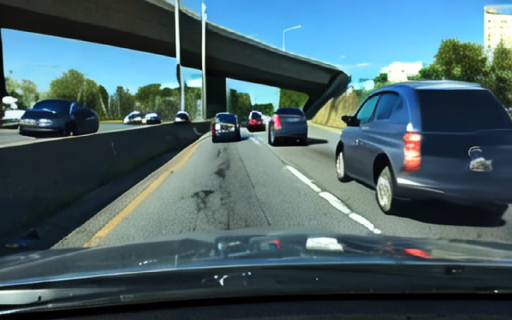}\\
      GB & \includegraphics[width=\hsize]{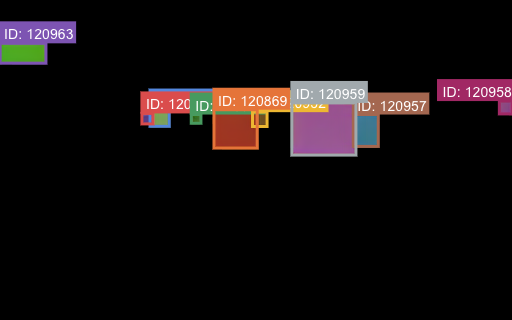}
      & \includegraphics[width=\hsize]{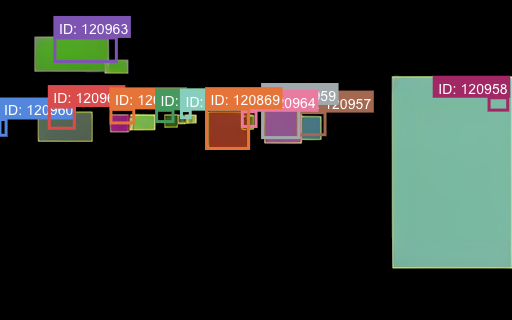}   
      & \includegraphics[width=\hsize]{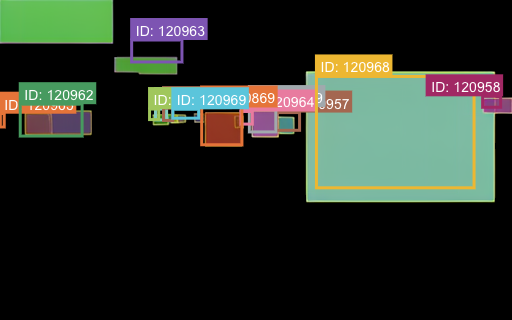}
      & \includegraphics[width=\hsize]{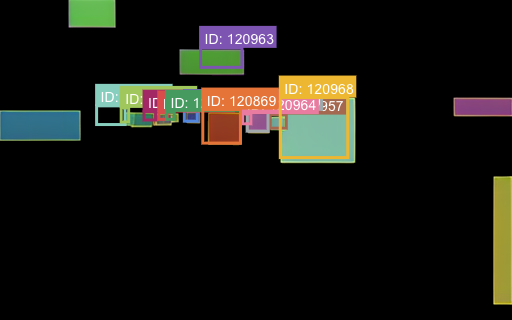}    
      & \includegraphics[width=\hsize]{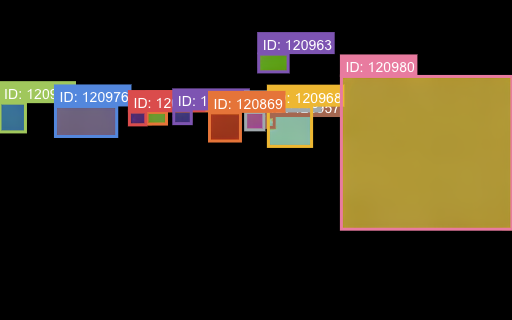}\\
      \hdashline
      GT & \includegraphics[width=\hsize]{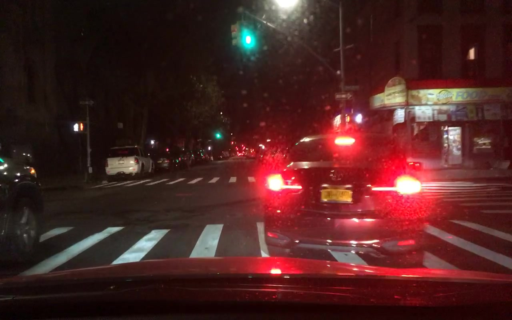} 
      & \includegraphics[width=\hsize]{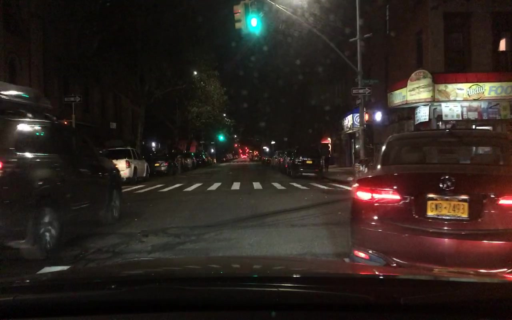} 
      & \includegraphics[width=\hsize]{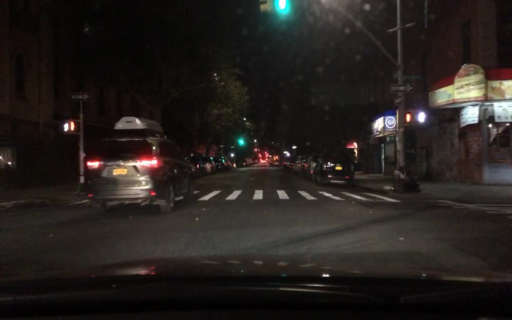}
      & \includegraphics[width=\hsize]{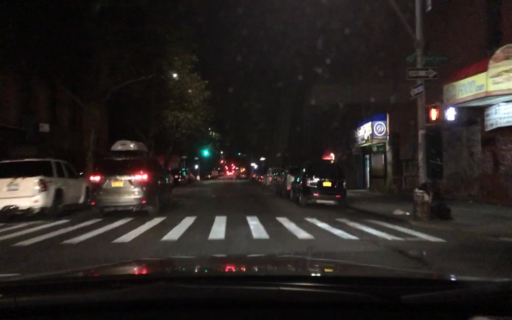}
      & \includegraphics[width=\hsize]{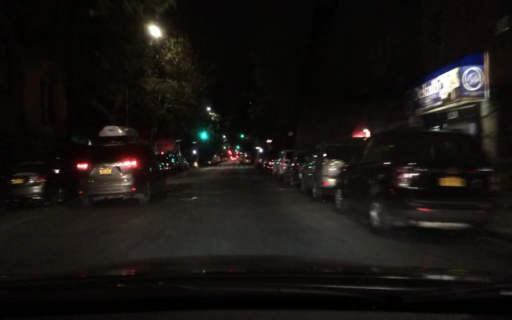} \\
      GF & \includegraphics[width=\hsize]{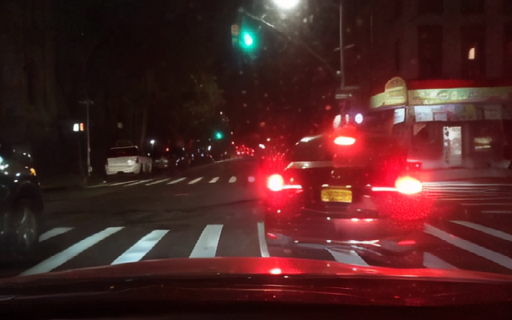}
      & \includegraphics[width=\hsize]{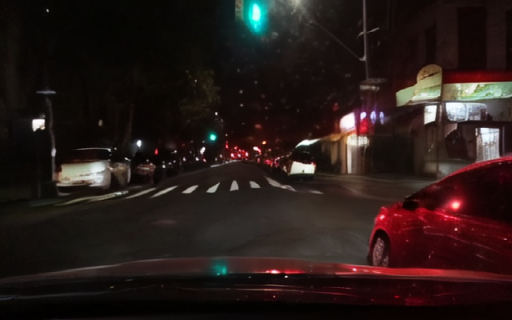}   
      & \includegraphics[width=\hsize]{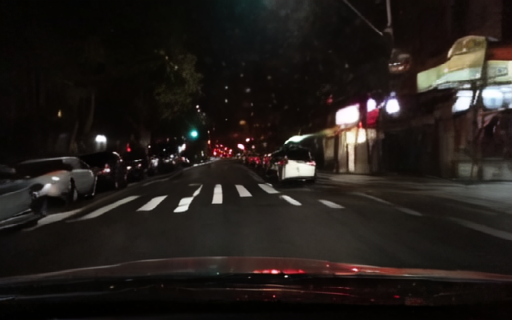}
      & \includegraphics[width=\hsize]{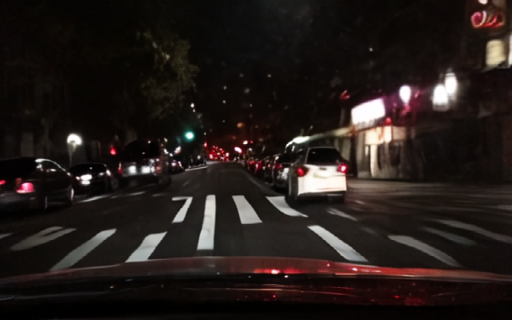}    
      & \includegraphics[width=\hsize]{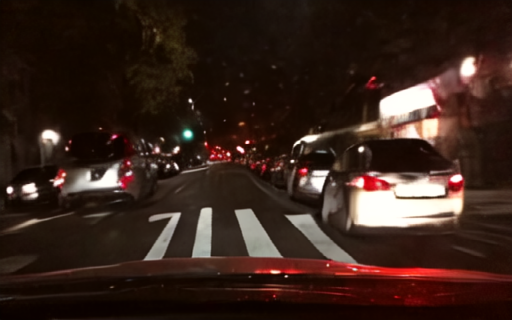}\\
      GB & \includegraphics[width=\hsize]{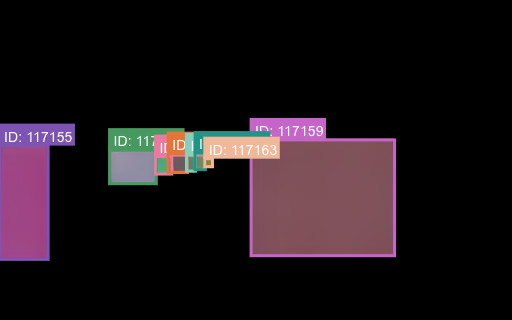}
      & \includegraphics[width=\hsize]{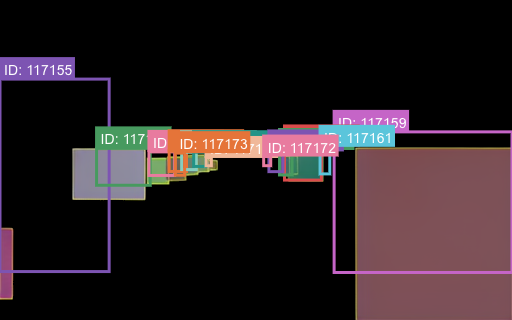}
      & \includegraphics[width=\hsize]{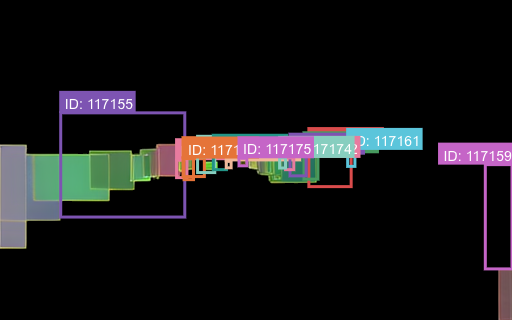}
      & \includegraphics[width=\hsize]{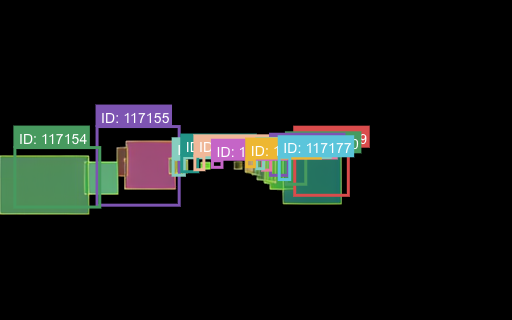}
      & \includegraphics[width=\hsize]{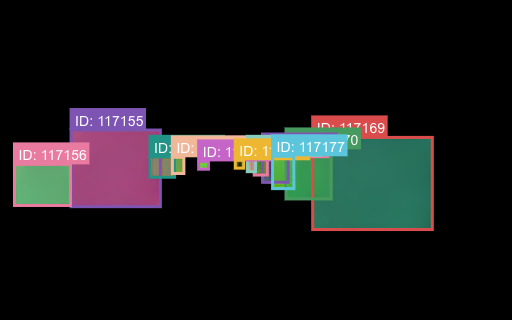}\\ 
      \end{tabular}
      \caption{2D bounding box frame generations and motion-controlled video generations for various scenes.}
      \label{fig:bdd100k_demos_I}
\end{figure}

\clearpage
\subsection{vKITTI Result Visualizations: 3D bounding box Generations and Motion Controlled Video Generations}
\begin{figure}[ht]
    \setlength\tabcolsep{3pt} 
    \centering
    \begin{tabular}{@{} r M{0.17\linewidth} M{0.17\linewidth} M{0.17\linewidth} M{0.17\linewidth} M{0.17\linewidth} @{}}
    & Frame 1 & Frame 7 & Frame 13 & Frame 19  & Frame 25\\
    GT & \includegraphics[width=\hsize]{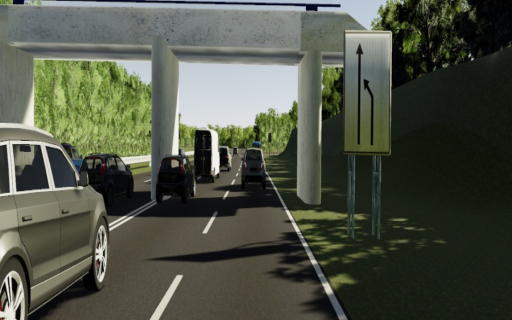}   
      & \includegraphics[width=\hsize]{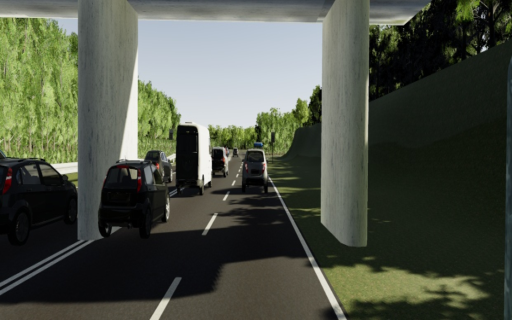} 
      & \includegraphics[width=\hsize]{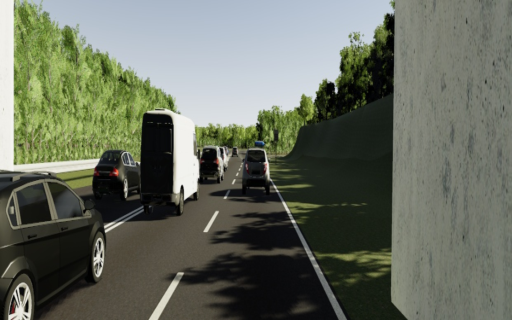}
      & \includegraphics[width=\hsize]{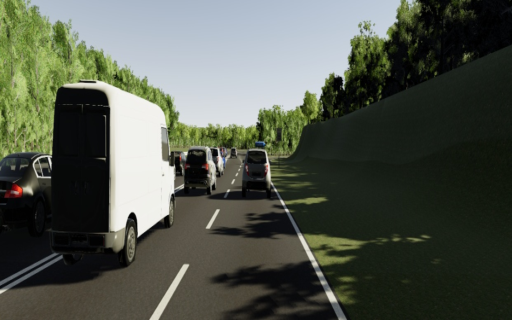}
      & \includegraphics[width=\hsize]{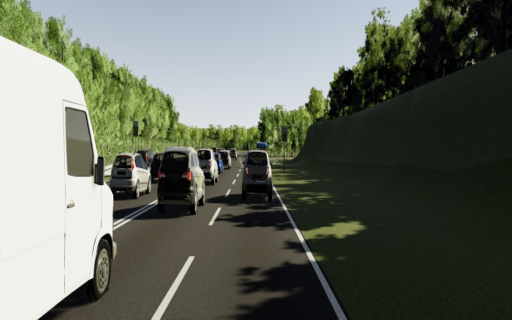} \\ 
      GF & \includegraphics[width=\hsize]{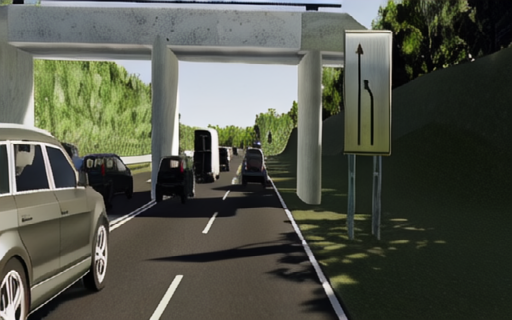}
      & \includegraphics[width=\hsize]{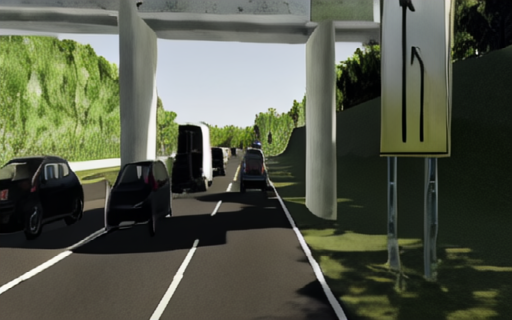}  
      & \includegraphics[width=\hsize]{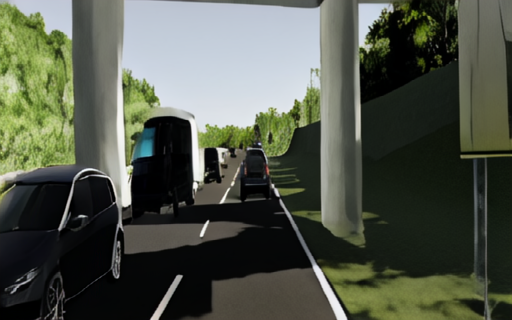}
      & \includegraphics[width=\hsize]{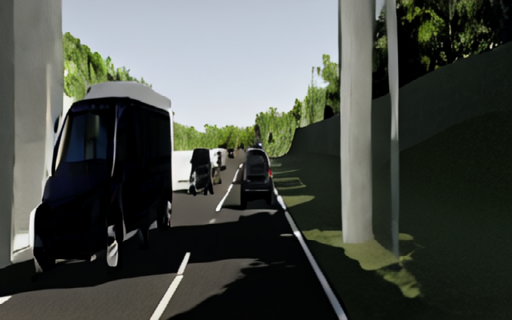} 
      & \includegraphics[width=\hsize]{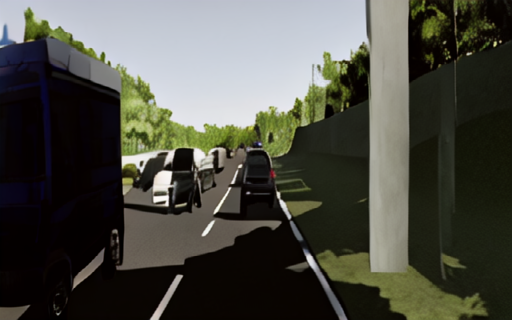}\\
      GB & \includegraphics[width=\hsize]{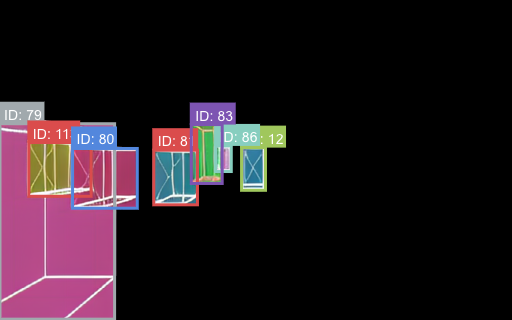}
      & \includegraphics[width=\hsize]{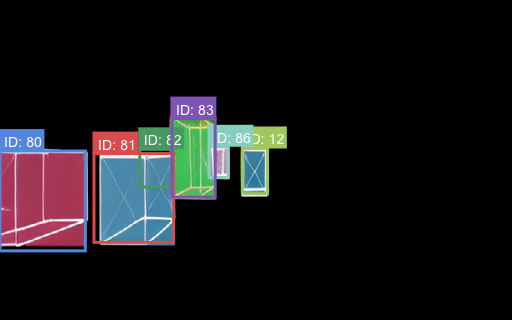}
      & \includegraphics[width=\hsize]{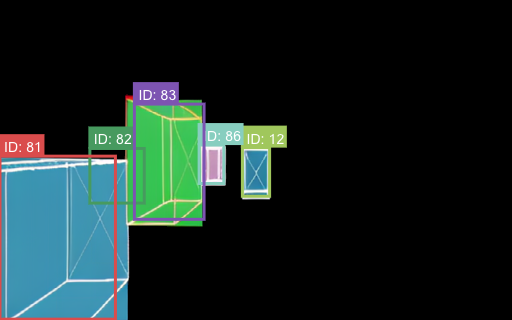}
      & \includegraphics[width=\hsize]{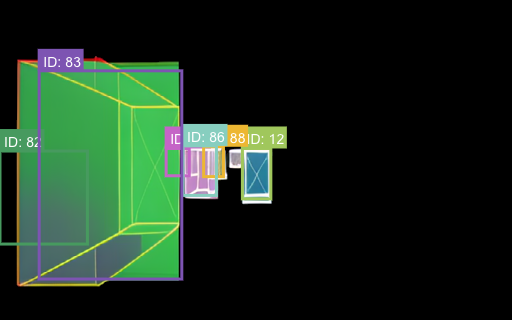}
      & \includegraphics[width=\hsize]{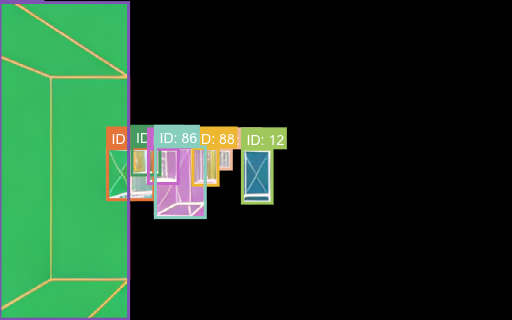}\\ 
    \hdashline
    GT & \includegraphics[width=\hsize]{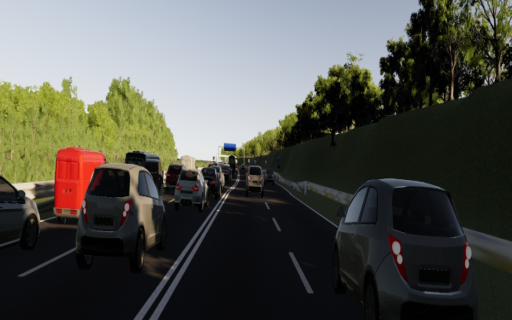}   
      & \includegraphics[width=\hsize]{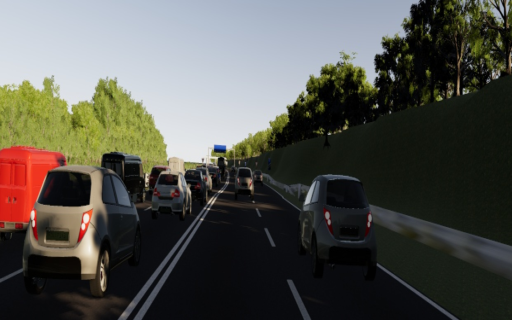} 
      & \includegraphics[width=\hsize]{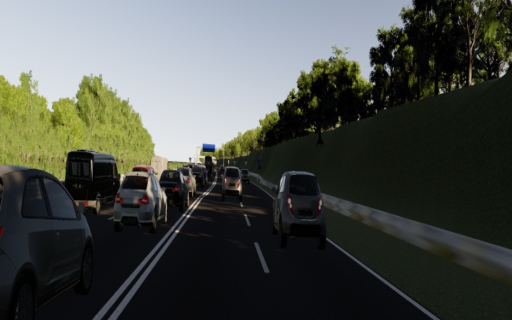}
      & \includegraphics[width=\hsize]{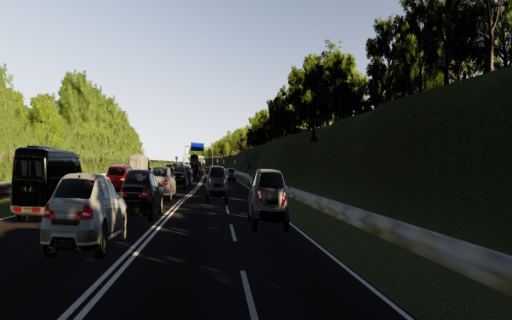}
      & \includegraphics[width=\hsize]{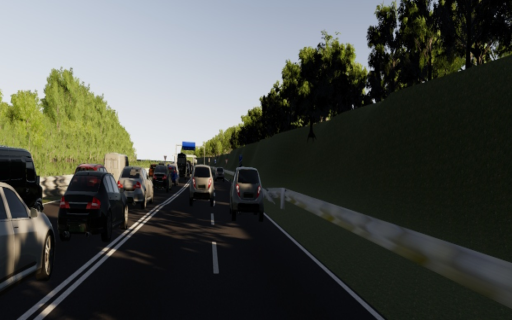} \\
      GF & \includegraphics[width=\hsize]{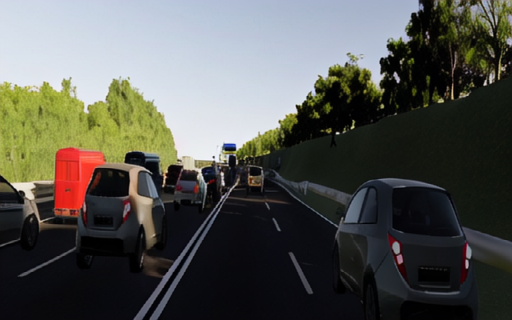}
      & \includegraphics[width=\hsize]{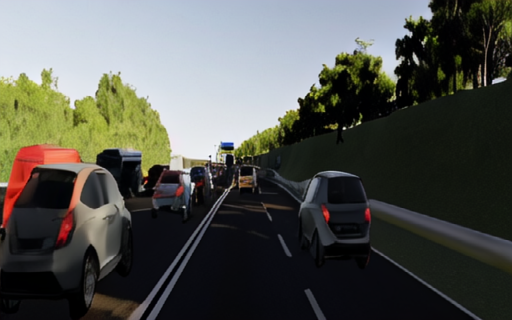}  
      & \includegraphics[width=\hsize]{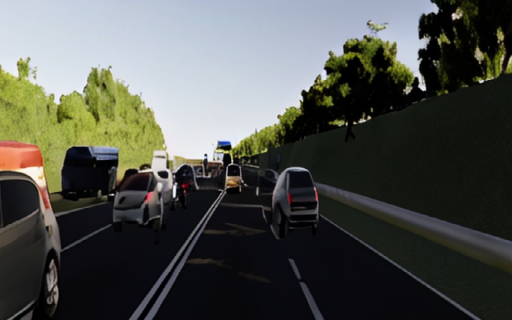}
      & \includegraphics[width=\hsize]{figures/demo_vkitti_daylight/gt_frame_19.png} 
      & \includegraphics[width=\hsize]{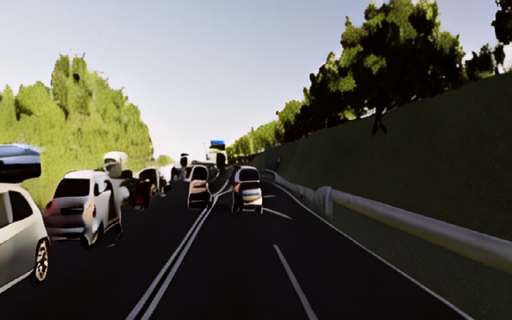}\\
      GB & \includegraphics[width=\hsize]{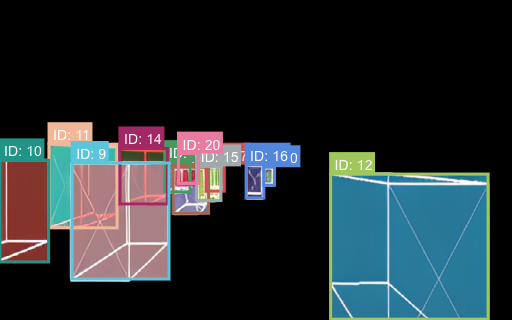}
      & \includegraphics[width=\hsize]{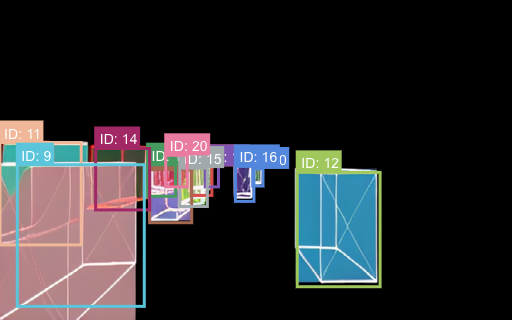}
      & \includegraphics[width=\hsize]{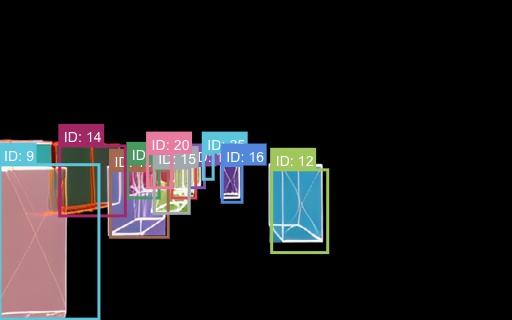}
      & \includegraphics[width=\hsize]{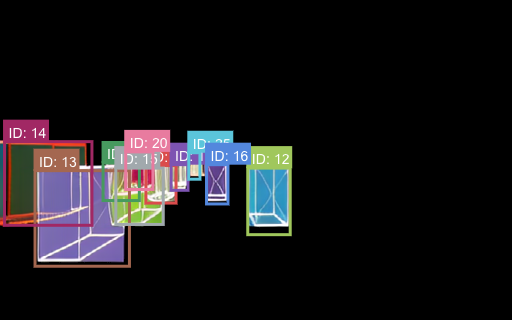}
      & \includegraphics[width=\hsize]{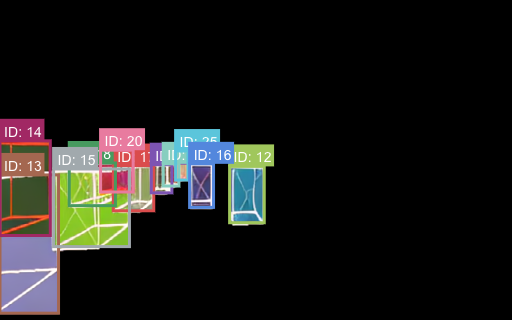}\\ 
    \hdashline
    GT & \includegraphics[width=\hsize]{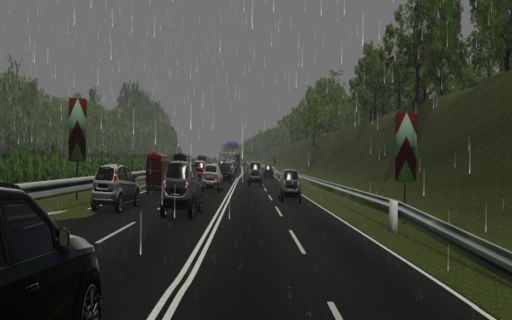}   
      & \includegraphics[width=\hsize]{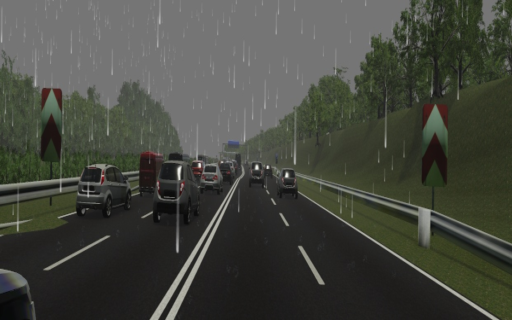} 
      & \includegraphics[width=\hsize]{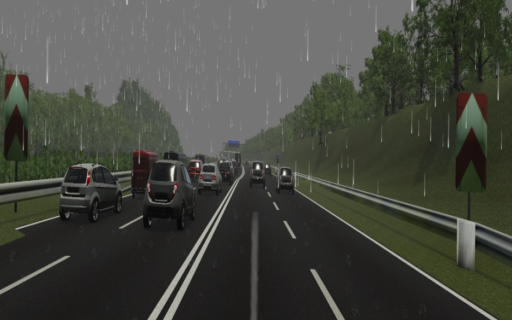}
      & \includegraphics[width=\hsize]{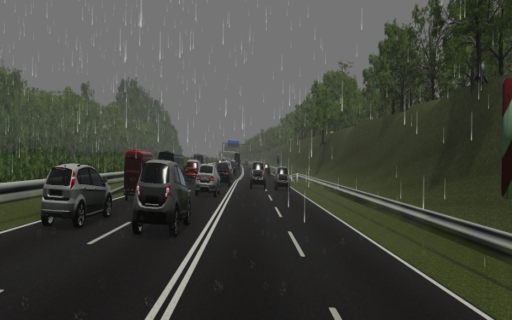}
      & \includegraphics[width=\hsize]{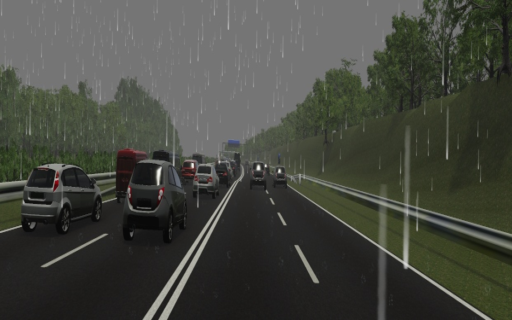} \\ 
      GF & \includegraphics[width=\hsize]{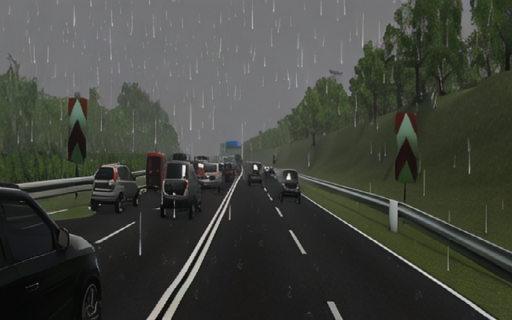}
      & \includegraphics[width=\hsize]{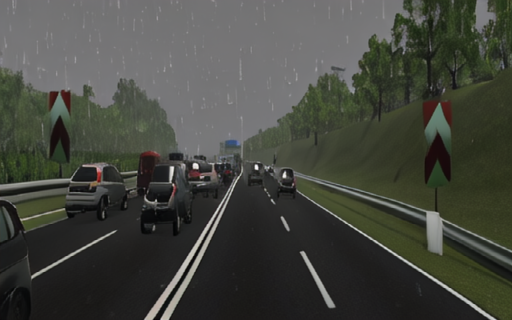}   
      & \includegraphics[width=\hsize]{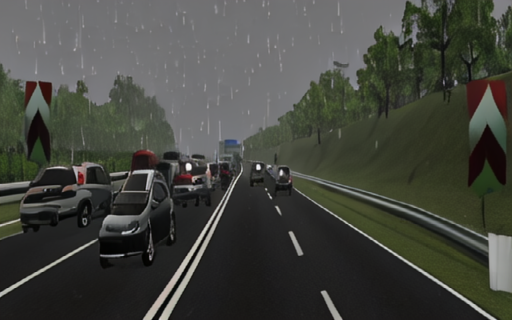}
      & \includegraphics[width=\hsize]{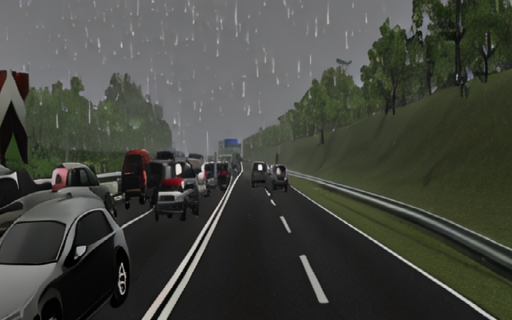} 
      & \includegraphics[width=\hsize]{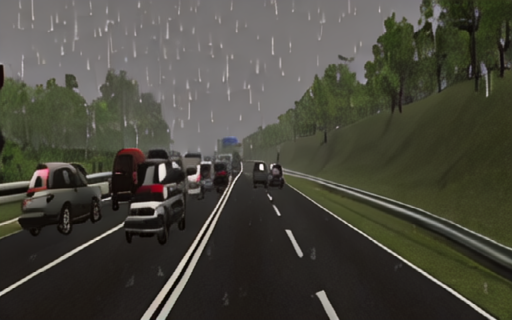}\\
      GB & \includegraphics[width=\hsize]{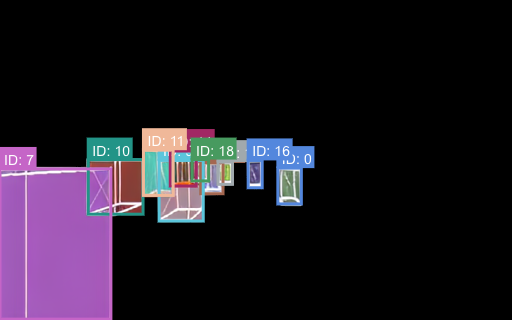}
      & \includegraphics[width=\hsize]{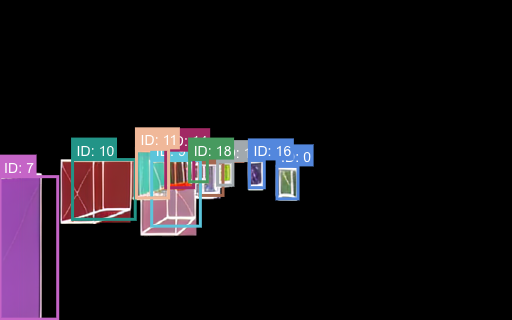}
      & \includegraphics[width=\hsize]{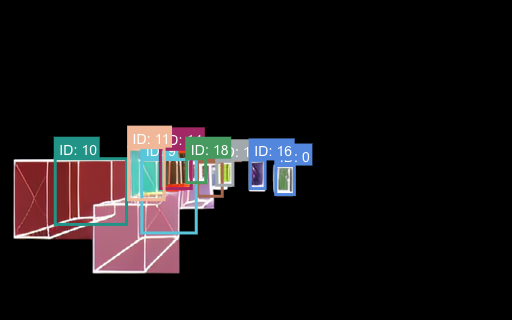}
      & \includegraphics[width=\hsize]{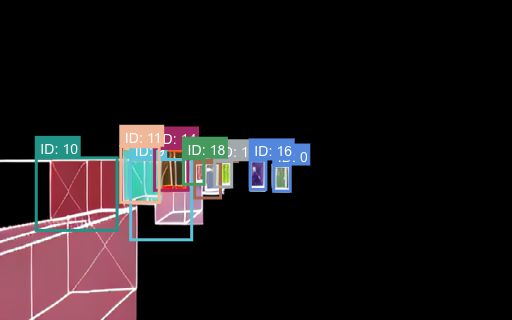}
      & \includegraphics[width=\hsize]{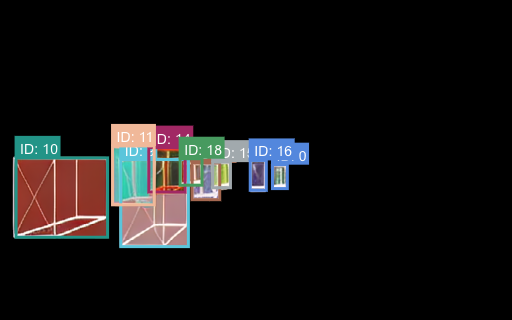}\\ 
    \end{tabular}
    \caption{3D bounding box generations and motion-controlled video generations for various scenes on vKITTI test-split.}
    \label{fig:vkitti_demos_test}
\end{figure}

In the following section, we showcase a range of vKITTI generation results produced by our 3-to-1 generation pipeline in different scene scenarios. In Figure~\ref{fig:vkitti_demos_test}, we showcase our generations on the vKITTI test split. However, it is worth noting that the vKITTI test split comprises samples from only one type of scene (unseen during training).
Each visualization displays every 6th frame from a 25-frame clip, with the actual video generated at a frame rate of 7 fps. In the leftmost column labels, GT represents ground truth, GB represents generated bounding box frames, and GF represents generated frames.

\clearpage
\subsection{KITTI Result Visualizations: bounding box Generations and Motion-Controlled Video Generations}
In the following section, we showcase a range of KITTI generation results produced by our 3-to-1 generation pipeline in different scene scenarios, such as urban area, city streets and highway. Each visualization displays every 6th frame from a 25-frame clip, with the actual video generated at a frame rate of 7 fps. In the leftmost column labels, GT represents ground truth, GB represents generated bounding box frames, and GF represents generated frames.
\begin{figure}[ht]
    \setlength\tabcolsep{3pt} 
    \centering
    \begin{tabular}{@{} r M{0.18\linewidth} M{0.18\linewidth} M{0.18\linewidth} M{0.18\linewidth} M{0.18\linewidth} @{}}
    & Frame 1 & Frame 7 & Frame 13 & Frame 19  & Frame 25\\
    GT & \includegraphics[width=\hsize]{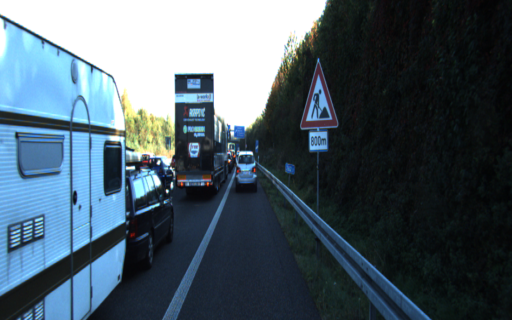}   
      & \includegraphics[width=\hsize]{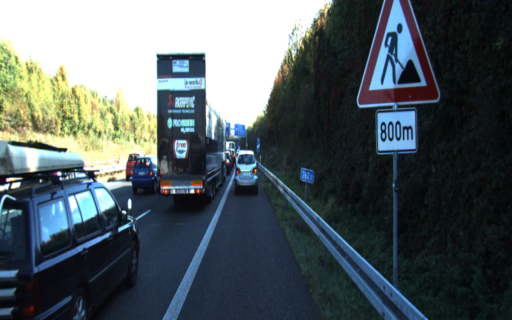} 
      & \includegraphics[width=\hsize]{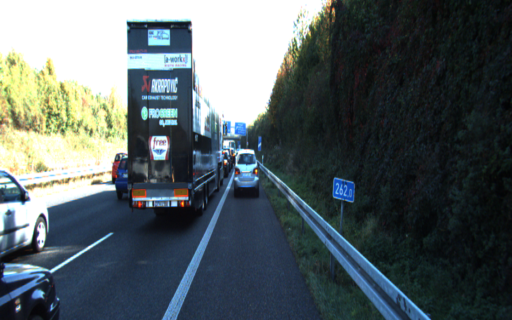}
      & \includegraphics[width=\hsize]{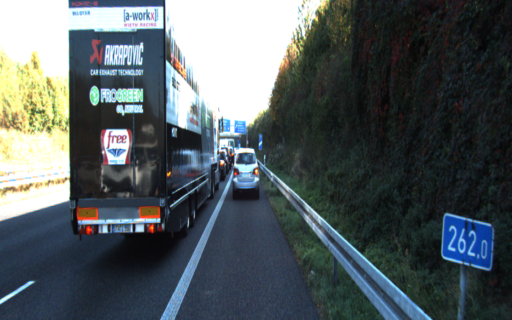}
      & \includegraphics[width=\hsize]{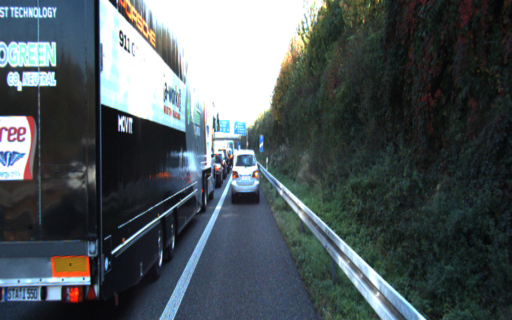} \\
      GF & \includegraphics[width=\hsize]{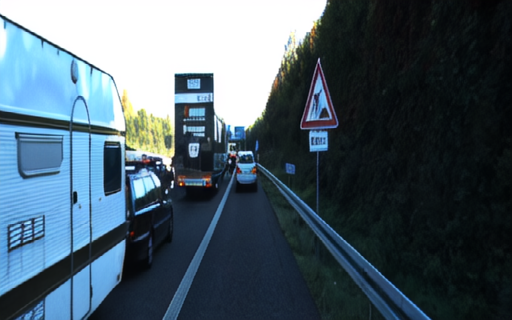}
      & \includegraphics[width=\hsize]{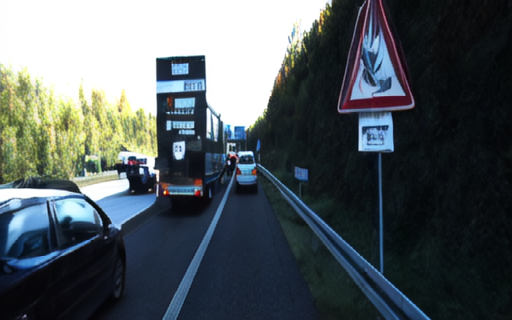}  
      & \includegraphics[width=\hsize]{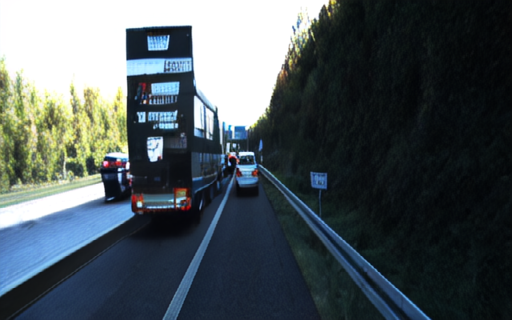}
      & \includegraphics[width=\hsize]{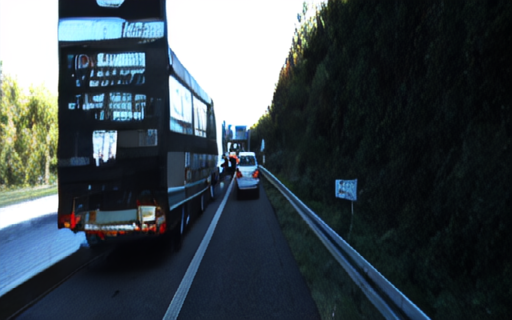} 
      & \includegraphics[width=\hsize]{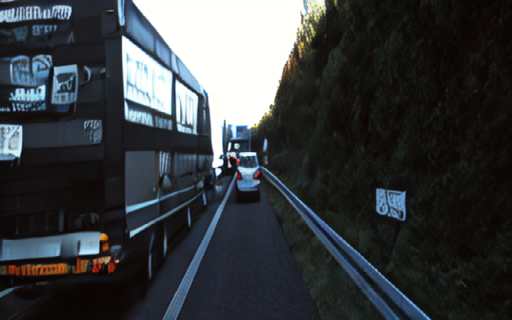}\\
      GB & \includegraphics[width=\hsize]{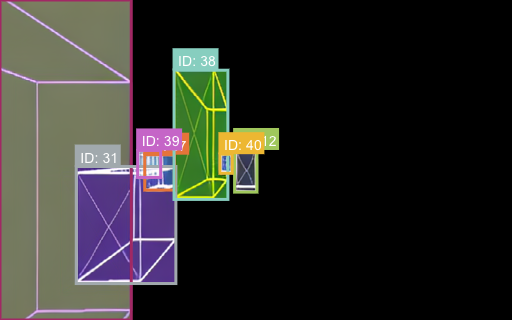}
      & \includegraphics[width=\hsize]{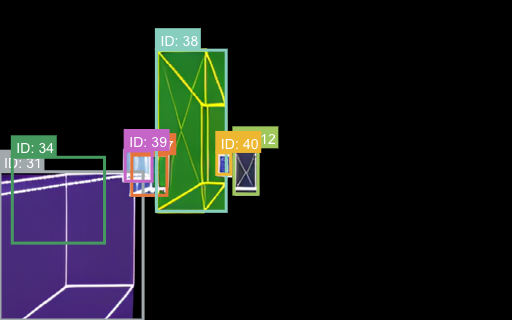}
      & \includegraphics[width=\hsize]{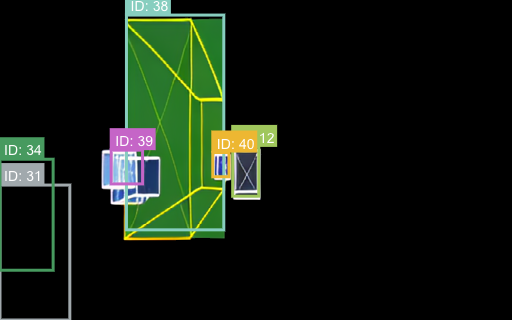}
      & \includegraphics[width=\hsize]{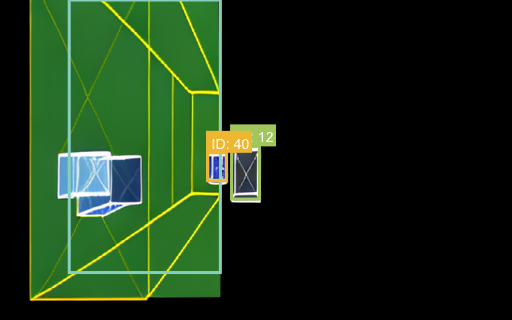}
      & \includegraphics[width=\hsize]{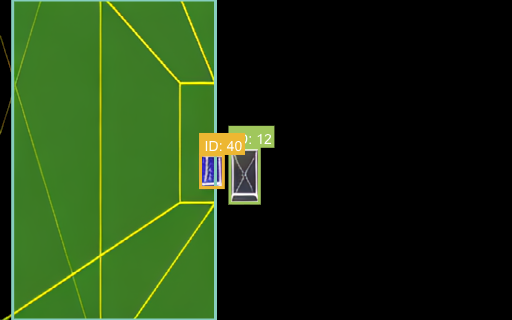}\\ 
    \hdashline
    GT & \includegraphics[width=\hsize]{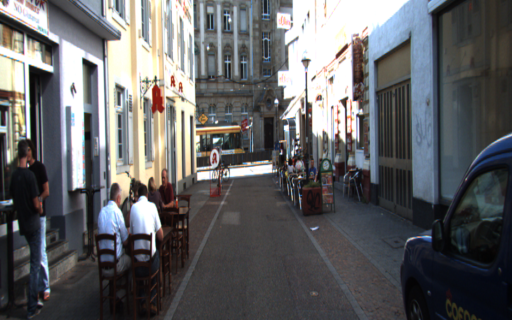}   
      & \includegraphics[width=\hsize]{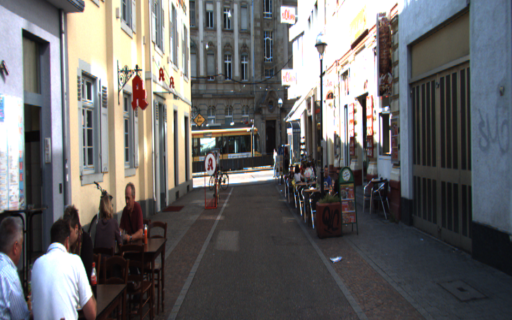} 
      & \includegraphics[width=\hsize]{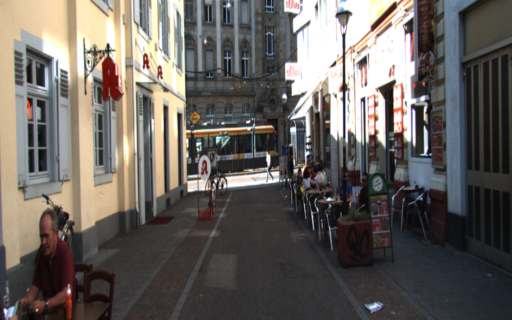}
      & \includegraphics[width=\hsize]{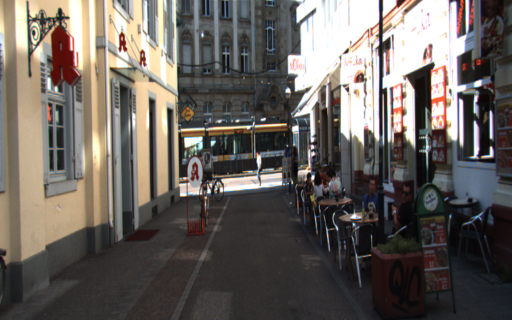}
      & \includegraphics[width=\hsize]{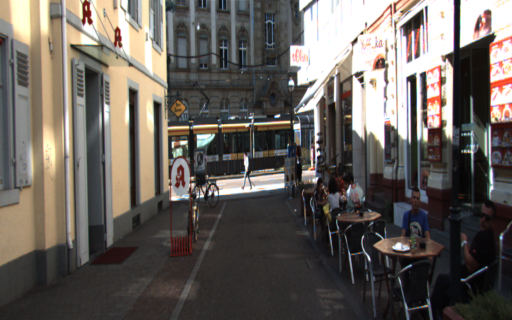} \\
      GF & \includegraphics[width=\hsize]{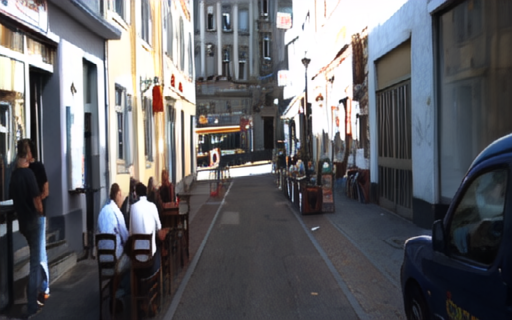}
      & \includegraphics[width=\hsize]{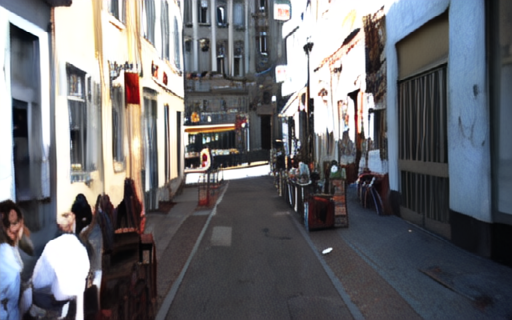}  
      & \includegraphics[width=\hsize]{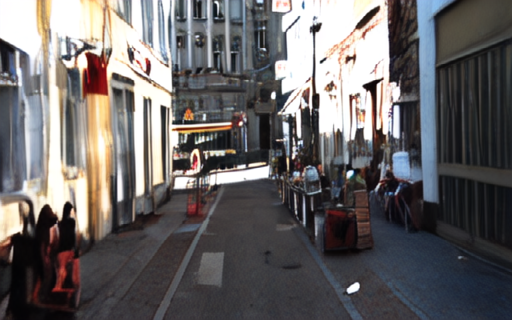}
      & \includegraphics[width=\hsize]{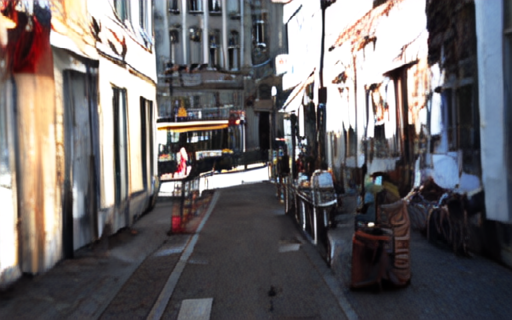} 
      & \includegraphics[width=\hsize]{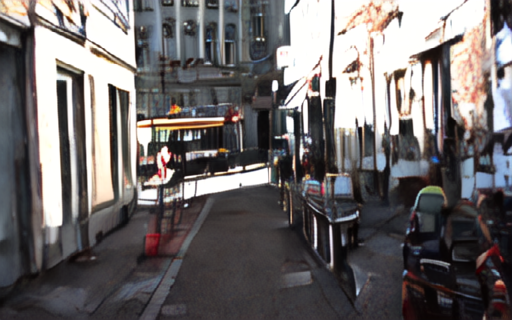}\\
      GB & \includegraphics[width=\hsize]{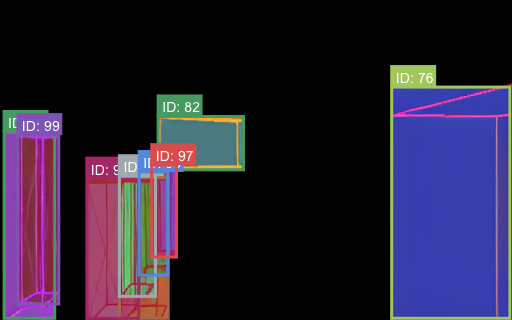}
      & \includegraphics[width=\hsize]{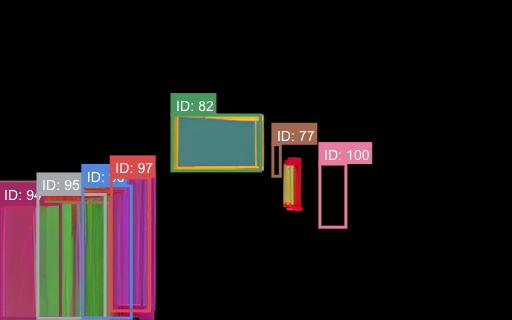}
      & \includegraphics[width=\hsize]{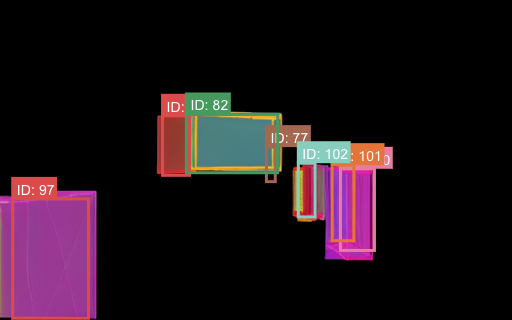}
      & \includegraphics[width=\hsize]{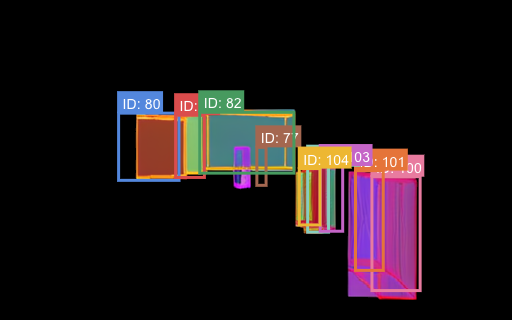}
      & \includegraphics[width=\hsize]{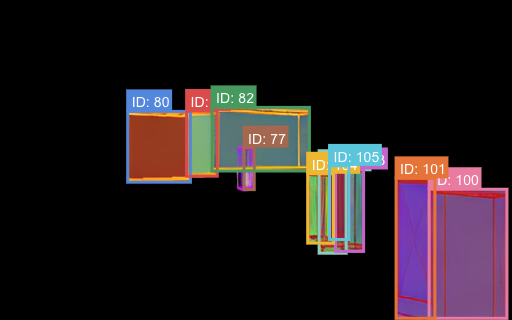}\\ 
      \hdashline
    GT & \includegraphics[width=\hsize]{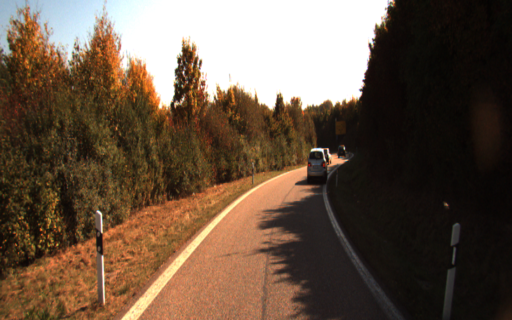}   
      & \includegraphics[width=\hsize]{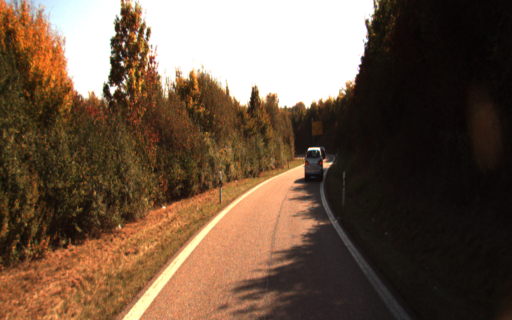} 
      & \includegraphics[width=\hsize]{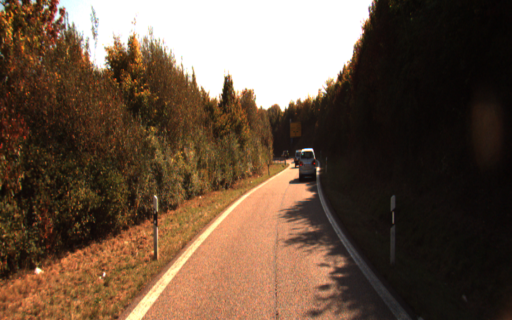}
      & \includegraphics[width=\hsize]{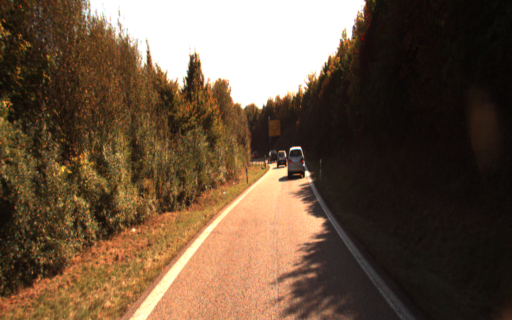}
      & \includegraphics[width=\hsize]{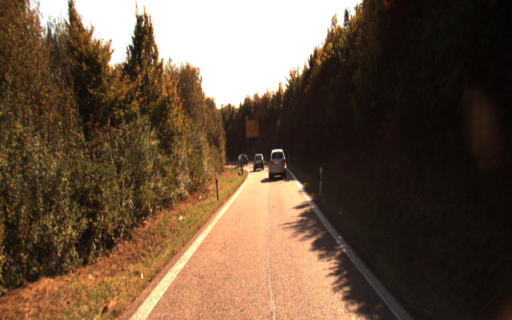} \\
      GF & \includegraphics[width=\hsize]{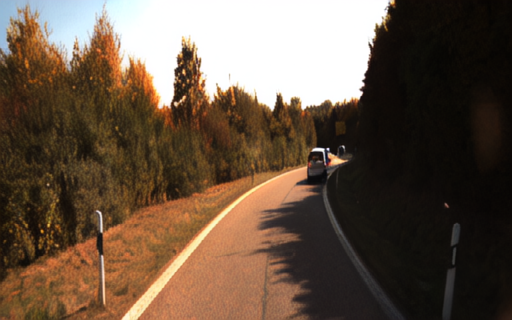}
      & \includegraphics[width=\hsize]{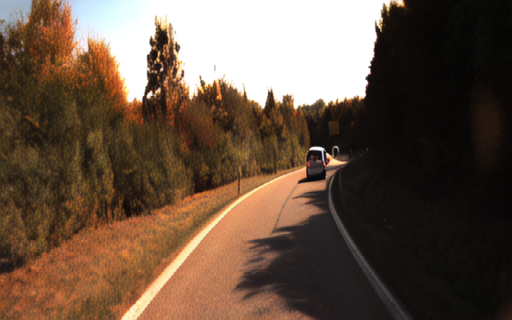}  
      & \includegraphics[width=\hsize]{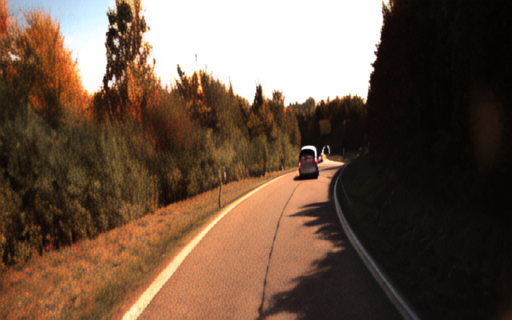}
      & \includegraphics[width=\hsize]{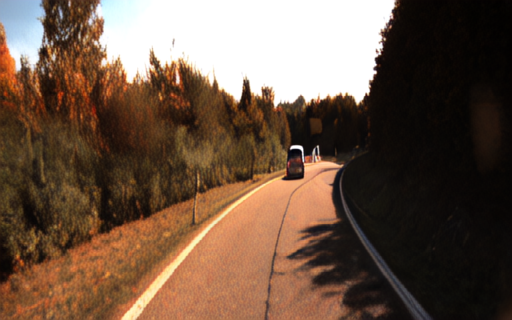} 
      & \includegraphics[width=\hsize]{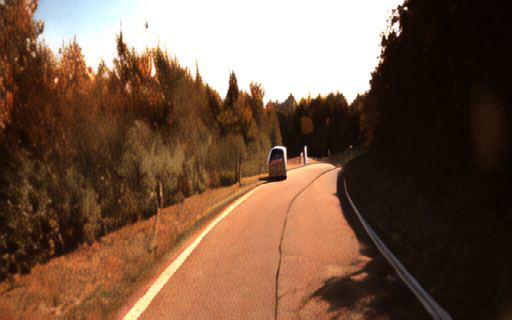}\\
      GB & \includegraphics[width=\hsize]{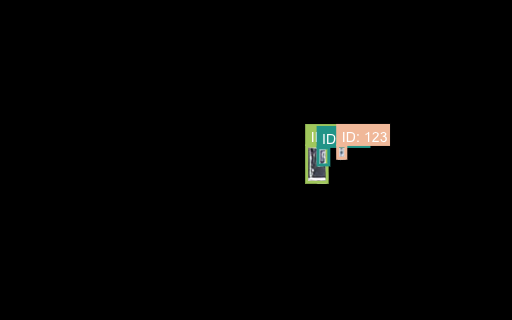}
      & \includegraphics[width=\hsize]{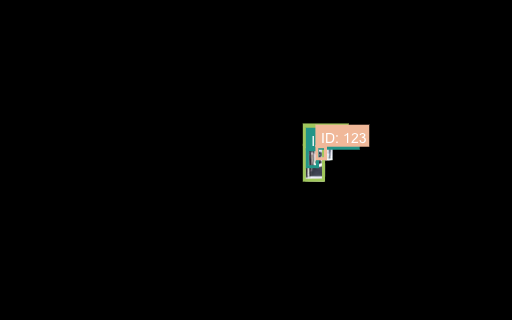}
      & \includegraphics[width=\hsize]{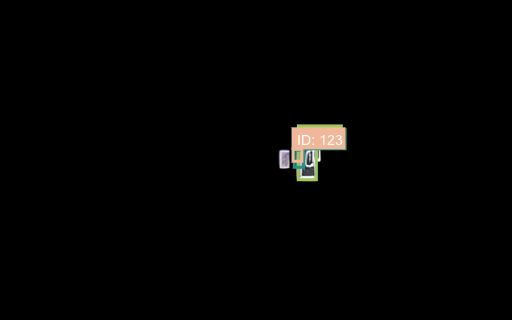}
      & \includegraphics[width=\hsize]{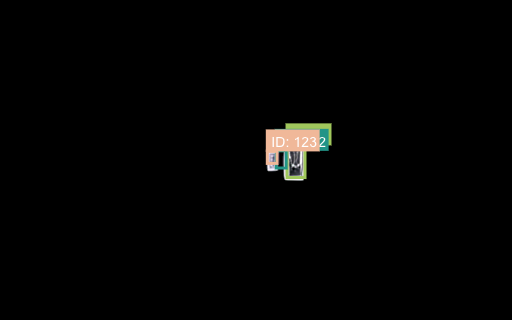}
      & \includegraphics[width=\hsize]{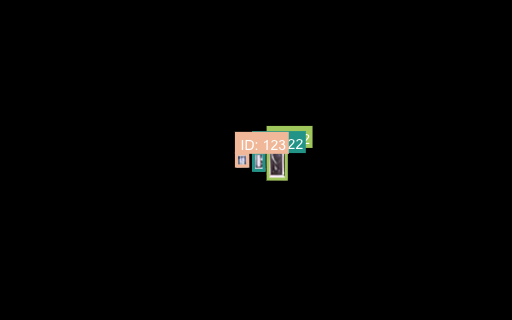}\\ 
\end{tabular}
    \caption{3D bounding box Generations and motion-controlled video generations for various scenes on KITTI test-split.}
    \label{fig:kitti_demos_train}
\end{figure}
\clearpage
\subsection{Nuscenes Result Visualization: bounding box Generations and Motion-Controlled Video Generation}
\begin{figure}[ht]
    \setlength\tabcolsep{3pt} 
    \centering
    \begin{tabular}{@{} r M{0.18\linewidth} M{0.18\linewidth} M{0.18\linewidth} M{0.18\linewidth} M{0.18\linewidth} @{}}
    & Frame 1 & Frame 7 & Frame 13 & Frame 19  & Frame 25\\
      GT & \includegraphics[width=\hsize]{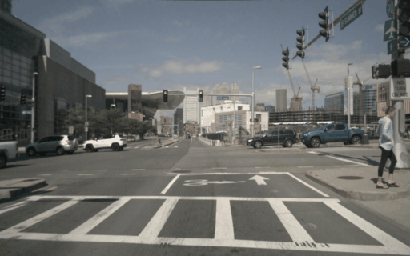}   
      & \includegraphics[width=\hsize]{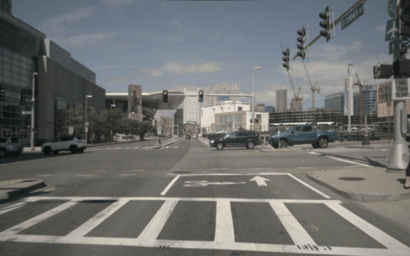} 
      & \includegraphics[width=\hsize]{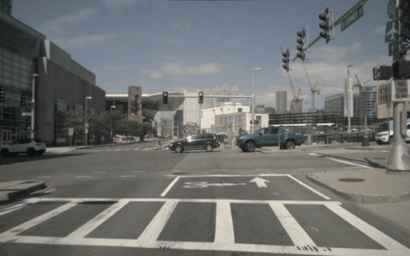}
      & \includegraphics[width=\hsize]{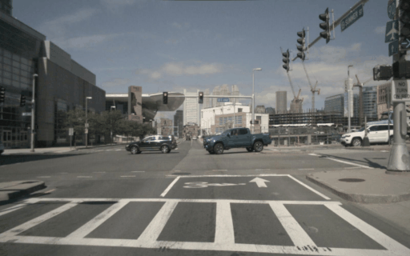}
      & \includegraphics[width=\hsize]{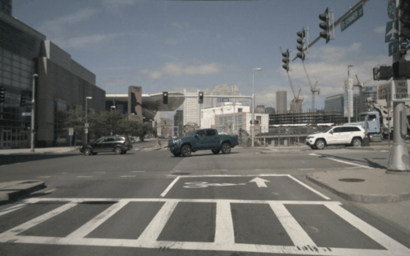} \\
      GF & \includegraphics[width=\hsize]{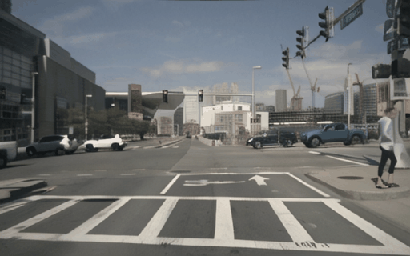}
      & \includegraphics[width=\hsize]{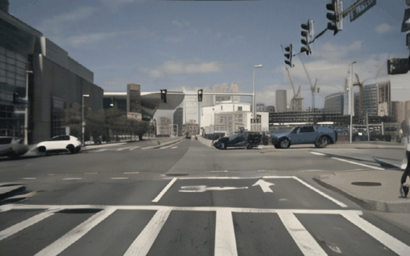}  
      & \includegraphics[width=\hsize]{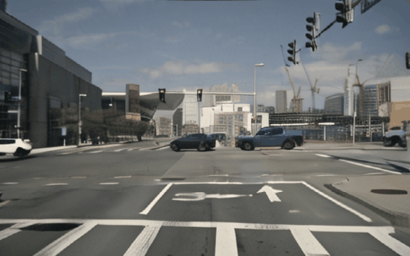}
      & \includegraphics[width=\hsize]{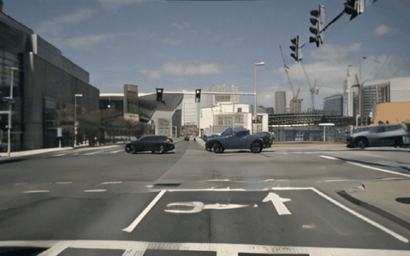} 
      & \includegraphics[width=\hsize]{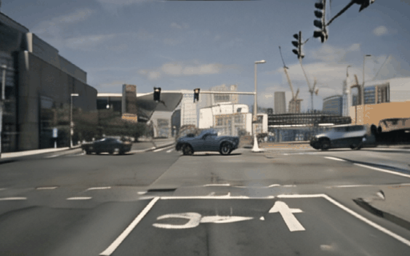}\\
      GB & \includegraphics[width=\hsize]{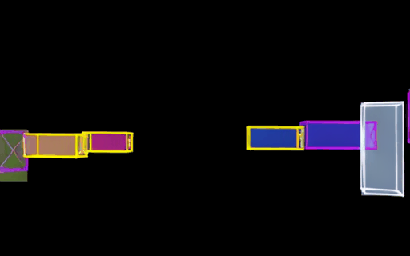}
      & \includegraphics[width=\hsize]{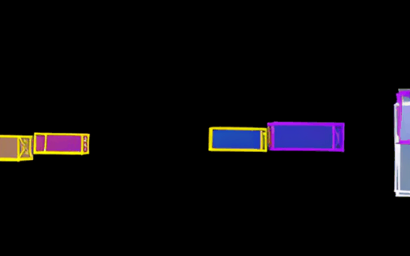}
      & \includegraphics[width=\hsize]{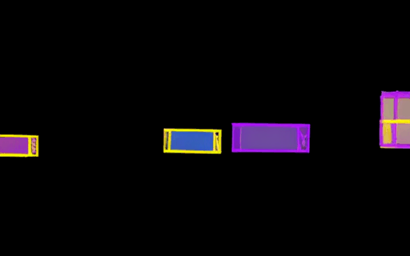}
      & \includegraphics[width=\hsize]{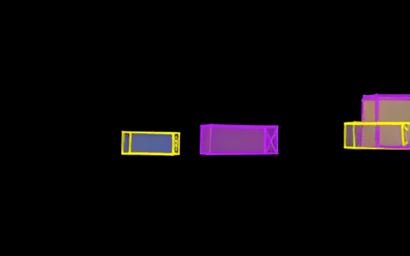}
      & \includegraphics[width=\hsize]{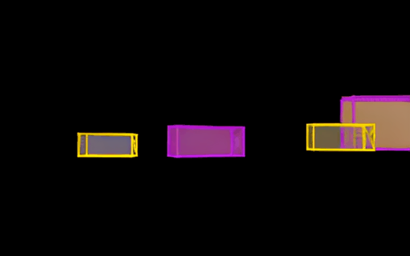}\\
    \hdashline
    GT & \includegraphics[width=\hsize]{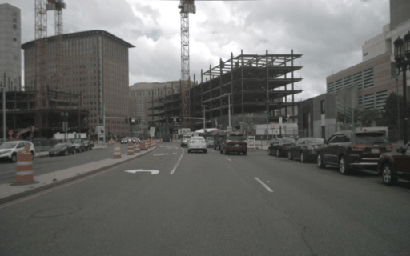}   
      & \includegraphics[width=\hsize]{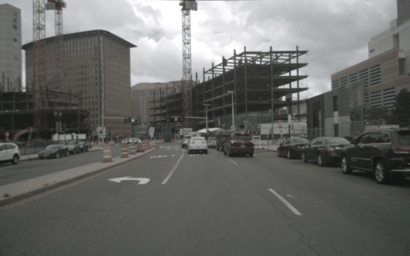} 
      & \includegraphics[width=\hsize]{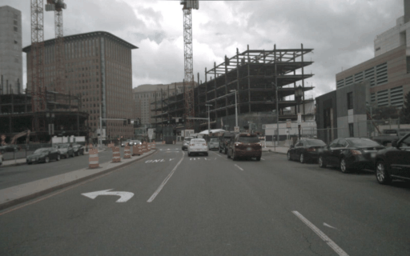}
      & \includegraphics[width=\hsize]{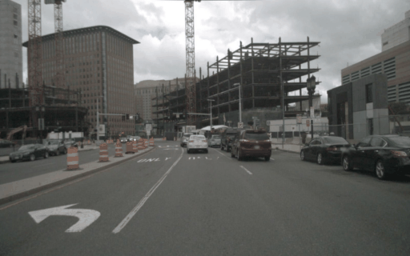}
      & \includegraphics[width=\hsize]{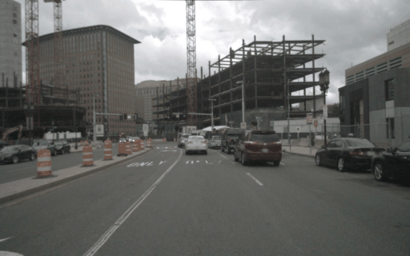} \\
      GF & \includegraphics[width=\hsize]{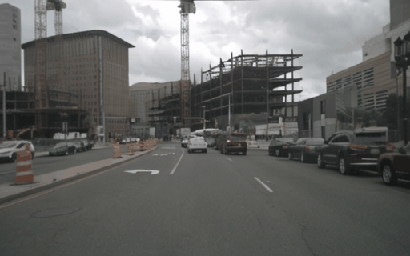}   
      & \includegraphics[width=\hsize]{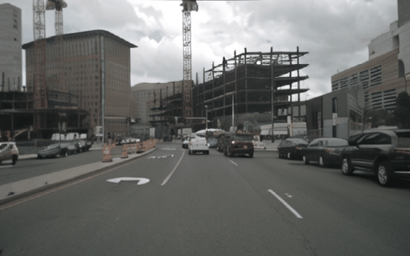} 
      & \includegraphics[width=\hsize]{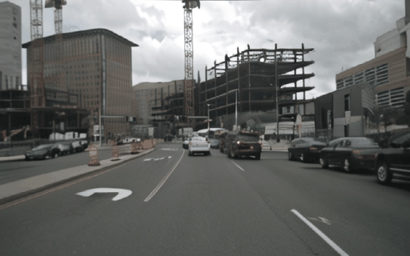}
      & \includegraphics[width=\hsize]{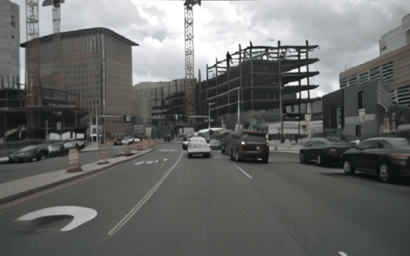}
      & \includegraphics[width=\hsize]{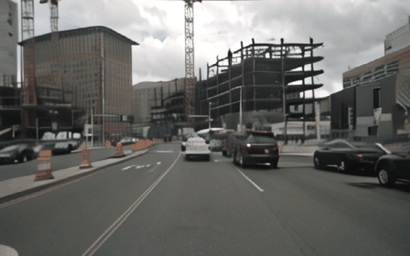} \\
      GB & \includegraphics[width=\hsize]{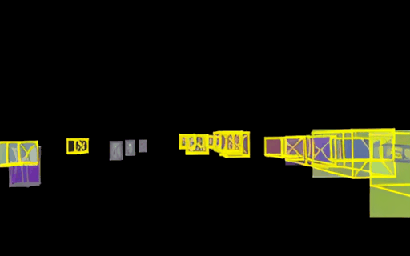}   
      & \includegraphics[width=\hsize]{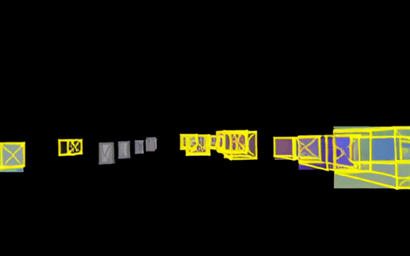} 
      & \includegraphics[width=\hsize]{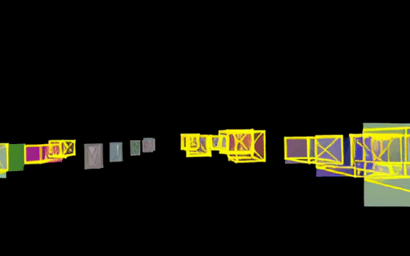}
      & \includegraphics[width=\hsize]{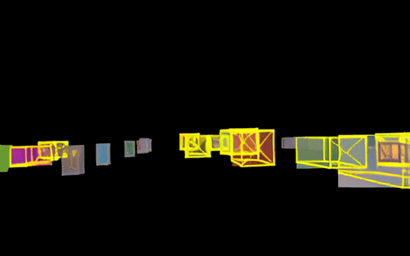}
      & \includegraphics[width=\hsize]{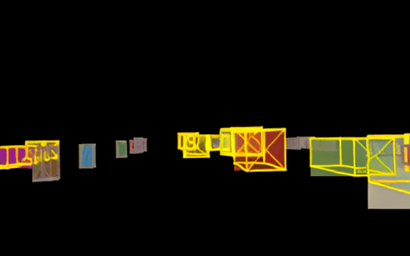} \\
      \hdashline
      \hdashline
    GT & \includegraphics[width=\hsize]{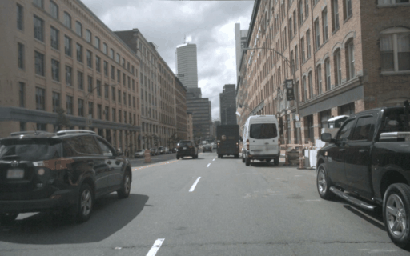}   
      & \includegraphics[width=\hsize]{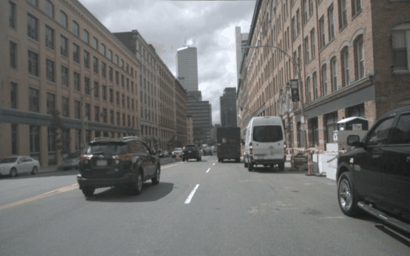} 
      & \includegraphics[width=\hsize]{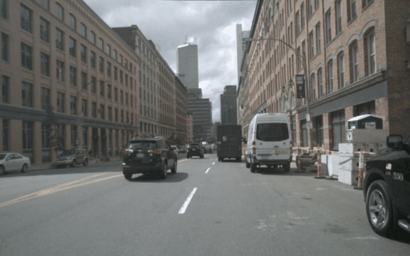}
      & \includegraphics[width=\hsize]{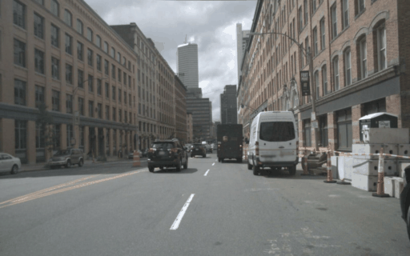}
      & \includegraphics[width=\hsize]{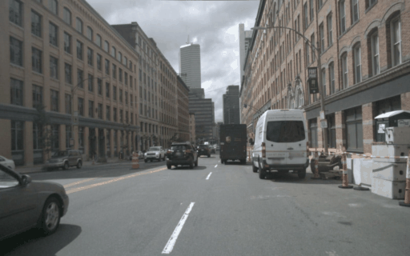} \\
      GF & \includegraphics[width=\hsize]{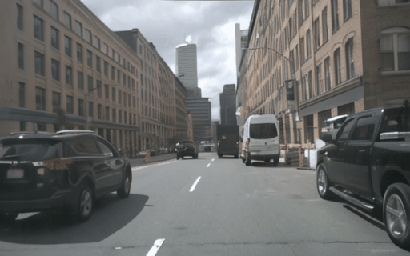}
      & \includegraphics[width=\hsize]{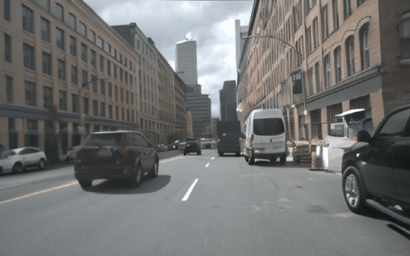}  
      & \includegraphics[width=\hsize]{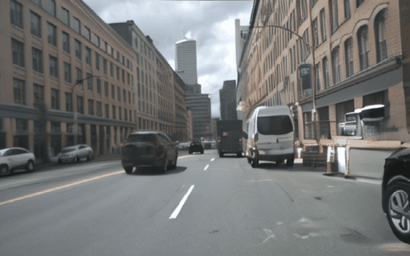}
      & \includegraphics[width=\hsize]{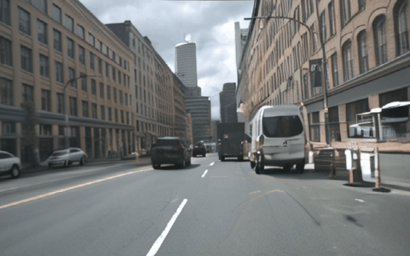} 
      & \includegraphics[width=\hsize]{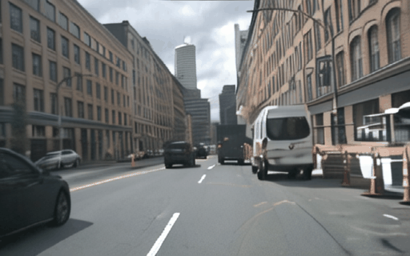}\\
      GB & \includegraphics[width=\hsize]{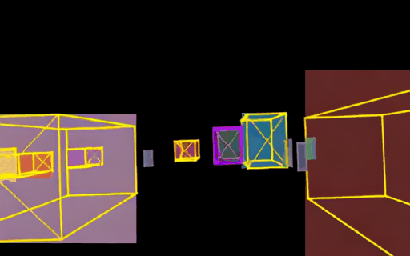}
      & \includegraphics[width=\hsize]{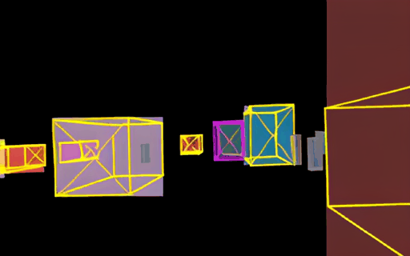}
      & \includegraphics[width=\hsize]{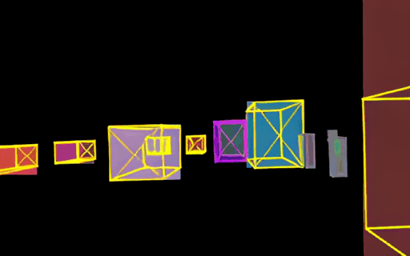}
      & \includegraphics[width=\hsize]{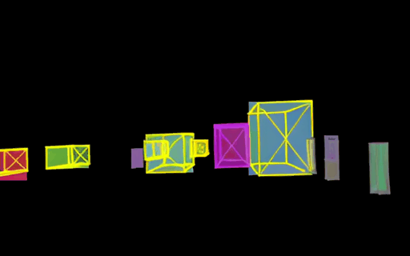}
      & \includegraphics[width=\hsize]{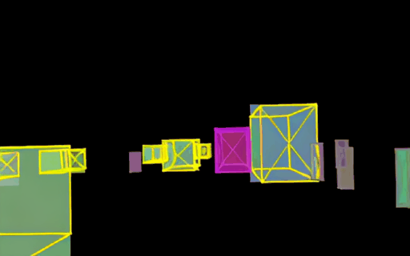}\\ 
\end{tabular}
    \caption{3D bounding box generations and motion-controlled video generations for various scenes on NuScenes.}
    \label{fig:nuscenes_demo}
\end{figure}
\clearpage
\subsection{Special Scenarios: Turning}
In the following section, we showcase a range of generation results with special scenarios, such as turning (day/night). These results are produced by our 3-to-1 generation pipeline. Each visualization displays every 6th frame from a 25-frame clip, with the actual video generated at a frame rate of 7 fps. In the leftmost column labels, GB represents generated bounding box frames, and GF represents generated frames.
\begin{figure}[ht]
    \setlength\tabcolsep{3pt} 
    \centering
    \begin{tabular}{@{} r M{0.18\linewidth} M{0.18\linewidth} M{0.18\linewidth} M{0.18\linewidth} M{0.18\linewidth} @{}}
    & Frame 1 & Frame 7 & Frame 13 & Frame 19  & Frame 25\\
      GT & \includegraphics[width=\hsize]{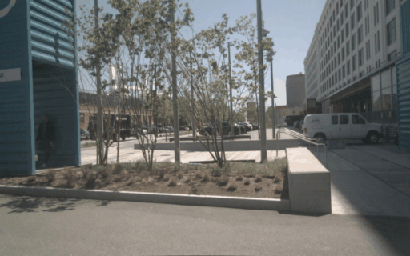}   
      & \includegraphics[width=\hsize]{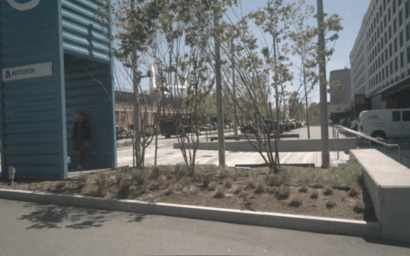} 
      & \includegraphics[width=\hsize]{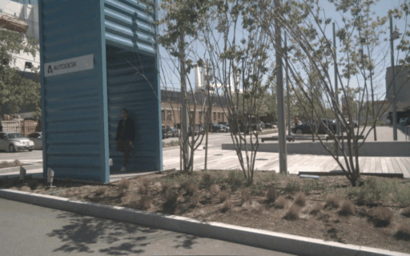}
      & \includegraphics[width=\hsize]{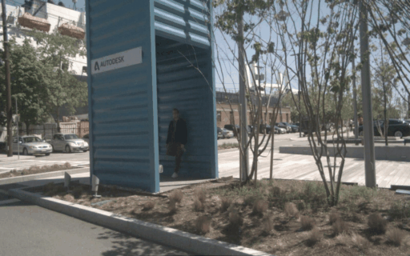}
      & \includegraphics[width=\hsize]{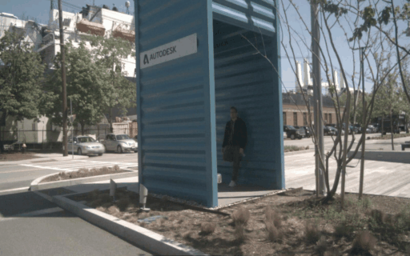} \\
      GF & \includegraphics[width=\hsize]{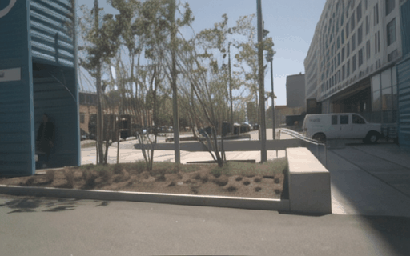}   
      & \includegraphics[width=\hsize]{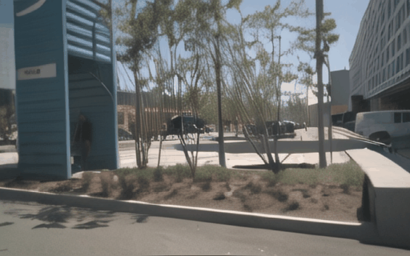} 
      & \includegraphics[width=\hsize]{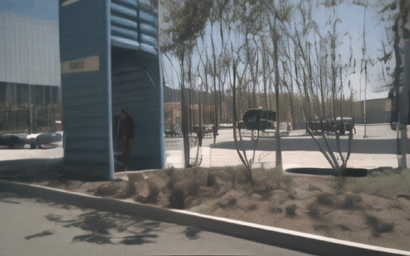}
      & \includegraphics[width=\hsize]{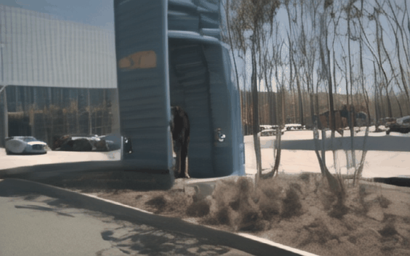}
      & \includegraphics[width=\hsize]{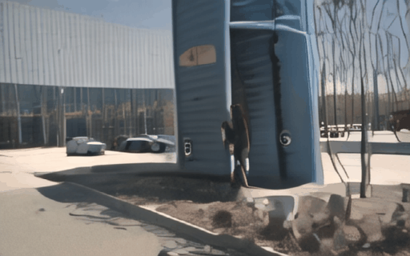} \\
      GB & \includegraphics[width=\hsize]{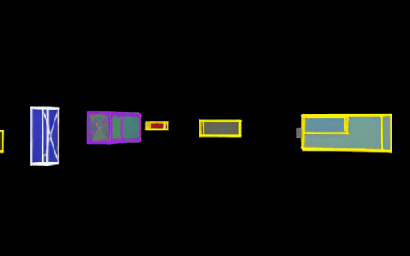}  
      & \includegraphics[width=\hsize]{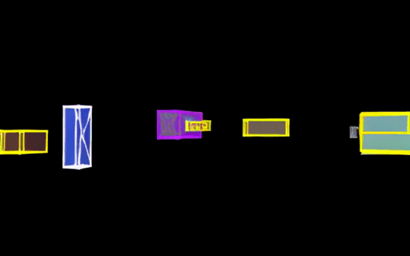} 
      & \includegraphics[width=\hsize]{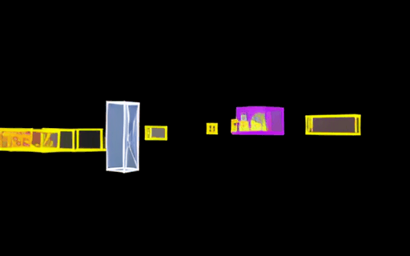}
      & \includegraphics[width=\hsize]{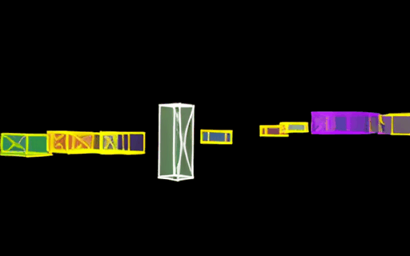}
      & \includegraphics[width=\hsize]{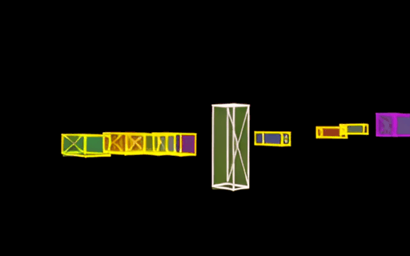} \\
    \hdashline
    GT & \includegraphics[width=\hsize]{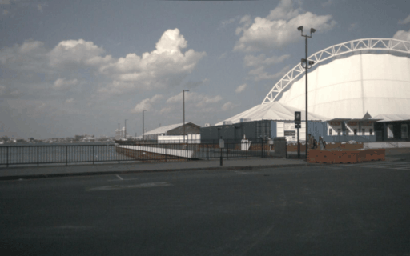}   
      & \includegraphics[width=\hsize]{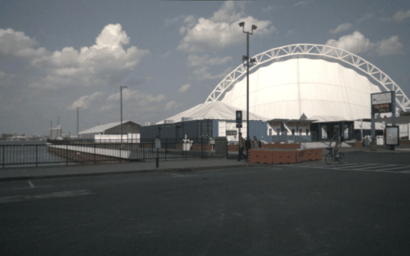} 
      & \includegraphics[width=\hsize]{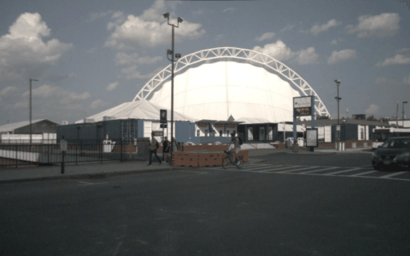}
      & \includegraphics[width=\hsize]{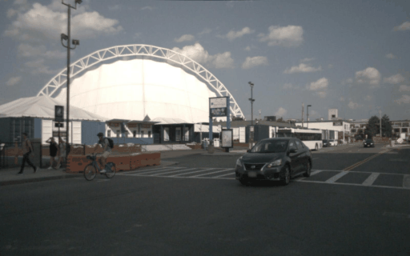}
      & \includegraphics[width=\hsize]{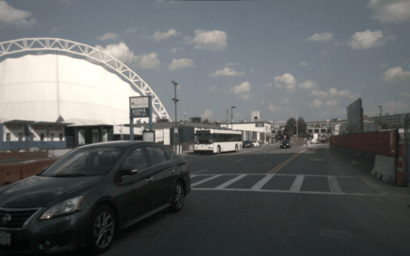} \\
      GF & \includegraphics[width=\hsize]{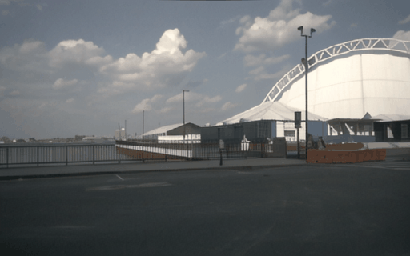}
      & \includegraphics[width=\hsize]{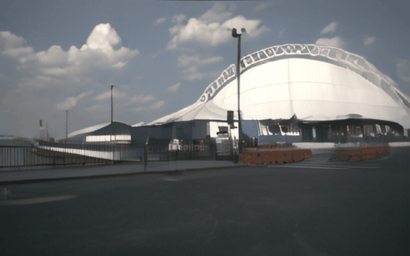}  
      & \includegraphics[width=\hsize]{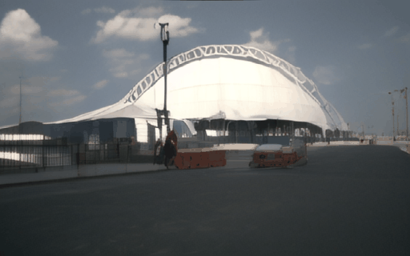}
      & \includegraphics[width=\hsize]{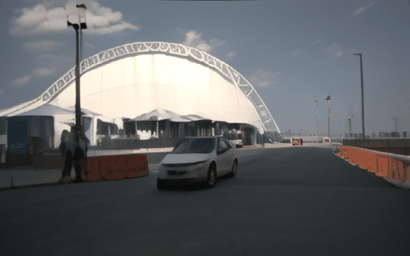} 
      & \includegraphics[width=\hsize]{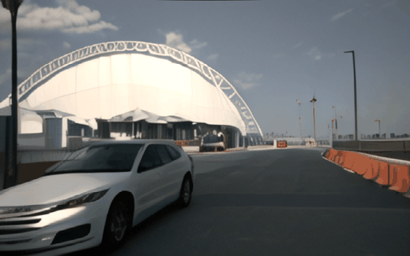}\\
      GB & \includegraphics[width=\hsize]{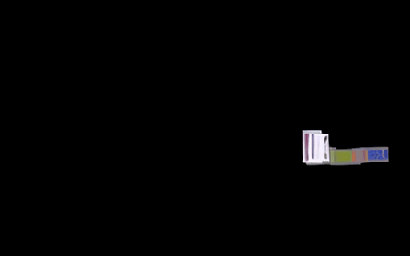}
      & \includegraphics[width=\hsize]{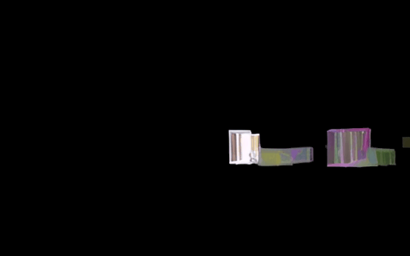}
      & \includegraphics[width=\hsize]{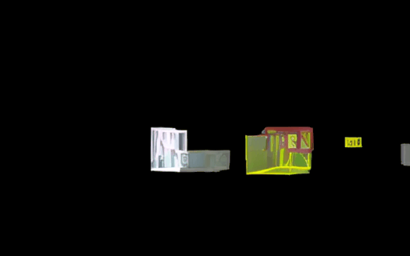}
      & \includegraphics[width=\hsize]{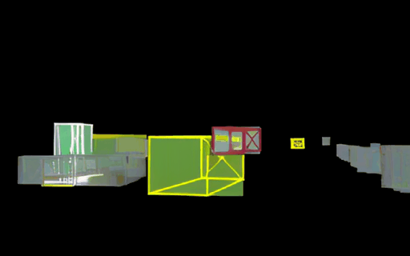}
      & \includegraphics[width=\hsize]{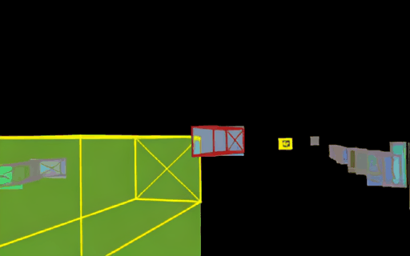}\\ 
      \hdashline
    GT & \includegraphics[width=\hsize]{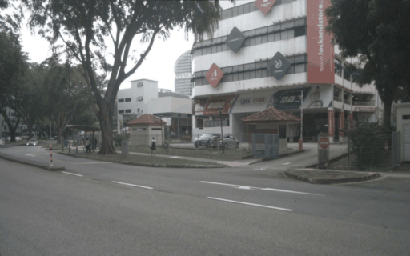}   
      & \includegraphics[width=\hsize]{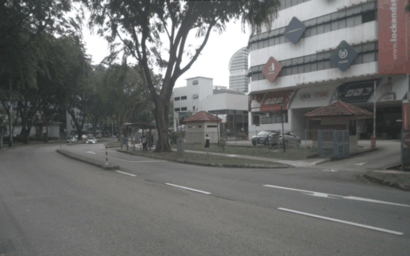} 
      & \includegraphics[width=\hsize]{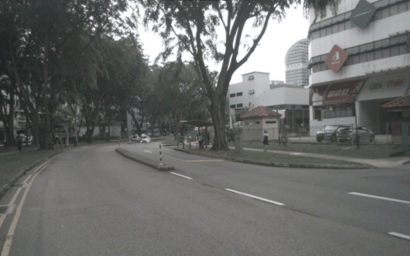}
      & \includegraphics[width=\hsize]{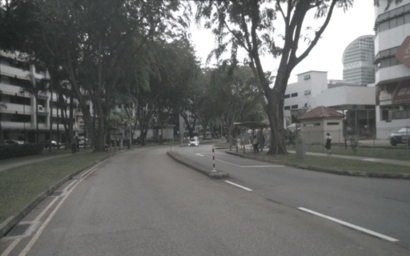}
      & \includegraphics[width=\hsize]{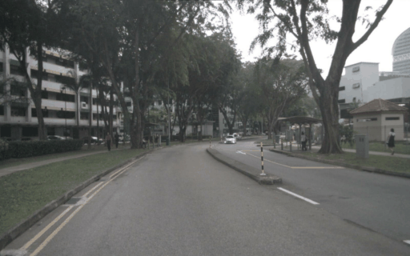} \\
      GF & \includegraphics[width=\hsize]{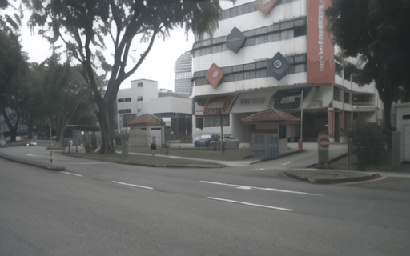}
      & \includegraphics[width=\hsize]{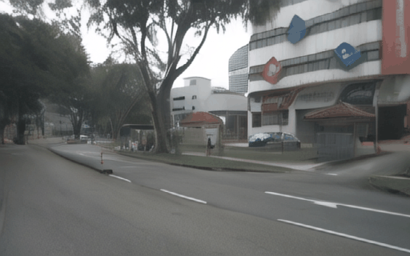}  
      & \includegraphics[width=\hsize]{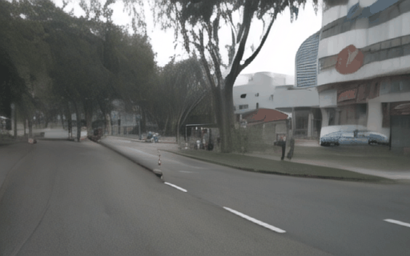}
      & \includegraphics[width=\hsize]{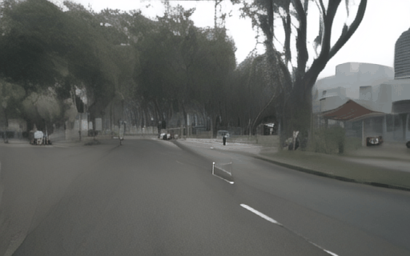} 
      & \includegraphics[width=\hsize]{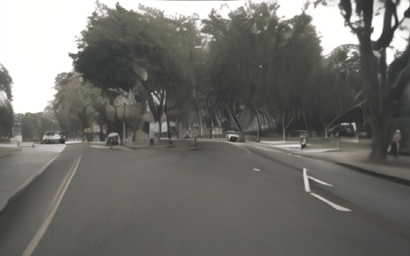}\\
      GB & \includegraphics[width=\hsize]{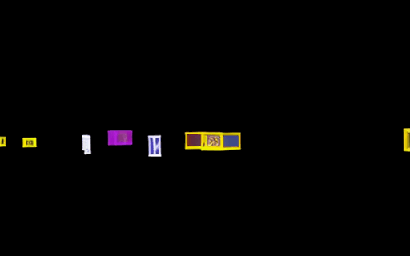}
      & \includegraphics[width=\hsize]{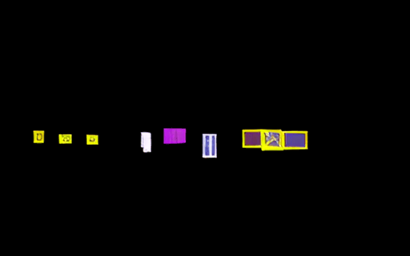}
      & \includegraphics[width=\hsize]{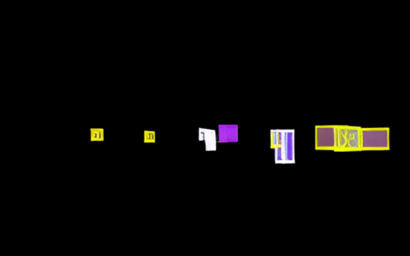}
      & \includegraphics[width=\hsize]{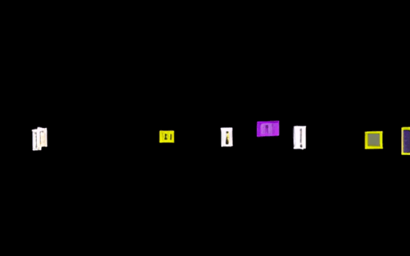}
      & \includegraphics[width=\hsize]{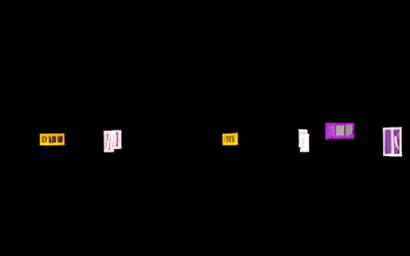}\\ 
\end{tabular}
    \caption{3D bounding box generations and motion-controlled video generations for turning scenarios on NuScenes.}
    \label{fig:nuscenes_turning_demo}
\end{figure}
\begin{figure}[ht]
    \setlength\tabcolsep{3pt} 
    \centering
    \begin{tabular}{@{} r M{0.18\linewidth} M{0.18\linewidth} M{0.18\linewidth} M{0.18\linewidth} M{0.18\linewidth} @{}}
    & Frame 1 & Frame 7 & Frame 13 & Frame 19  & Frame 25\\
      GF & \includegraphics[width=\hsize]{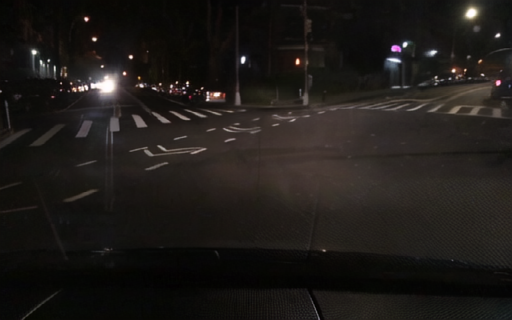}
      & \includegraphics[width=\hsize]{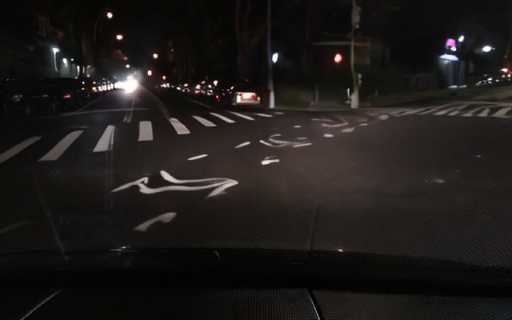}  
      & \includegraphics[width=\hsize]{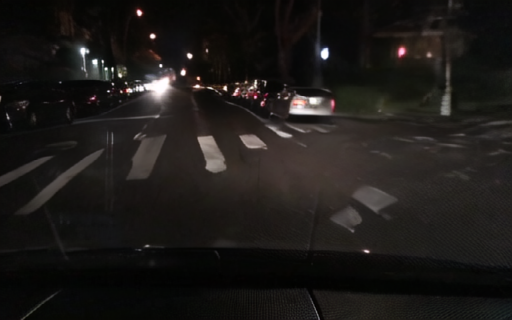}
      & \includegraphics[width=\hsize]{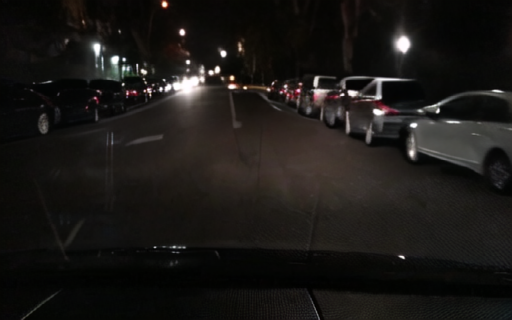} 
      & \includegraphics[width=\hsize]{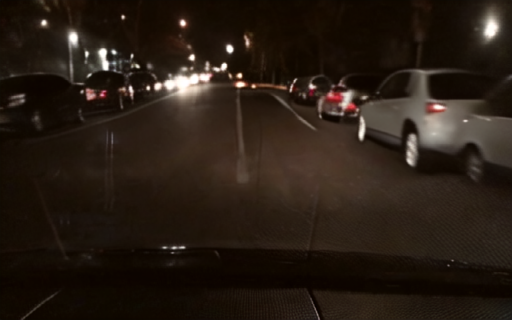}\\
      GB & \includegraphics[width=\hsize]{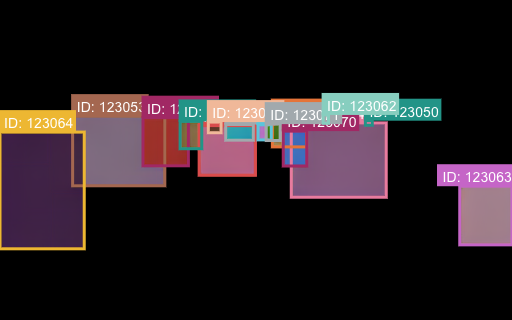}
      & \includegraphics[width=\hsize]{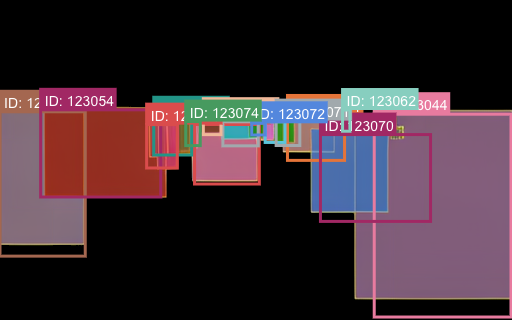}  
      & \includegraphics[width=\hsize]{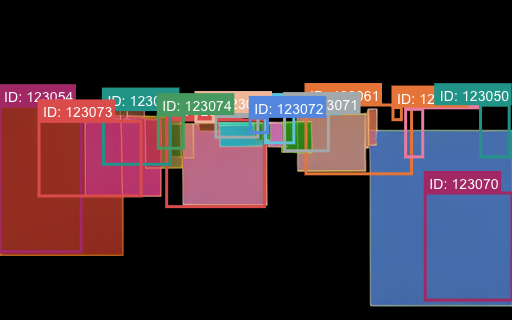}
      & \includegraphics[width=\hsize]{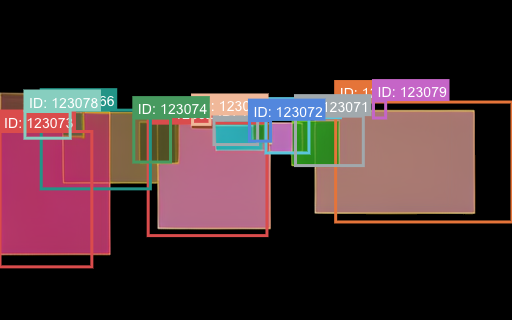} 
      & \includegraphics[width=\hsize]{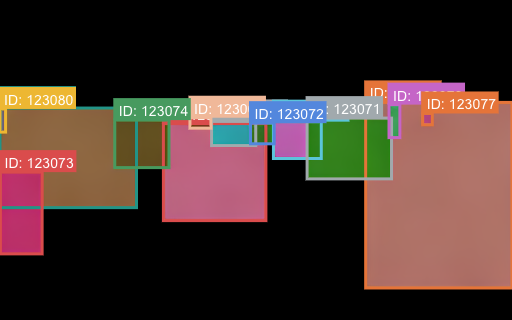}\\
    \hdashline
      GF & \includegraphics[width=\hsize]{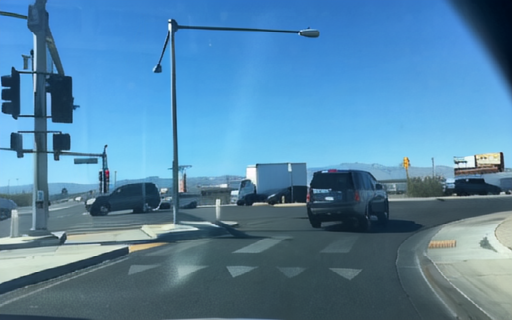}
      & \includegraphics[width=\hsize]{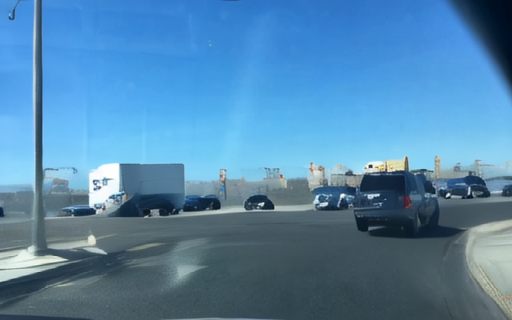}  
      & \includegraphics[width=\hsize]{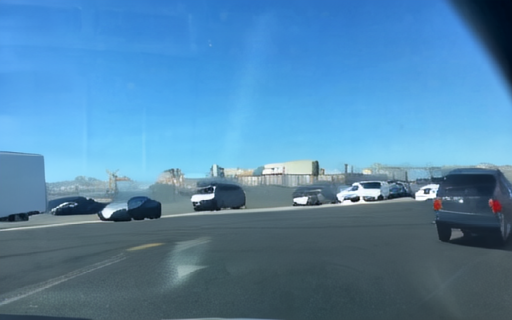}
      & \includegraphics[width=\hsize]{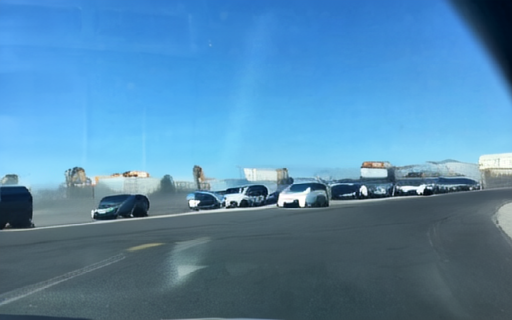} 
      & \includegraphics[width=\hsize]{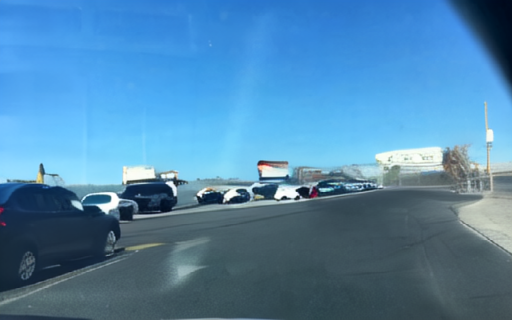}\\
      GB & \includegraphics[width=\hsize]{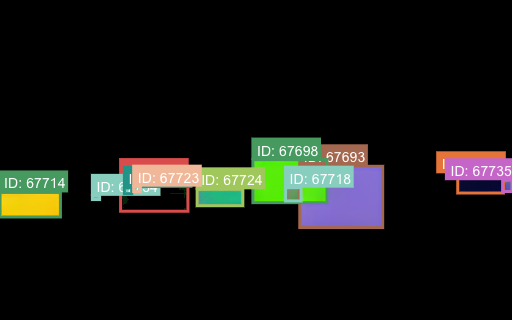}
      & \includegraphics[width=\hsize]{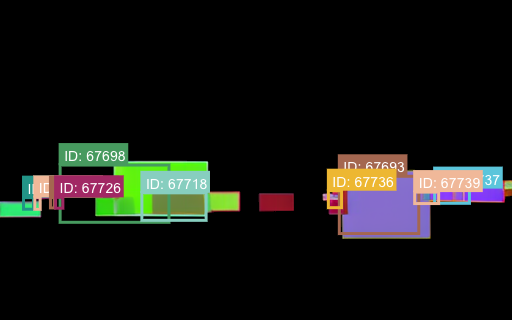}  
      & \includegraphics[width=\hsize]{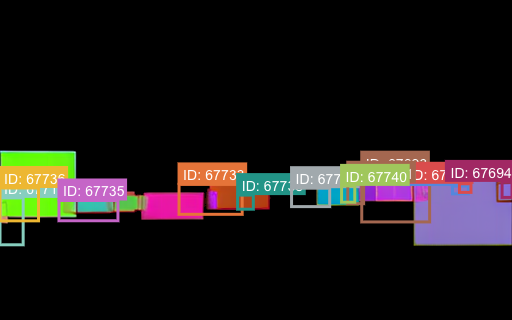}
      & \includegraphics[width=\hsize]{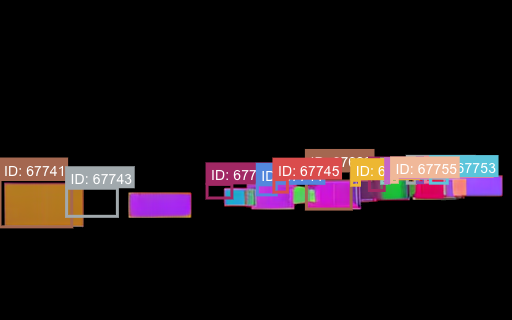} 
      & \includegraphics[width=\hsize]{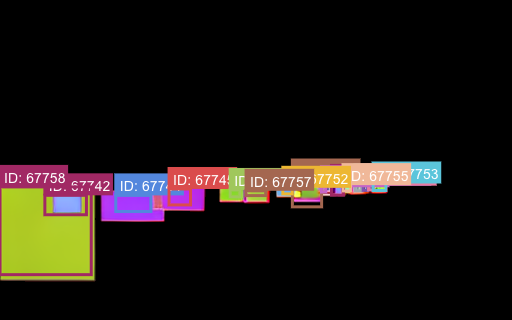}\\
      \hdashline
      GF & \includegraphics[width=\hsize]{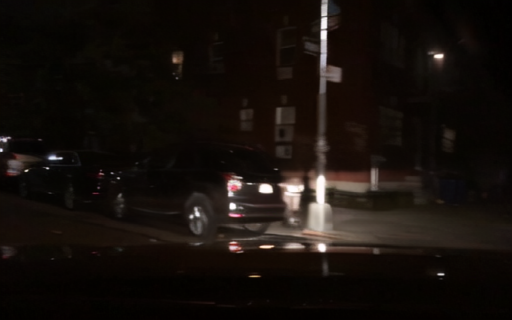}
      & \includegraphics[width=\hsize]{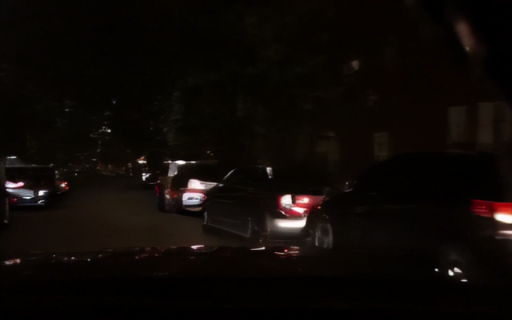}  
      & \includegraphics[width=\hsize]{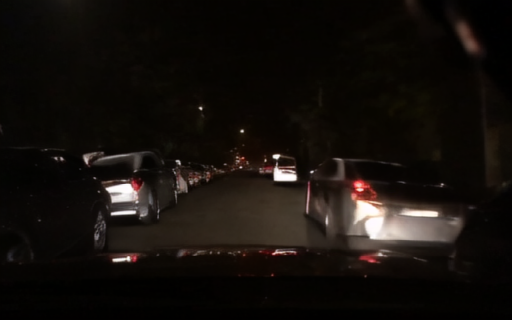}
      & \includegraphics[width=\hsize]{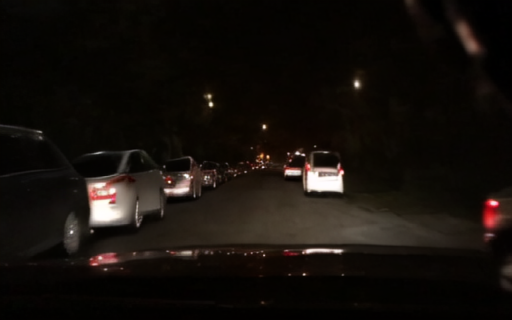} 
      & \includegraphics[width=\hsize]{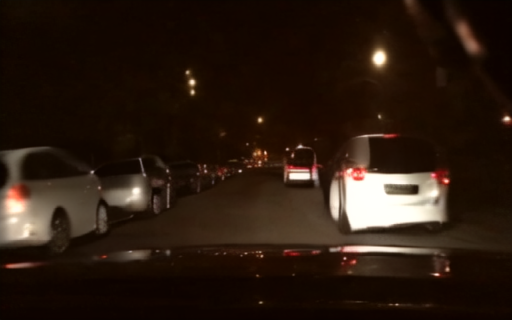}\\
      GB & \includegraphics[width=\hsize]{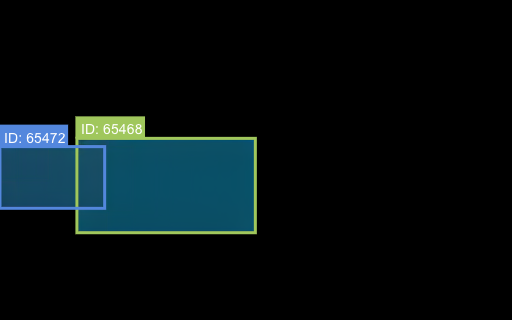}
      & \includegraphics[width=\hsize]{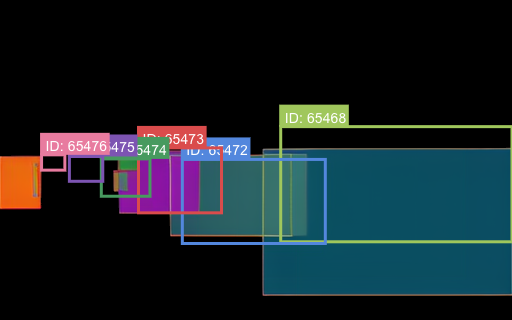}  
      & \includegraphics[width=\hsize]{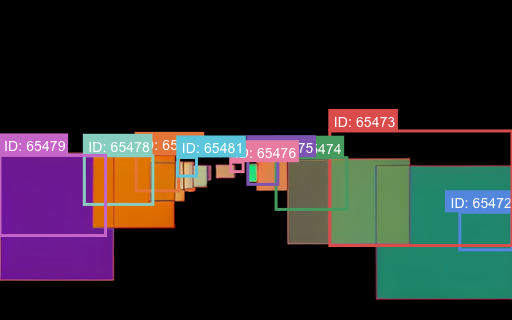}
      & \includegraphics[width=\hsize]{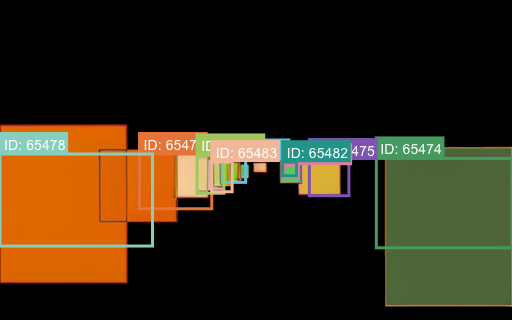} 
      & \includegraphics[width=\hsize]{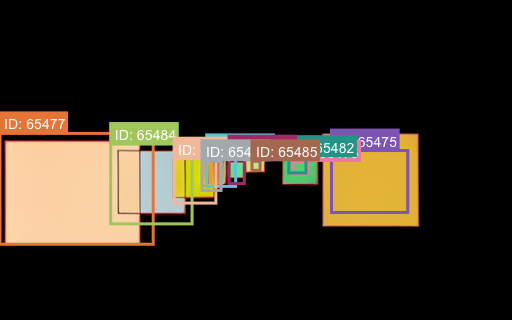}\\
      \hdashline
      GF & \includegraphics[width=\hsize]{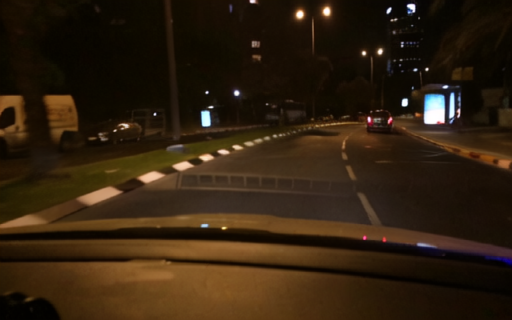}
      & \includegraphics[width=\hsize]{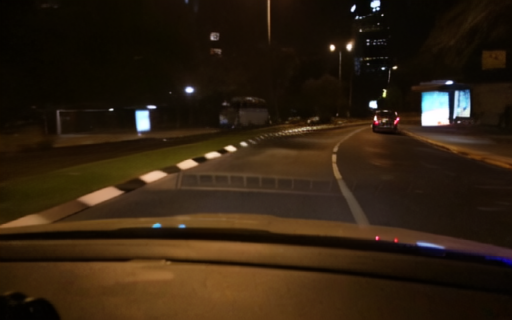}  
      & \includegraphics[width=\hsize]{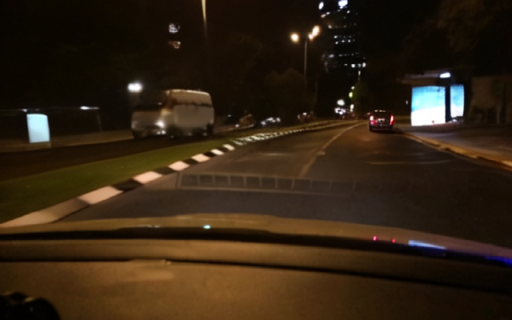}
      & \includegraphics[width=\hsize]{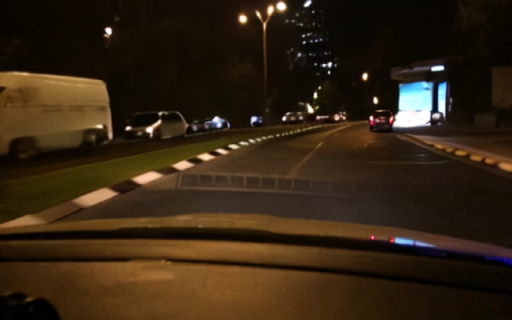} 
      & \includegraphics[width=\hsize]{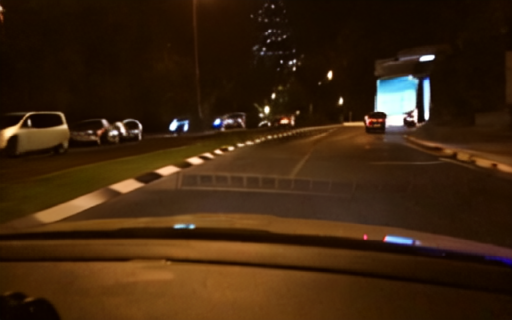}\\
      GB & \includegraphics[width=\hsize]{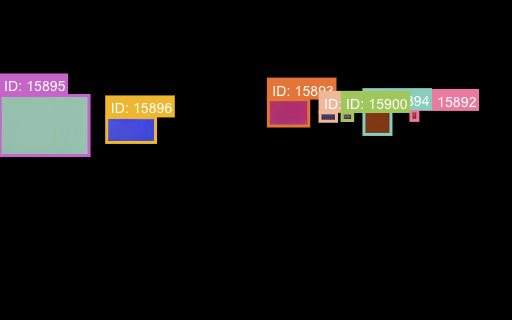}
      & \includegraphics[width=\hsize]{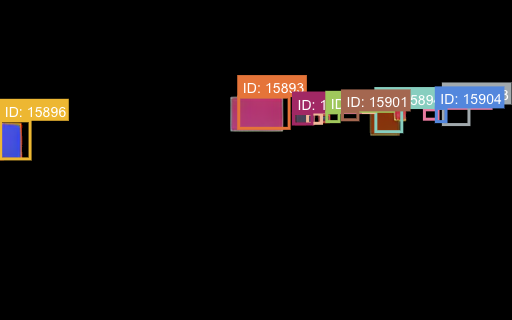}  
      & \includegraphics[width=\hsize]{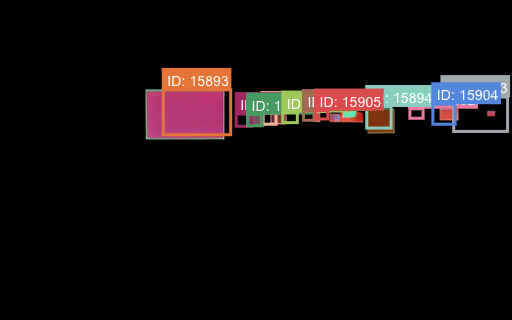}
      & \includegraphics[width=\hsize]{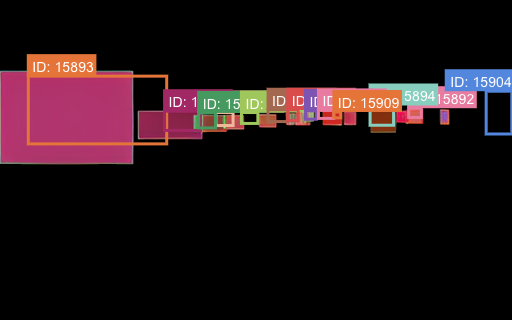} 
      & \includegraphics[width=\hsize]{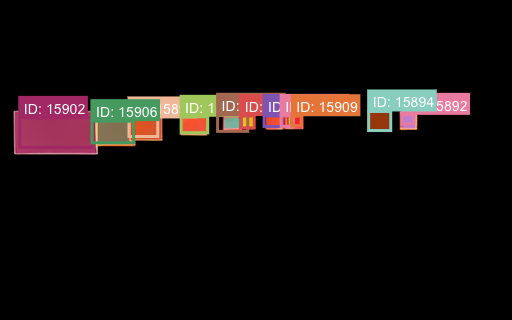}\\
\end{tabular}
    \caption{2D bounding box generations and motion-controlled video generations for turning scenarios on BDD test-split.}
    \label{fig:turning_demos_train}
\end{figure}

%% file: sections/appendix/frame-by-frame_control.tex
\clearpage
\subsection{Visualizing Motion Controlled Generation via \modelvid}
\begin{figure}[h]
    \setlength\tabcolsep{3pt} 
    \centering
    \begin{tabular}{@{} M{0.18\linewidth} M{0.18\linewidth} M{0.18\linewidth} M{0.18\linewidth} M{0.18\linewidth} @{}}
     \includegraphics[width=\hsize]{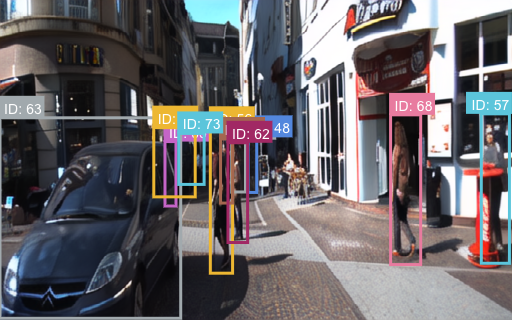}   
      & \includegraphics[width=\hsize]{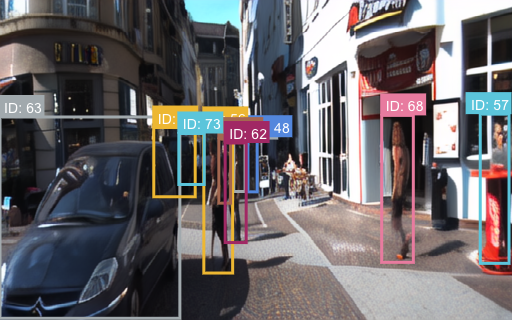} 
      & \includegraphics[width=\hsize]{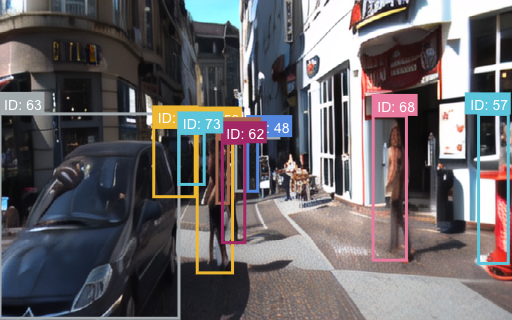}
      & \includegraphics[width=\hsize]{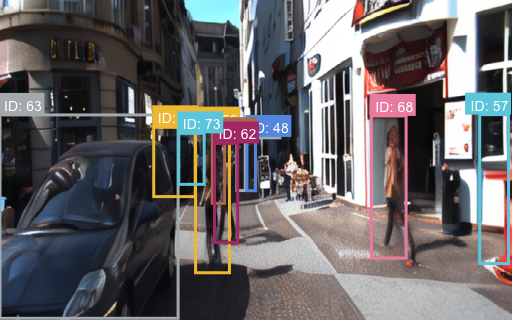}
      & \includegraphics[width=\hsize]{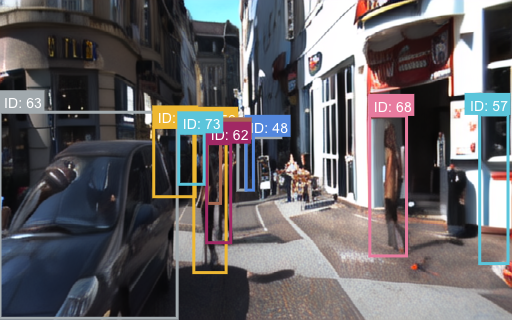} \\ 
      \includegraphics[width=\hsize]{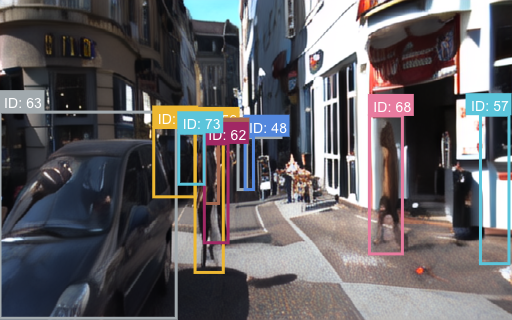}   
      & \includegraphics[width=\hsize]{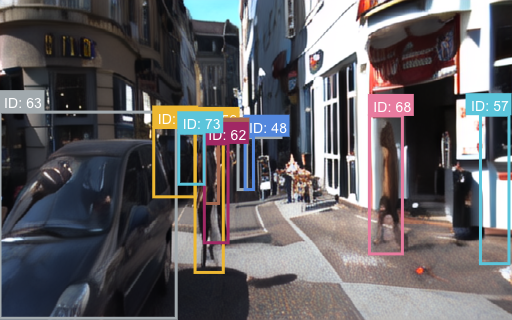} 
      & \includegraphics[width=\hsize]{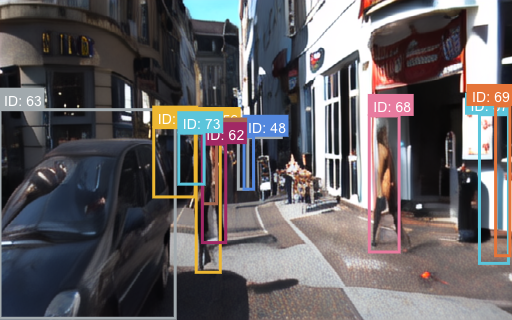}
      & \includegraphics[width=\hsize]{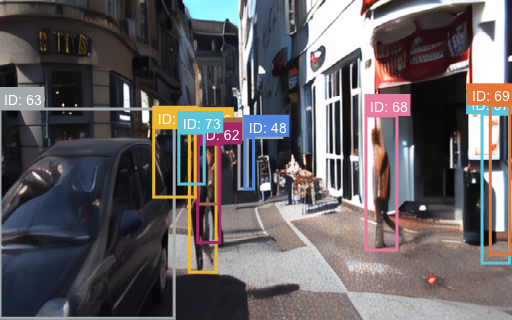}
      & \includegraphics[width=\hsize]{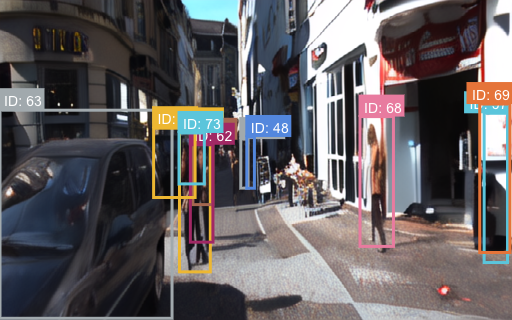} \\ 
       \includegraphics[width=\hsize]{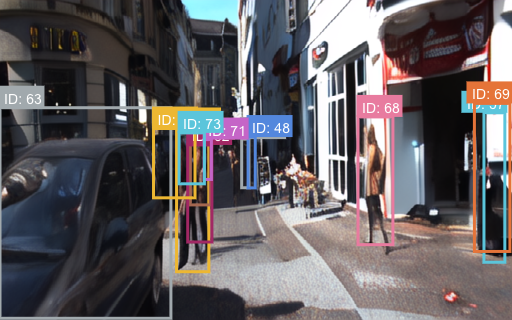}   
      & \includegraphics[width=\hsize]{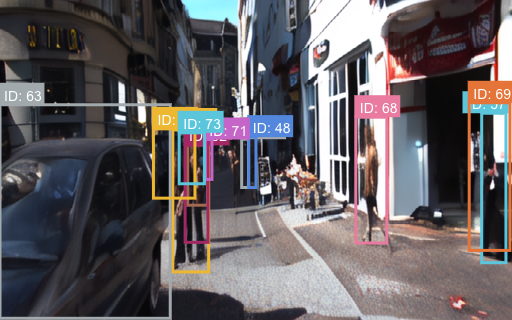} 
      & \includegraphics[width=\hsize]{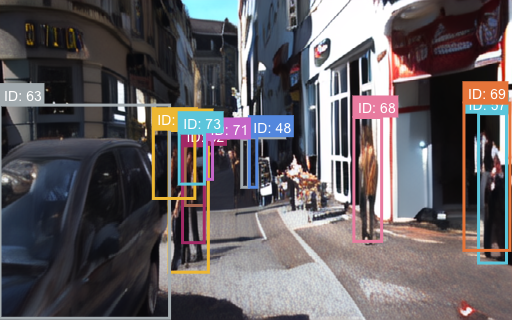}
      & \includegraphics[width=\hsize]{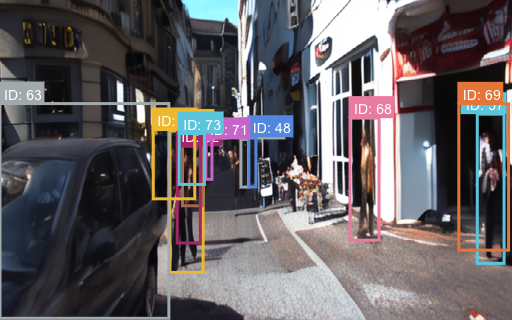}
      & \includegraphics[width=\hsize]{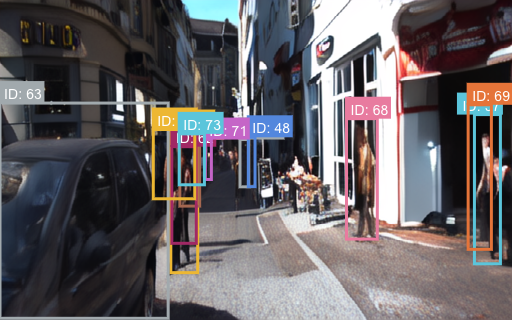} \\ 
      \includegraphics[width=\hsize]{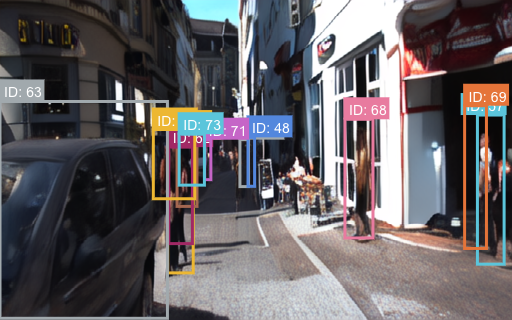}   
      & \includegraphics[width=\hsize]{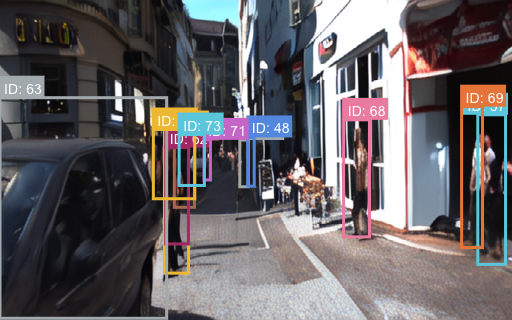} 
      & \includegraphics[width=\hsize]{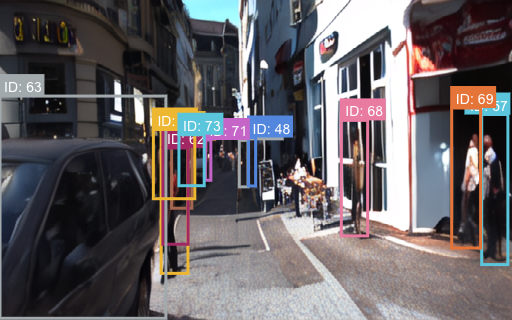}
      & \includegraphics[width=\hsize]{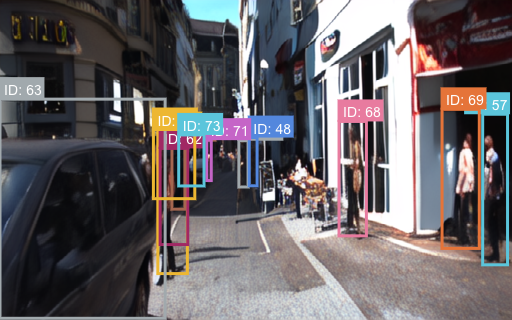}
      & \includegraphics[width=\hsize]{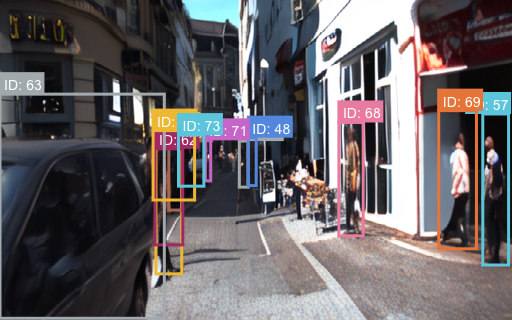} \\ 
      \includegraphics[width=\hsize]{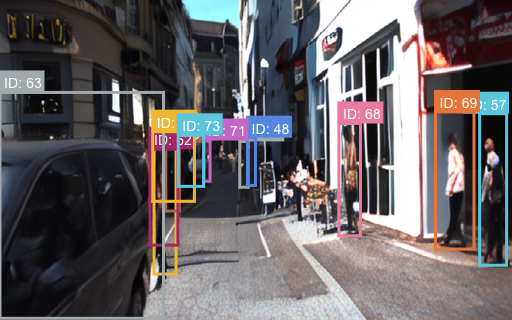}   
      & \includegraphics[width=\hsize]{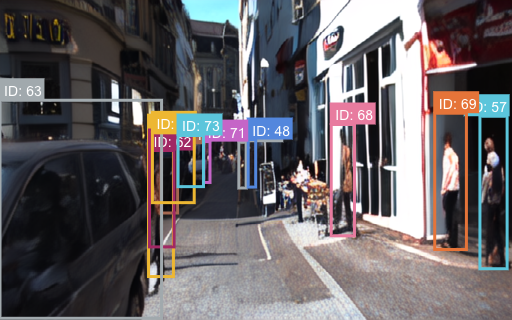} 
      & \includegraphics[width=\hsize]{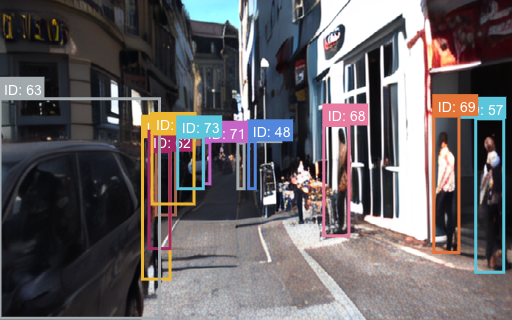}
      & \includegraphics[width=\hsize]{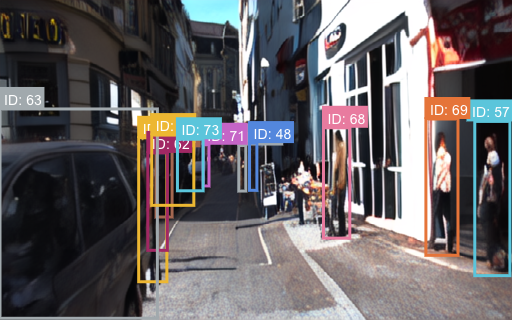}
      & \includegraphics[width=\hsize]{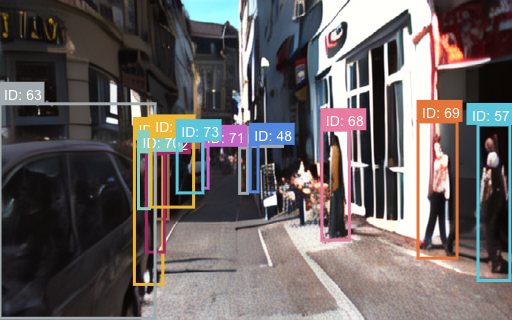} \\ 
\end{tabular}
    \caption{A frame-by-frame visualization of \modelvid~generation, conditioned on the ground-truth bbox frame sequence from the KITTI dataset, with conditions outlined in the plots.}
    \label{fig:kitti_controlnet_frame_by_frame}
\end{figure}

\begin{figure}[h]
    \setlength\tabcolsep{3pt} 
    \centering
    \begin{tabular}{@{} M{0.18\linewidth} M{0.18\linewidth} M{0.18\linewidth} M{0.18\linewidth} M{0.18\linewidth} @{}}
     \includegraphics[width=\hsize]{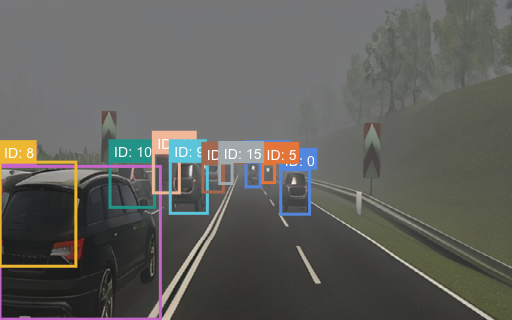}   
      & \includegraphics[width=\hsize]{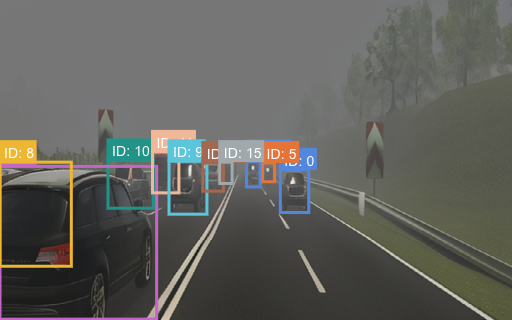} 
      & \includegraphics[width=\hsize]{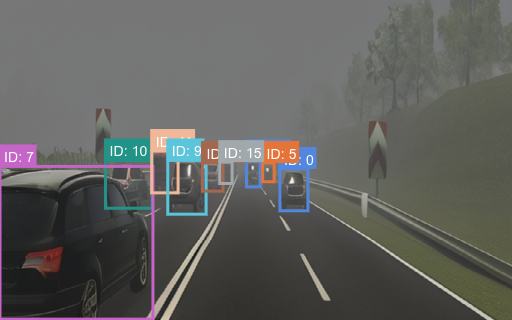}
      & \includegraphics[width=\hsize]{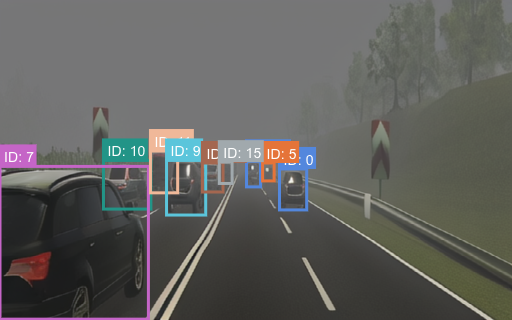}
      & \includegraphics[width=\hsize]{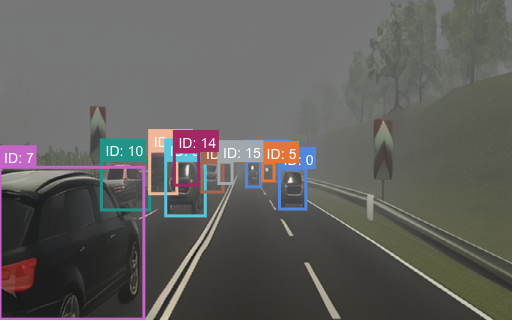} \\ 
      \includegraphics[width=\hsize]{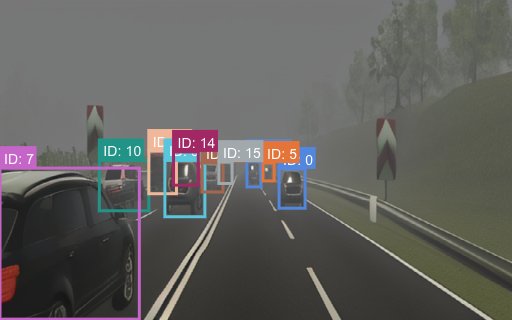}   
      & \includegraphics[width=\hsize]{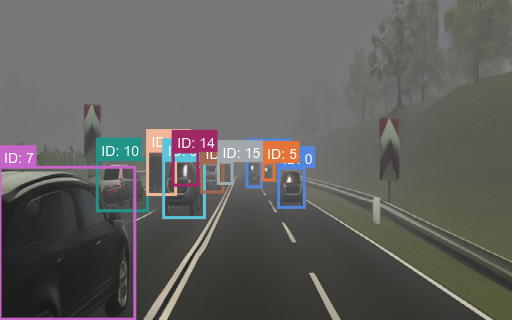} 
      & \includegraphics[width=\hsize]{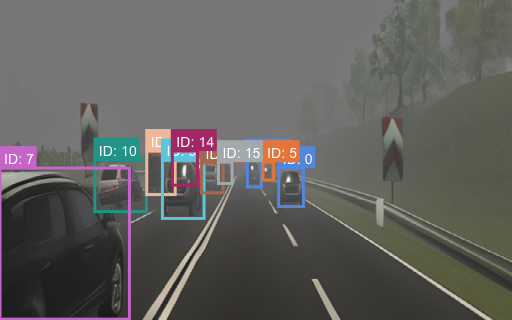}
      & \includegraphics[width=\hsize]{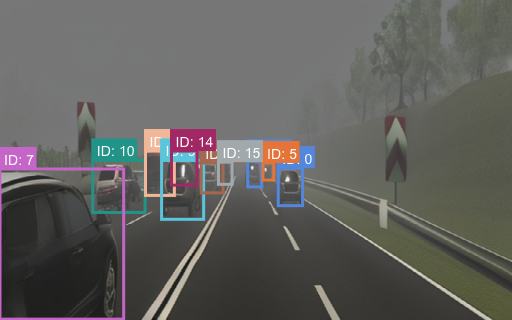}
      & \includegraphics[width=\hsize]{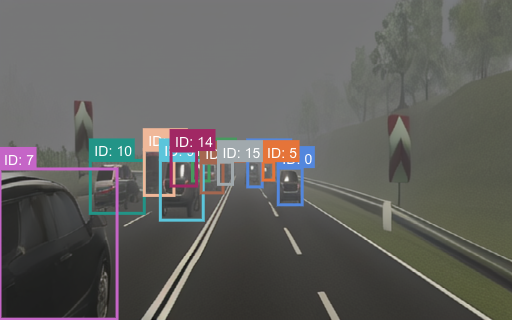} \\ 
       \includegraphics[width=\hsize]{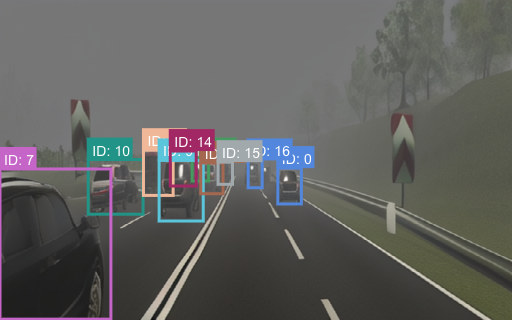}   
      & \includegraphics[width=\hsize]{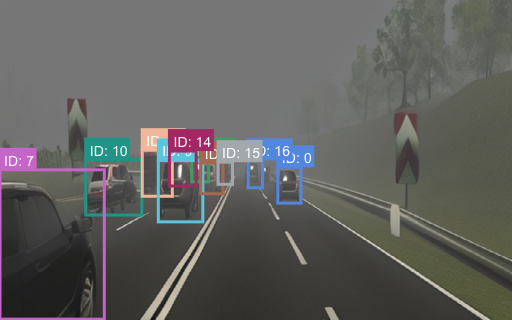} 
      & \includegraphics[width=\hsize]{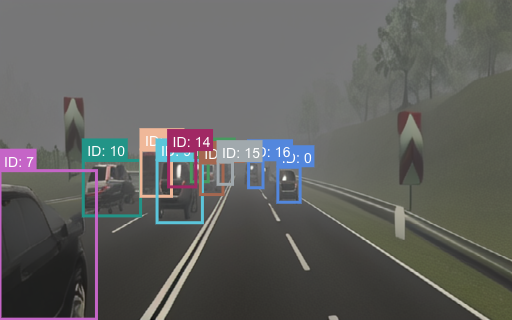}
      & \includegraphics[width=\hsize]{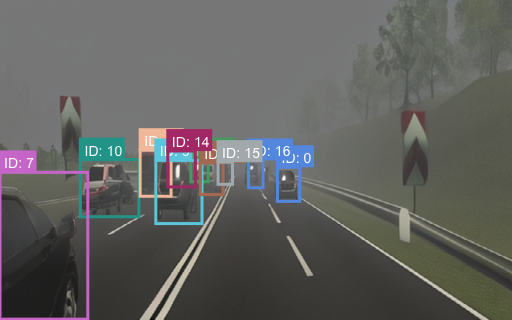}
      & \includegraphics[width=\hsize]{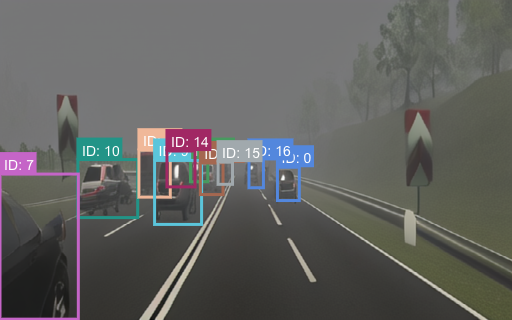} \\ 
      \includegraphics[width=\hsize]{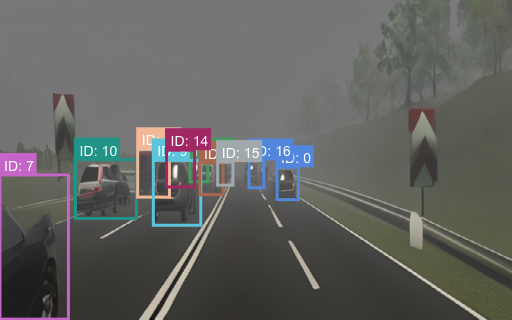}   
      & \includegraphics[width=\hsize]{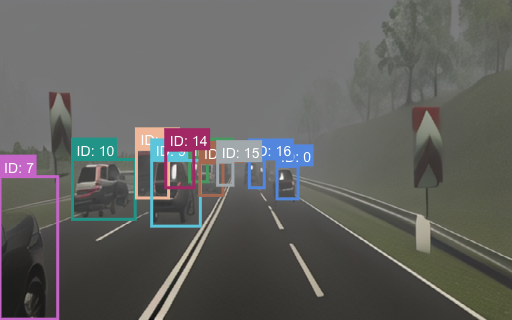} 
      & \includegraphics[width=\hsize]{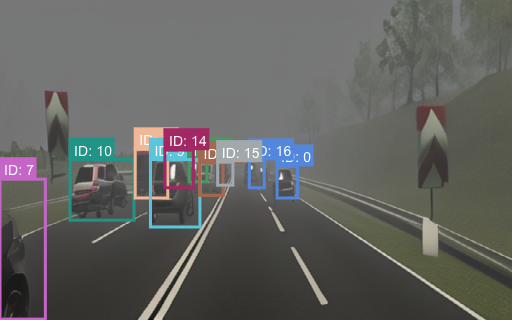}
      & \includegraphics[width=\hsize]{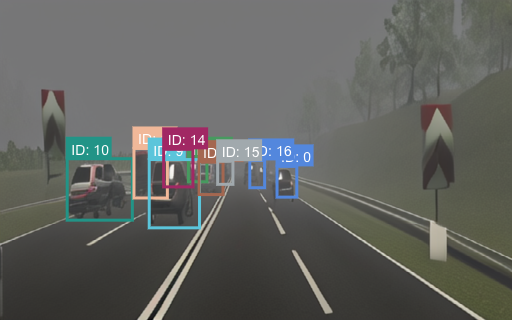}
      & \includegraphics[width=\hsize]{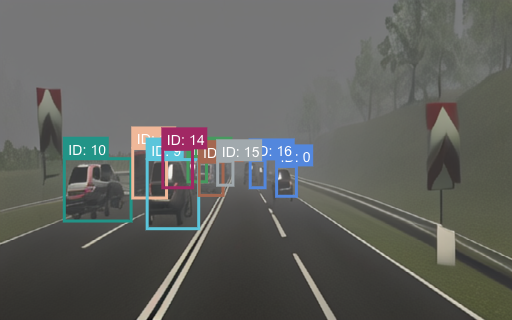} \\ 
      \includegraphics[width=\hsize]{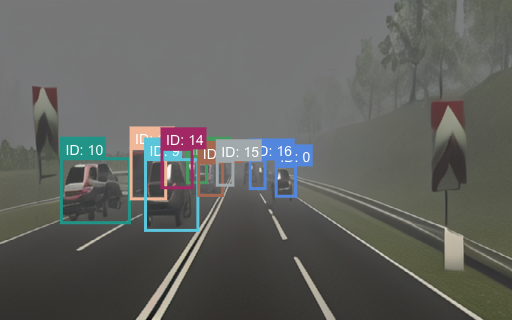}   
      & \includegraphics[width=\hsize]{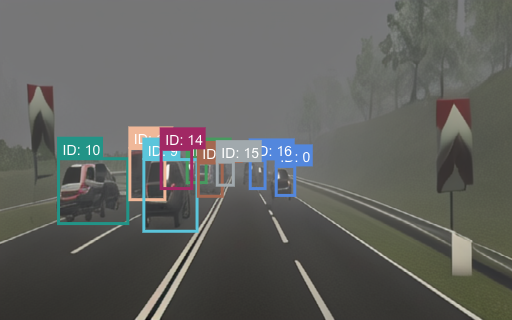} 
      & \includegraphics[width=\hsize]{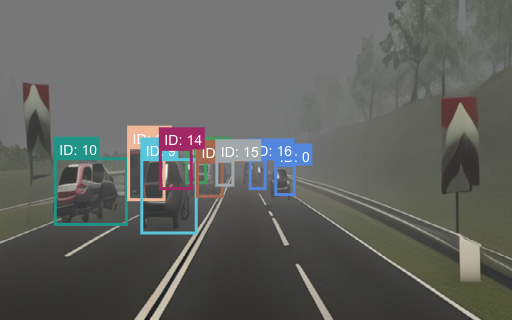}
      & \includegraphics[width=\hsize]{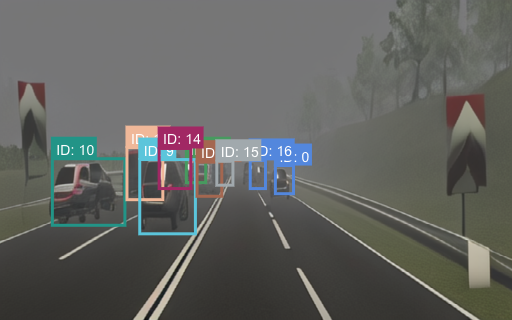}
      & \includegraphics[width=\hsize]{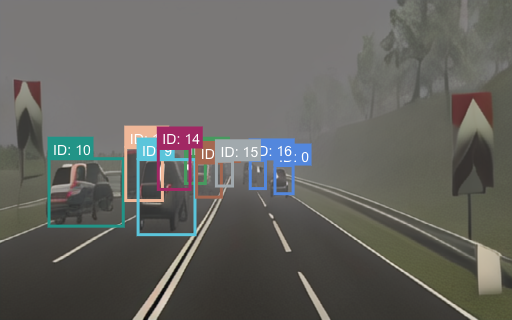} \\ 
\end{tabular}
    \caption{A frame-by-frame visualization of \modelvid~generation, conditioned on the ground-truth bbox frame sequence from the vKITTI dataset. The ground-truth bboxes are outlined in the plots.}
    \label{fig:vkitti_controlnet_frame_by_frame}
\end{figure}

\begin{figure}[h]
    \setlength\tabcolsep{3pt} 
    \centering
    \begin{tabular}{@{} M{0.18\linewidth} M{0.18\linewidth} M{0.18\linewidth} M{0.18\linewidth} M{0.18\linewidth} @{}}
     \includegraphics[width=\hsize]{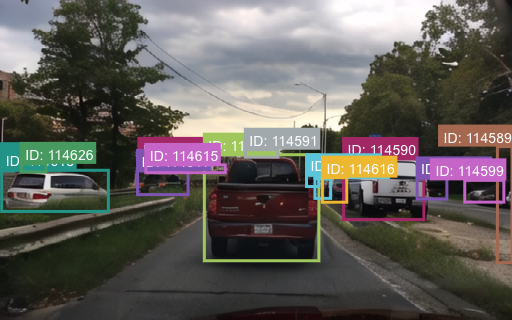}   
      & \includegraphics[width=\hsize]{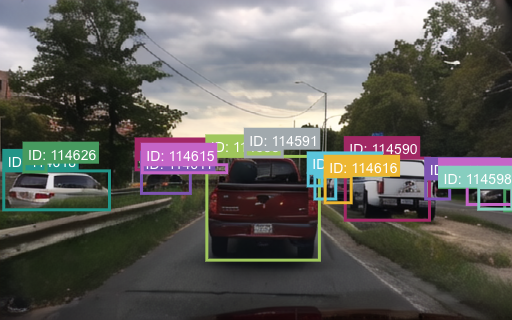} 
      & \includegraphics[width=\hsize]{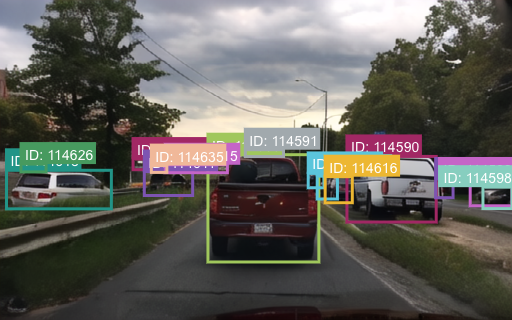}
      & \includegraphics[width=\hsize]{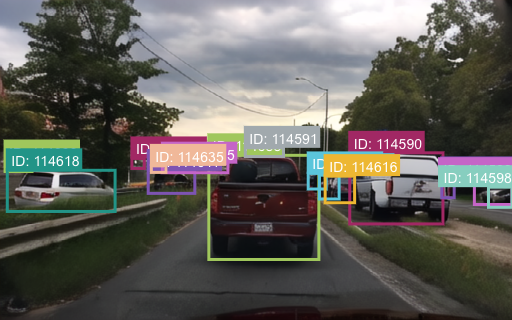}
      & \includegraphics[width=\hsize]{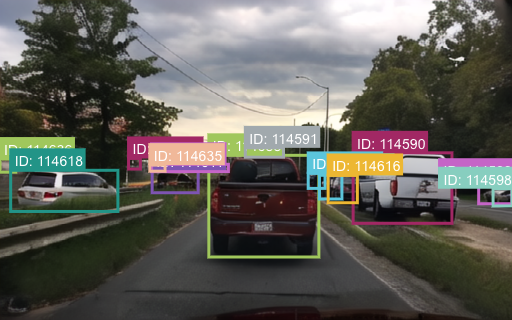} \\ 
      \includegraphics[width=\hsize]{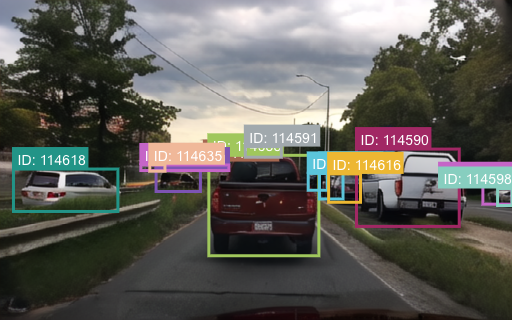}   
      & \includegraphics[width=\hsize]{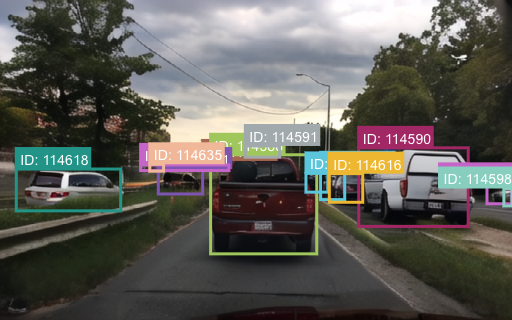} 
      & \includegraphics[width=\hsize]{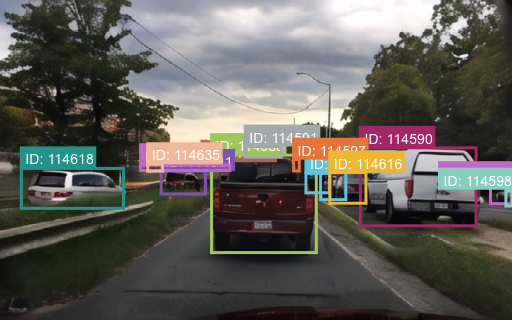}
      & \includegraphics[width=\hsize]{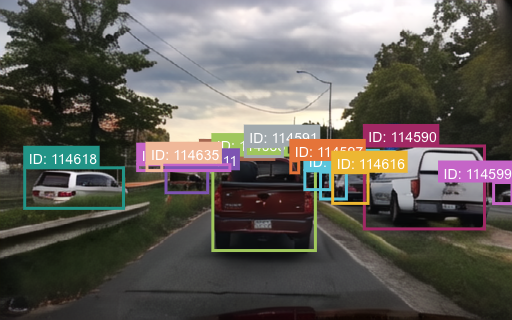}
      & \includegraphics[width=\hsize]{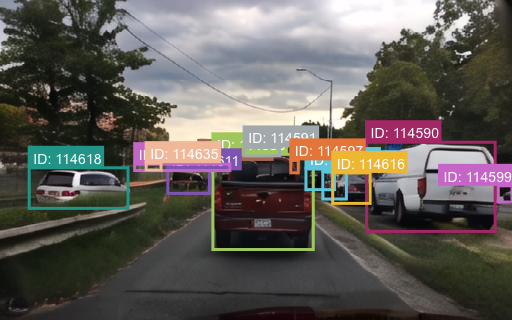} \\ 
       \includegraphics[width=\hsize]{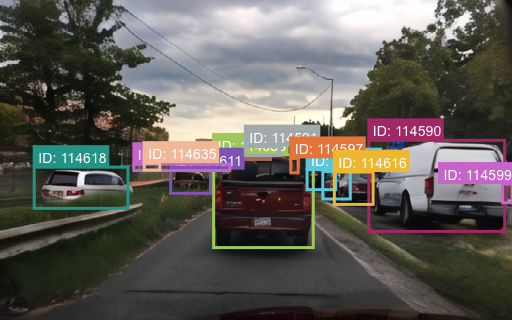}   
      & \includegraphics[width=\hsize]{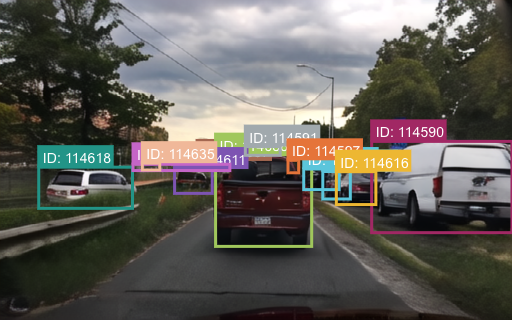} 
      & \includegraphics[width=\hsize]{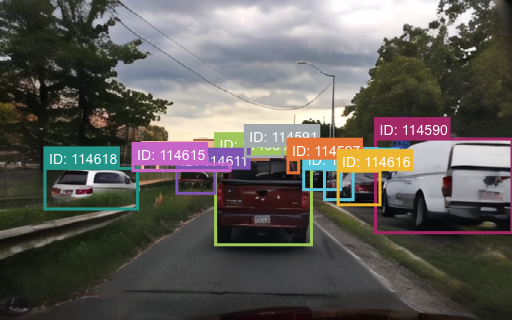}
      & \includegraphics[width=\hsize]{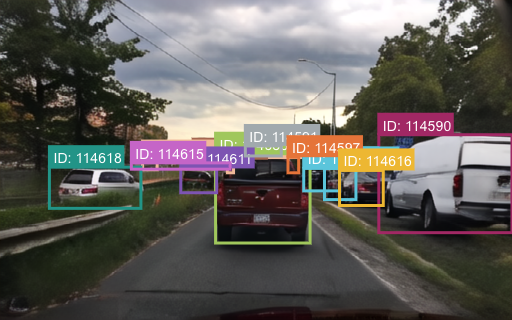}
      & \includegraphics[width=\hsize]{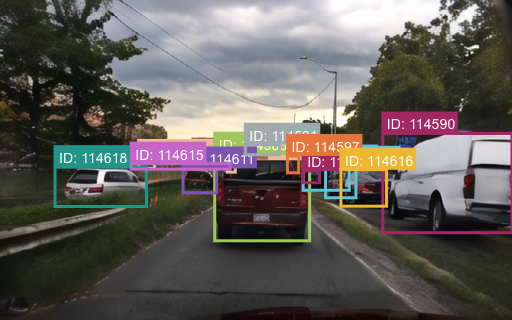} \\ 
      \includegraphics[width=\hsize]{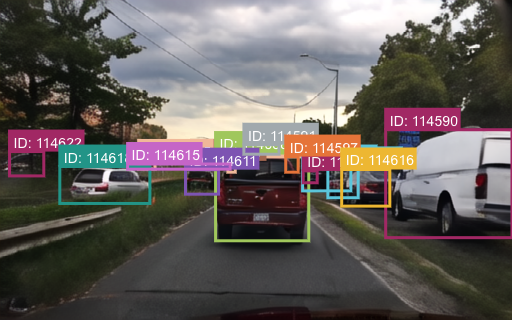}   
      & \includegraphics[width=\hsize]{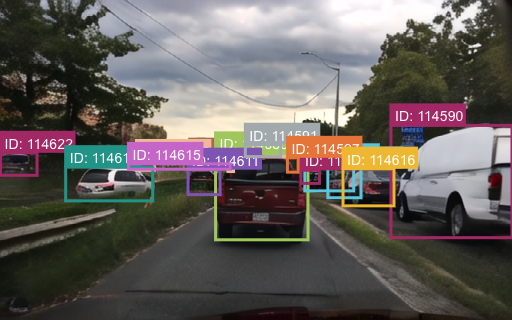} 
      & \includegraphics[width=\hsize]{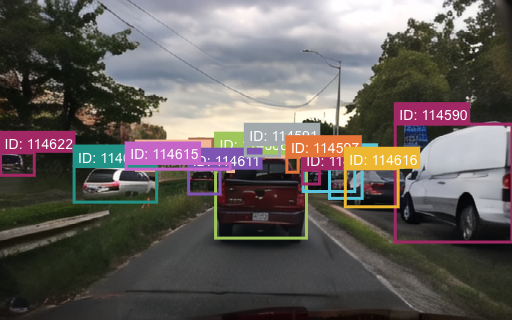}
      & \includegraphics[width=\hsize]{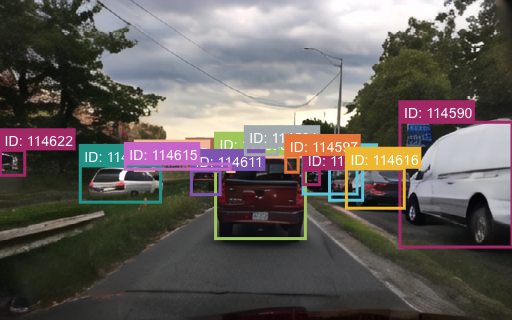}
      & \includegraphics[width=\hsize]{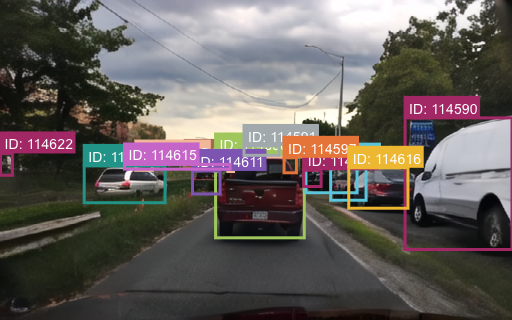} \\ 
      \includegraphics[width=\hsize]{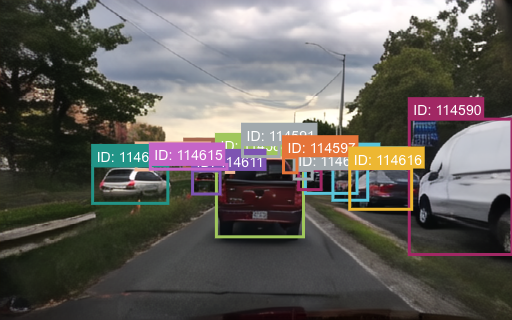}   
      & \includegraphics[width=\hsize]{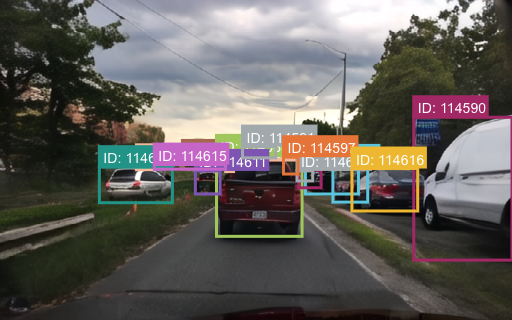} 
      & \includegraphics[width=\hsize]{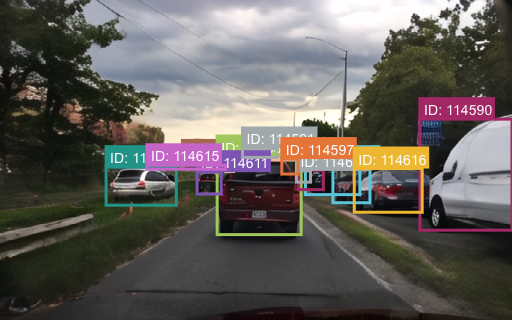}
      & \includegraphics[width=\hsize]{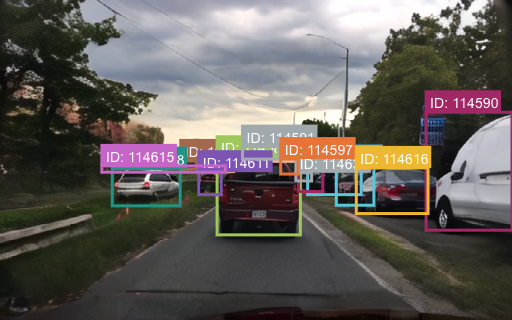}
      & \includegraphics[width=\hsize]{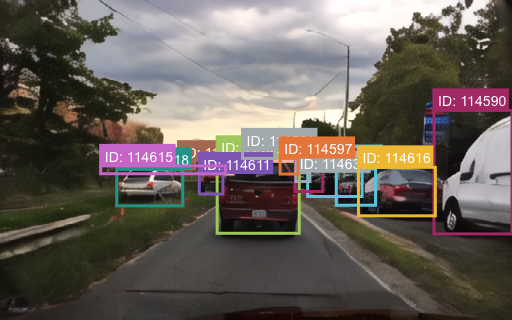} \\ 
\end{tabular}
    \caption{A frame-by-frame visualization of \modelvid~generation, conditioned on the ground-truth bbox frame sequence from the BDD dataset. The ground-truth bounding boxes are outlined in the plots.}
    \label{fig:bdd_controlnet_frame_by_frame}
\end{figure}

%% file: sections/appendix/additional_bdd_results.tex
\subsection{BDD100K Results under Varying Conditioning Scenarios\label{sec:additional_bdd}}
We investigate the impact of the number of conditioning bounding box frames on the quality and alignment of the output to the ground truth. We train distinct models for each conditioning configuration. Our findings are summarized in Table~\ref{tab:quality_vs_conditioning} and Table~\ref{tab:alignment_vs_conditioning}.

\begin{figure}[h]
    \setlength\tabcolsep{3pt} 
    \centering
    \begin{tabular}{@{} M{0.18\linewidth} M{0.18\linewidth} M{0.18\linewidth} M{0.18\linewidth} M{0.18\linewidth} @{}}
    Frame 1 & Frame 7 & Frame 13 & Frame 19  & Frame 25\\
     \includegraphics[width=\hsize]{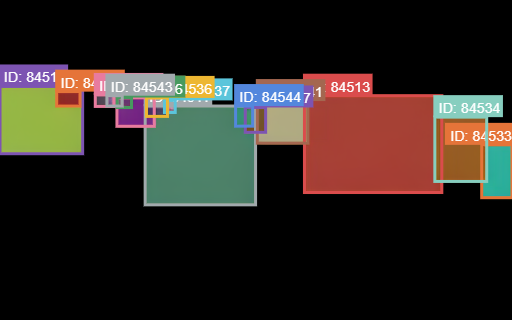}   
      & \includegraphics[width=\hsize]{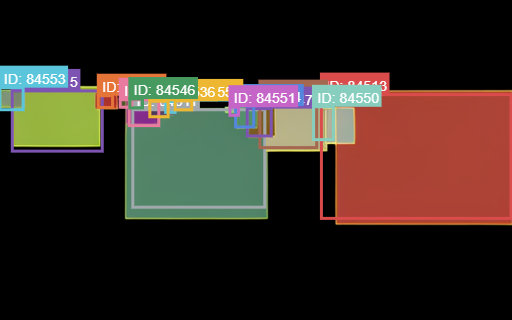} 
      & \includegraphics[width=\hsize]{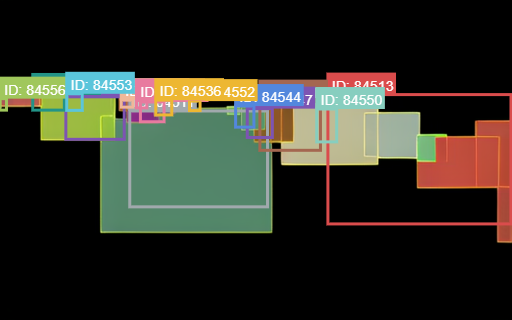}
      & \includegraphics[width=\hsize]{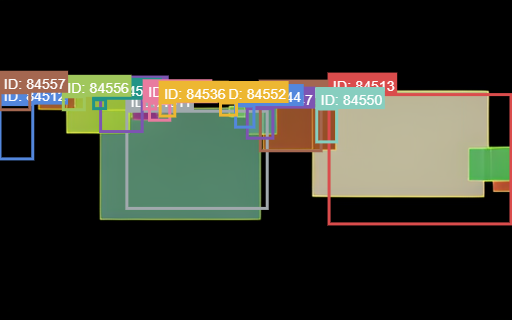}
      & \includegraphics[width=\hsize]{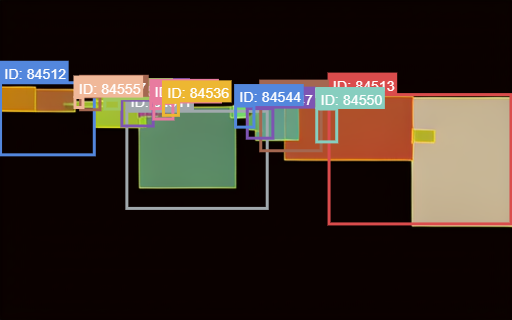} \\ 
      \includegraphics[width=\hsize]{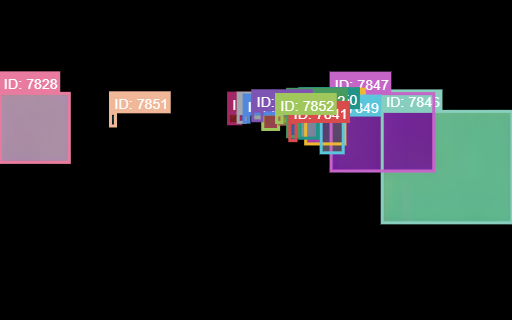}   
      & \includegraphics[width=\hsize]{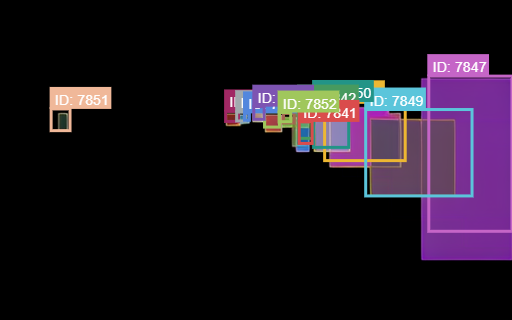} 
      & \includegraphics[width=\hsize]{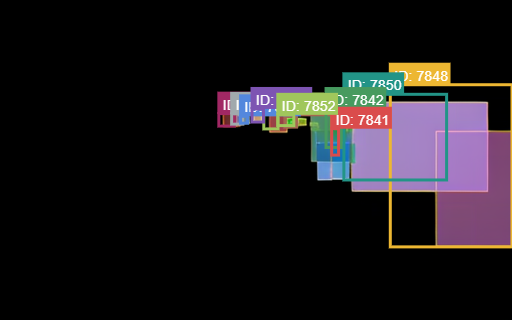}
      & \includegraphics[width=\hsize]{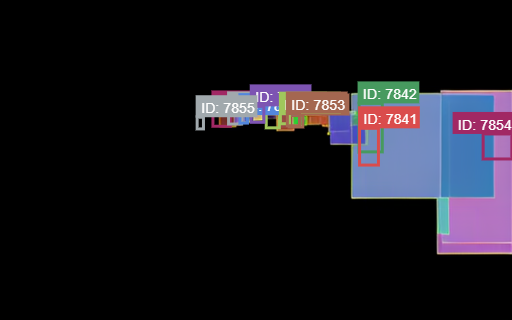}
      & \includegraphics[width=\hsize]{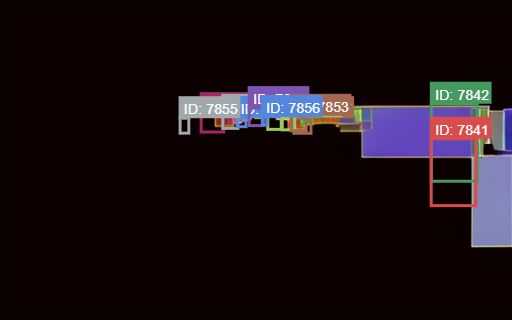} \\ 
    \end{tabular}
    \caption{Bounding box trajectory generations are produced without conditioning on the last frame's bounding box locations. The provided demos display generated trajectories with ground-truth bounding box trajectories overlaid for visual comparison.\label{fig:bbox_generaltion_without_final}}
\end{figure}

\begin{figure}[ht]
    \setlength\tabcolsep{3pt} 
    \centering
    \begin{tabular}{@{} r M{0.18\linewidth} M{0.18\linewidth} M{0.18\linewidth} M{0.18\linewidth} M{0.18\linewidth} @{}}
    & Frame 1 & Frame 7 & Frame 13 & Frame 19  & Frame 25\\
      GT & \includegraphics[width=\hsize]{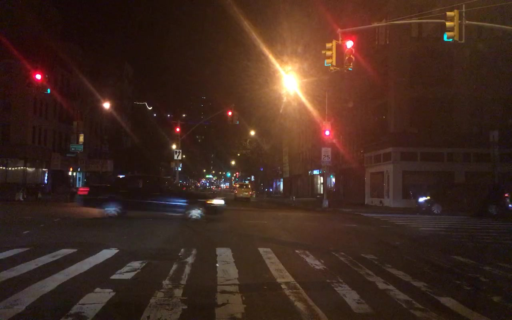}
      & \includegraphics[width=\hsize]{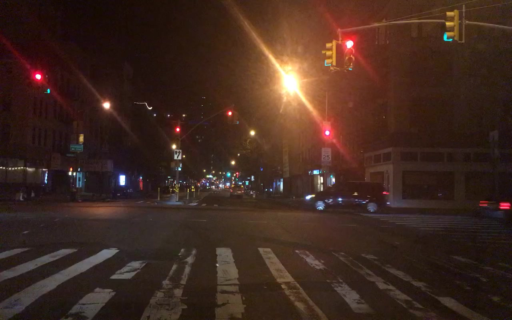}  
      & \includegraphics[width=\hsize]{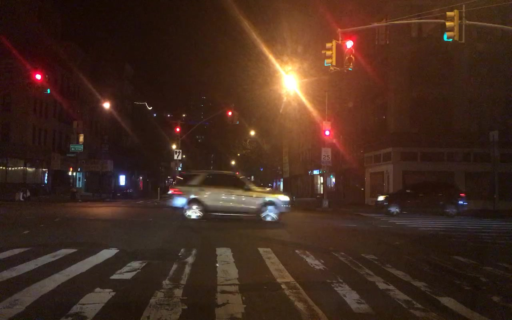}
      & \includegraphics[width=\hsize]{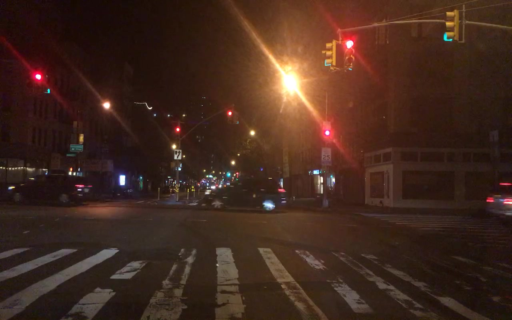} 
      & \includegraphics[width=\hsize]{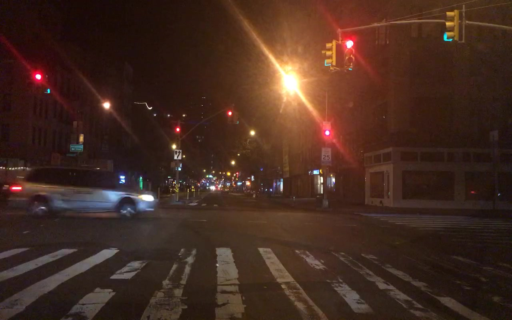}\\
      GF & \includegraphics[width=\hsize]{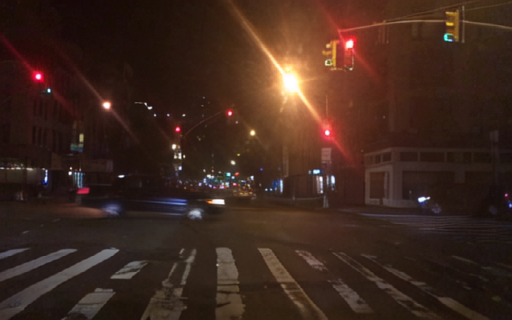}
      & \includegraphics[width=\hsize]{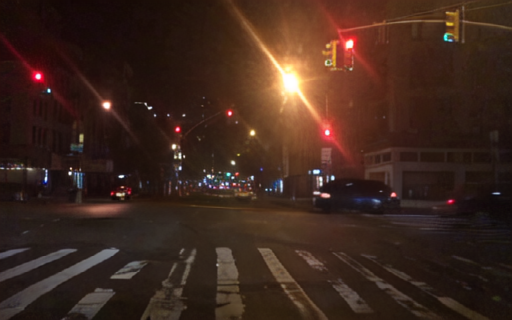}  
      & \includegraphics[width=\hsize]{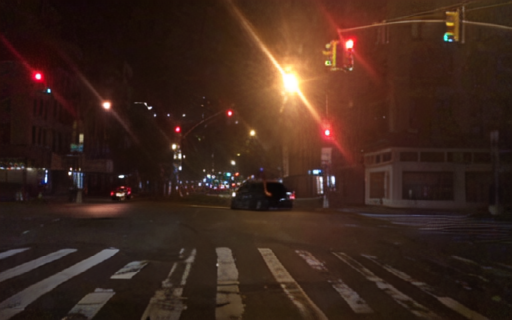}
      & \includegraphics[width=\hsize]{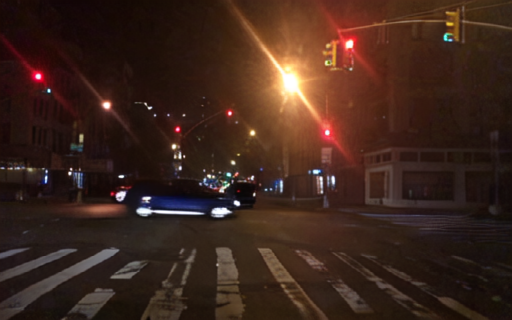} 
      & \includegraphics[width=\hsize]{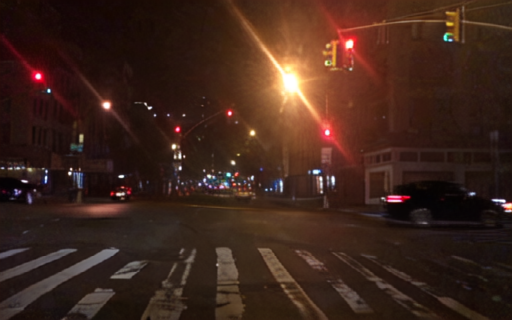}\\
      GB & \includegraphics[width=\hsize]{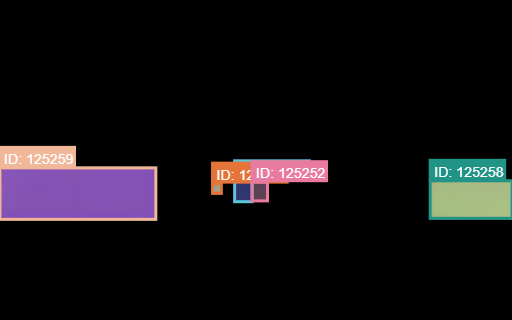}
      & \includegraphics[width=\hsize]{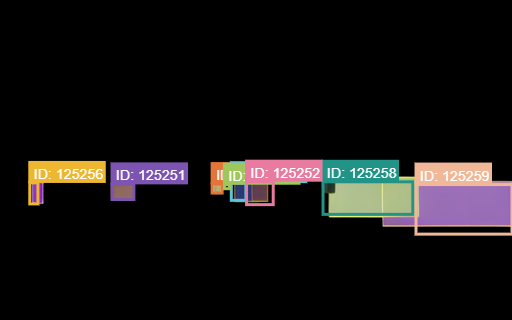}  
      & \includegraphics[width=\hsize]{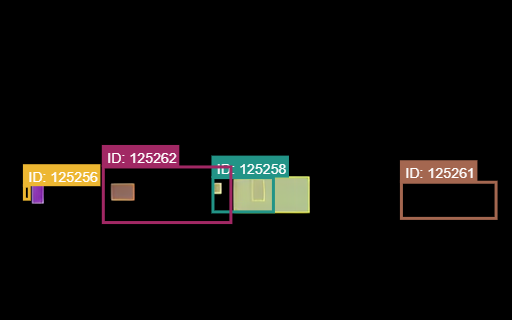}
      & \includegraphics[width=\hsize]{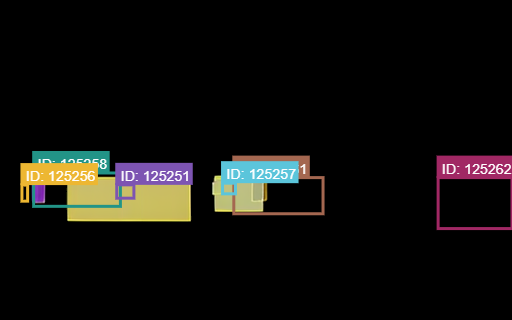} 
      & \includegraphics[width=\hsize]{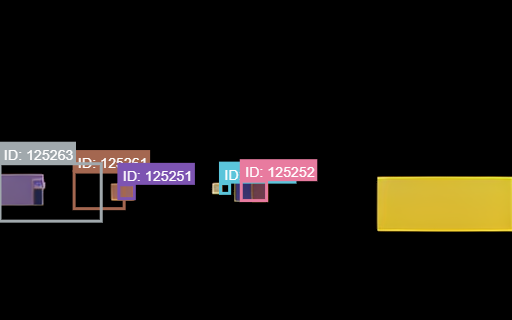}\\
    \hdashline
    GT & \includegraphics[width=\hsize]{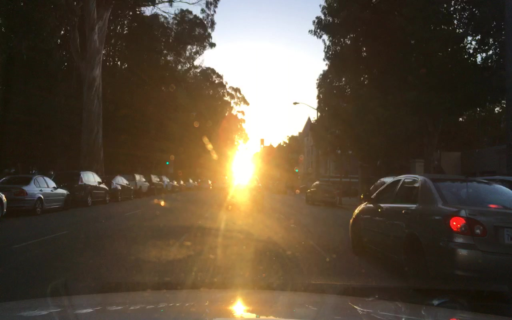}
      & \includegraphics[width=\hsize]{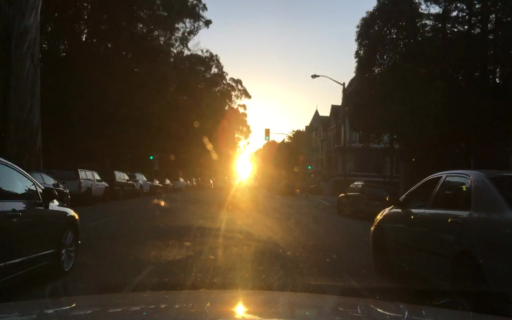}  
      & \includegraphics[width=\hsize]{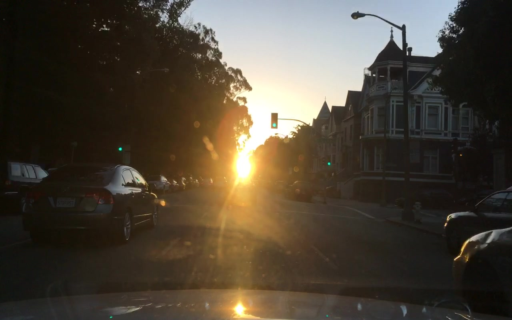}
      & \includegraphics[width=\hsize]{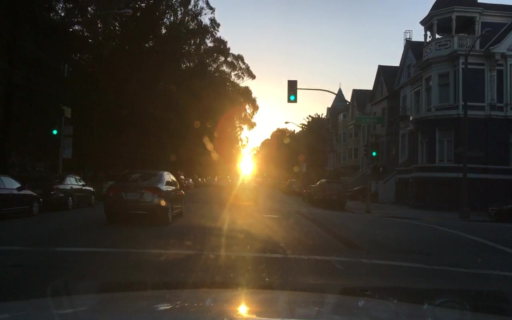} 
      & \includegraphics[width=\hsize]{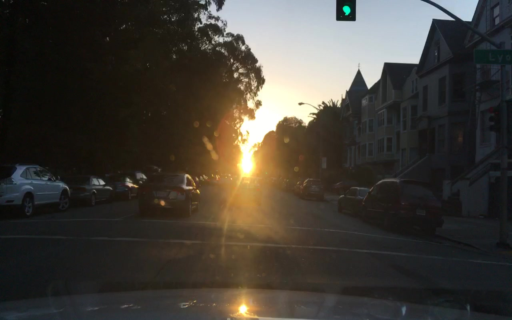}\\
      GF & \includegraphics[width=\hsize]{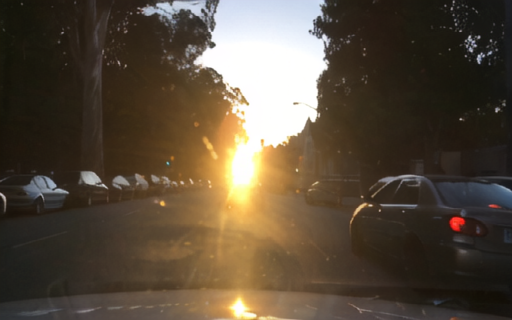}
      & \includegraphics[width=\hsize]{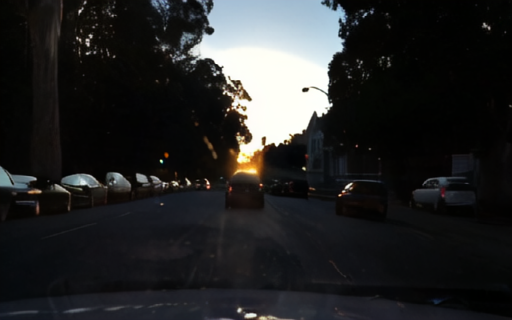}  
      & \includegraphics[width=\hsize]{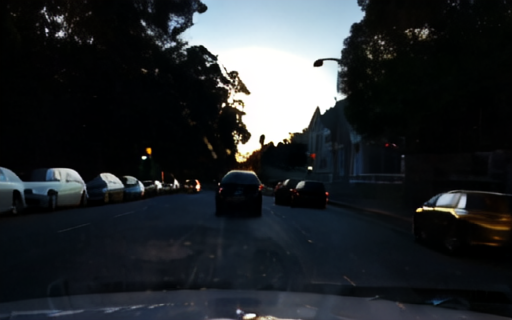}
      & \includegraphics[width=\hsize]{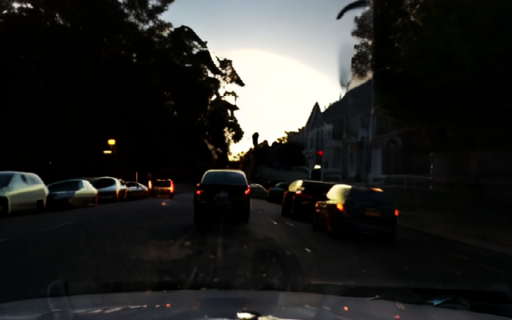} 
      & \includegraphics[width=\hsize]{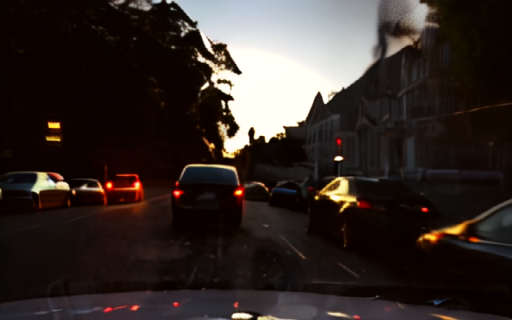}\\
      GB & \includegraphics[width=\hsize]{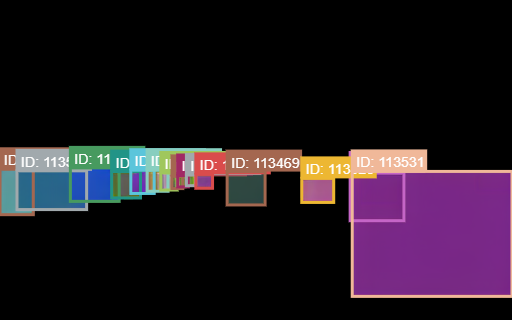}
      & \includegraphics[width=\hsize]{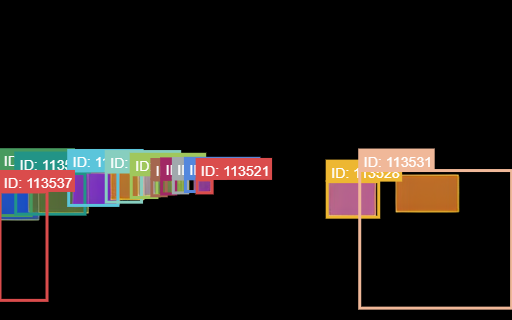}  
      & \includegraphics[width=\hsize]{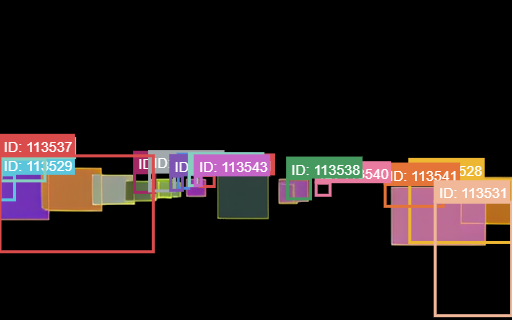}
      & \includegraphics[width=\hsize]{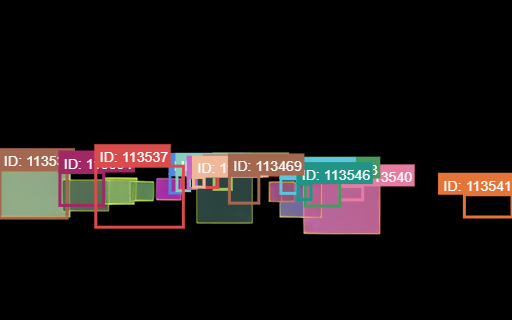} 
      & \includegraphics[width=\hsize]{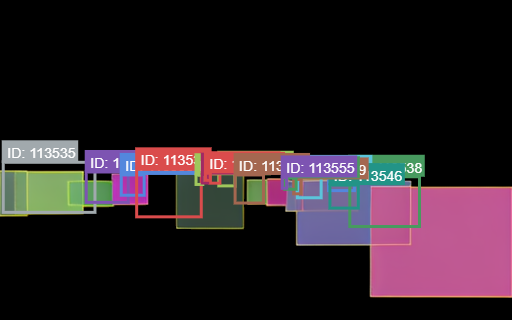}\\
    \end{tabular}
    \caption{2D bounding box frame generations and motion-controlled video generations on BDD100K test split where no final bounding box frame conditioning was provided. \label{fig:bdd_no_final_generations}}
\end{figure}

Key insights include:
\begin{itemize}
    \item 
    Generation quality peaks when conditioned on the entire ground-truth bounding box trajectory.
    \item
    Quality metrics show slight disagreement on the worst-performing setting, with conditioning on either: first+last bounding box frames or first three frames.
    \item 
    Comparing to the no-final-frame conditioning baseline, our results show that minimal conditioning with the central point of the final bounding box frame (termed "trajectory frame"; see Figure~\ref{fig:example_traj_frame}) significantly enhances generation quality.
    \item 
    Table~\ref{tab:alignment_vs_conditioning} demonstrates that incorporating the last bounding box frame significantly improves alignment between generated and ground-truth bounding box trajectories, as evidenced by the improved performance when conditioning on the first and last frames versus the first three frames.
    \item 
    Figure~\ref{fig:bbox_generaltion_without_final} showcases two predicted bounding box trajectories generated by \modelbbox~without conditioning on last-frame bounding box locations. Notably, despite inaccurate last-frame bounding boxes compared to ground-truth, the \modelbbox~still predict convincing bounding box trajectories independently of last-frame bounding box locations.
    \item 
    Figure~\ref{fig:bdd_no_final_generations} presents exemplary clips generated by the \modelvid~model, conditioned on bounding box sequences derived without last-frame bounding box information.
    \item
    Figure~\ref{fig:bbox_generaltion_without_final} and Figure~\ref{fig:bdd_no_final_generations}'s visualization displays every 6th frame from a 25-frame clip, with the actual video generated at a frame rate of 7fps. In the leftmost column labels of Figure~\ref{fig:bdd_no_final_generations}, GT represents ground-truth, GB represents generated bounding box frames, and GF represents generated frames.
    \item 
    Comparing the ground-truth and generated clips in Figure~\ref{fig:bdd_no_final_generations} reveals notable discrepancies. In the first example, a novel vehicle emerges (darker yellow bounding box), not present initially. This demonstrates \modelbbox's ability to create new bounding box instances, which \modelvid~then uses to generate corresponding visual content
\end{itemize}

\begin{table}[h]
    \centering
    \begin{tabular}{lccccc}
    \toprule
      \textbf{Pipeline}& \textbf{\# Cond. BBox} & \textbf{FVD\(\downarrow\)} & \textbf{LPIPS\(\downarrow\)} & \textbf{SSIM\(\uparrow\)} & \textbf{PSNR\(\uparrow\)}\\
    \midrule
    \modelbbox~+ \modelvid & 1-to-1 & 412.8 & 0.2967 &  0.5470 & 17.52\\
    \modelbbox~+ \modelvid  & 3-to-1 & 373.1 &  0.3071 & 0.5407 & 17.37\\
    \modelbbox~+ \modelvid & 3-to-0 & 389.7 & 0.3085 &  0.5401 & 17.11 \\
    \modelbbox\(^{\text{Traj}}\) + \modelvid & 3-to-1\(^{\text{Traj}}\) & 375.8 & 0.2973 & 0.5481 & 17.55\\
    Teacher-forced \modelvid & All & {\bfseries 348.9} & {\bfseries 0.2926} & {\bfseries 0.5836} & {\bfseries 18.39}\\
    \bottomrule
    \end{tabular}
    \caption{Summary of generation quality metrics on BDD100K dataset: evaluating the impact of bounding box conditioning types on generation quality.\label{tab:quality_vs_conditioning}}
\end{table}

\begin{table}[h]
\centering
\resizebox{\textwidth}{!}{\begin{tabular}{cccccccc}
\toprule
 \textbf{Method} & \specialcell[]{\textbf{\# Cond.}\\\textbf{BBox}} & \textbf{maskIoU\(\uparrow\)} & \textbf{maskP\(\uparrow\)} & \textbf{maskR\(\uparrow\)} & \specialcell[]{\textbf{maskIoU\(\uparrow\)}\\(first+last)} & \specialcell[]{\textbf{maskP\(\uparrow\)}\\(first+last)} &  \specialcell[]{\textbf{maskR\(\uparrow\)}\\(first+last)} \\
 \midrule
\multirow{4}{*}{ BBox Generator} & 1-to-1 & \(.587\pm .214\)& \(.747\pm.187\)& \(.712\pm.194\) & \(.954\pm.047\) & \(\textbf{.955}\pm.047\)&\(\textbf{.999}\pm.002\)\\
& 3-to-1 & \(\textbf{.647}\pm.176\)   & \(\textbf{.784}\pm.150\) &  \(\textbf{.783}\pm.156\) &  \(\textbf{.955}\pm.043\) & \(\textbf{.955}\pm.042\) &  \(.997\pm.001\) \\
 & 3-to-0 & \(.522\pm.204\)   & \(.686\pm.204\) &  \(.673\pm.201\) &  \(.578\pm.214\) & \(.735\pm.222\) &  \(.734\pm.201\) \\
& 3-to-1\(^{\text{Traj}}\) & \(.606\pm.201\) & \(.773\pm .162\) & \(.722\pm.189\) & \(.798\pm.155\) & \(.886\pm.137\)& \(.894\pm.118\) \\
\bottomrule
\end{tabular}}
\caption{Comparing trajectory alignment: comparing generated and ground-truth trajectory alignment under varying conditioning.~\label{tab:alignment_vs_conditioning}}
\end{table}


%% file: sections/appendix/trajeglish_baseline.tex
To evaluate both the performance and simplicity of our bounding box generator model, we implemented a baseline method inspired by the Trajeglish model \citep{philion2023trajeglish}, which we refer to as the ``Trajeglish-Style" model.

The original Trajeglish model employs a discrete sequence modeling approach to represent multi-agent road-user trajectories. Vehicle trajectories are modeled as a discrete sequence of actions, where each action corresponds to a token that describes the agent's relative displacement. The model utilizes a GPT-like encoder-decoder architecture, taking as input the agents' initial timestep information and map objects, and outputting a distribution over possible displacement actions. Trajeglish operates in bird's-eye view (BEV) and achieved state-of-the-art (SOTA) performance on the 2023 WOMD Sim Agents Benchmark.

To adapt this method to our problem, we made several modifications, which are detailed below. Our ``Trajeglish-Style" model focuses on predicting 2D bounding boxes from the ego perspective (POV). Unlike the BEV representation, these bounding boxes can overlap in 3D space and are subject to resizing during rollout. Additionally, the boxes can dynamically enter and exit the frame at any point in the sequence.

For our baseline, we define the set of possible actions as the displacement of a bounding box corner. Each corner's displacement is represented by a 2D vector, where the vector's magnitude is discretized uniformly into 16 values, ranging from 0.0 to 0.1 of the image size in the respective dimension. The direction of the displacement is discretized into 24 values, covering the full 360-degree range. This joint discretization yields a vocabulary size of 384 (16 magnitudes * 24 directions). At each timestep, the movement of a bounding box is determined by selecting two actions: one for the displacement of the top-left corner and one for the bottom-right corner.

As input to the model's encoder, we provide the initial and final bounding box coordinates (1 or 3 frames), as well as the type of each agent (e.g., car, pedestrian, cyclist). In lieu of HDMap information (missing in most of the datasets used in this study) we use the initial natural image as contextual input to the encoder. The image is encoded using the same method as our diffusion-based bounding box generator (VAE image encoding combined with CLIP embeddings) to ensure fair comparison.

While the original Trajeglish model achieved SOTA performance in its intended task of predicting BEV agent trajectories, we do not claim that our ``Trajeglish-Style" model reaches SOTA for the current task of ego POV 2D bounding box trajectory prediction. In fact, we consider this task to be significantly more challenging and less suited to the modeling approach of Trajeglish, which was not originally designed for the complexities predicting the motion of bounding boxes in the ego perspective, as outlined above. This increased difficulty accounts for the drop in performance, as seen in Table~\ref{tab:predictor_scores}. Nevertheless, we believe that the ``Trajeglish-Style" model serves as a valuable baseline for evaluating the performance of other models, including our own diffusion-based approach.

Despite the complexities inherent in this task, our diffusion-based BBox Generator, which operates directly on pixel images of bounding boxes, offers a compelling alternative due to its simplicity and strong performance.

The code for the ``Trajeglish-Style" model implementation is available in the GitHub repository associated with this work.

%% file: sections/appendix/future_work.tex
\subsection{Failure Cases}
\begin{figure}[h]%
    \centering
    \subfloat[Ground-Truth Image]{{\includegraphics[width=0.45\linewidth]{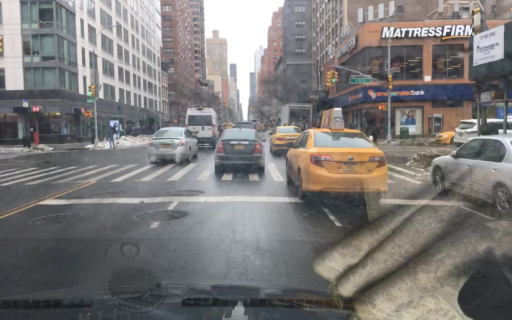} }}%
    \qquad
    \subfloat[VAE Reconstruction]{{\includegraphics[width=0.45\linewidth]{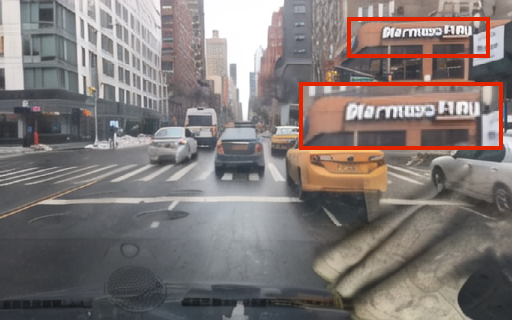} }}%
    \caption{
    Failure case: VAE's inability to reconstruct building signage.}%
    \label{fig:sign_failure}%
\end{figure}

Ctrl-V struggles to accurately encode and decode specific visual details, particularly road signs, building lettering, license-plate information and lane markings. This limitation stems from the constraints of the off-the-shelf VAE network used for image encoding and decoding. Figure~\ref{fig:sign_failure} illustrates the VAE model's difficulty in accurately encoding and decoding the "MATTRESS FIRM" sign on the building. A potential direction for future research is exploring super-resolution techniques to recover fine-grained text details.

\begin{figure}[h]%
    \centering
    \subfloat[Example of final frame generation]{{\includegraphics[width=0.45\linewidth]{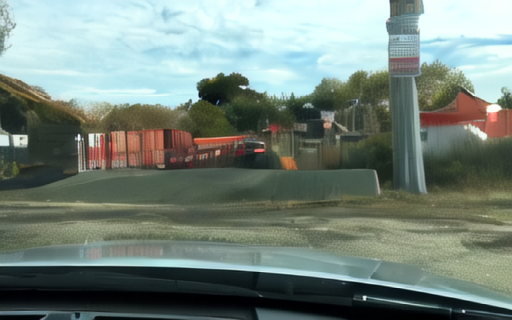} }}%
    \qquad
    \subfloat[Example of final frame generation]{{\includegraphics[width=0.45\linewidth]{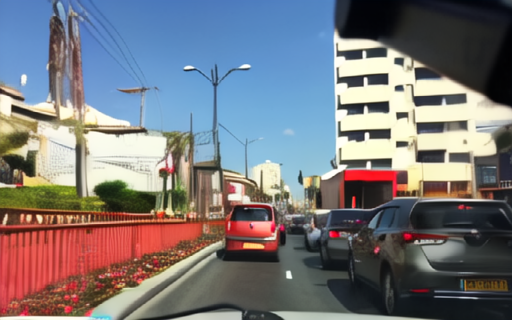} }}%
    \caption{Failure case: resolution degradation over time.}%
    \label{fig:final_frame}%
\end{figure}

Resolution degradation is a pervasive issue in video generation, primarily caused by error accumulation through temporal propagation. Notably, Ctrl-V exhibits this problem, particularly when the ego vehicle travels at higher speeds or changing directions. Figure~\ref{fig:final_frame} shows instances of degraded generation quality, where the last frame outputs suffer from a loss of realism. Future work can investigate multistage training paradigms, leveraging auxiliary networks to refine and enhance the visual quality of generated frames.

\subsection{Noteworthy Discoveries in our \modelvid~Model}
\begin{figure}[h]
    \setlength\tabcolsep{3pt} 
    \centering
    \begin{tabular}{@{} M{0.18\linewidth} M{0.18\linewidth} M{0.18\linewidth} M{0.18\linewidth} M{0.18\linewidth} @{}}
     \includegraphics[width=\hsize]{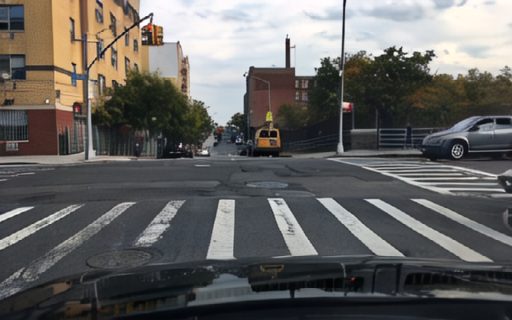}   
      & \includegraphics[width=\hsize]{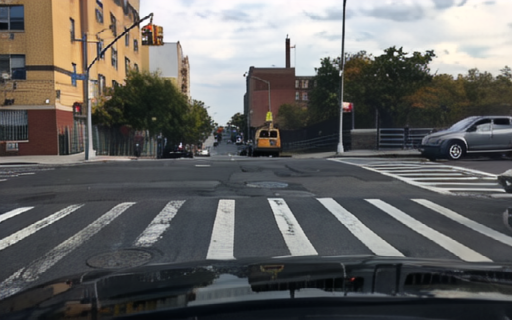} 
      & \includegraphics[width=\hsize]{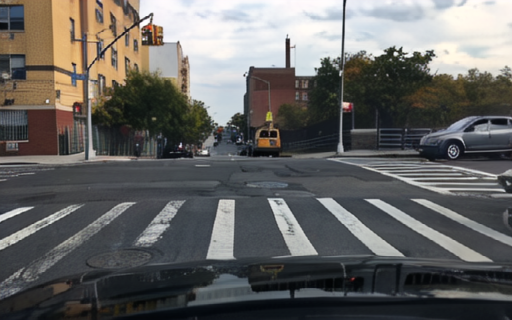}
      & \includegraphics[width=\hsize]{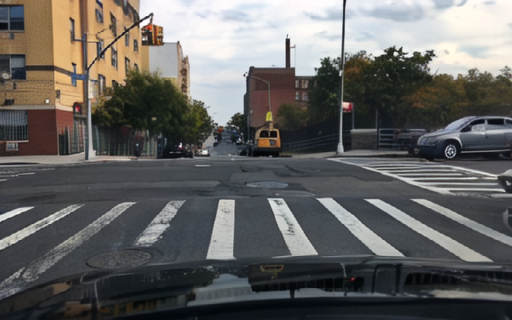}
      & \includegraphics[width=\hsize]{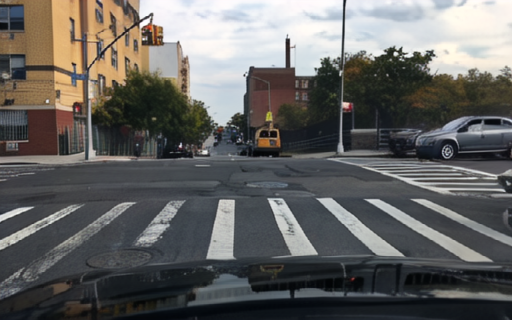} \\ 
      \includegraphics[width=\hsize]{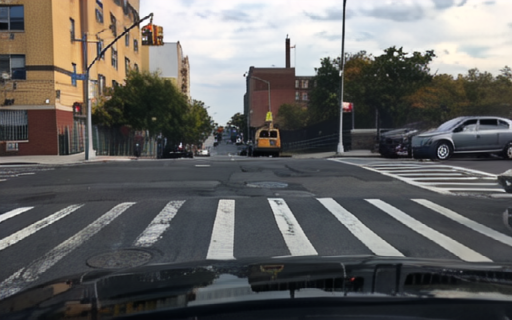}   
      & \includegraphics[width=\hsize]{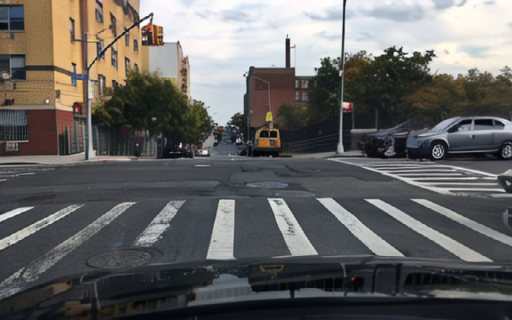} 
      & \includegraphics[width=\hsize]{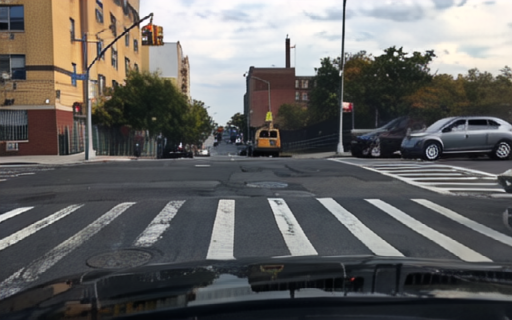}
      & \includegraphics[width=\hsize]{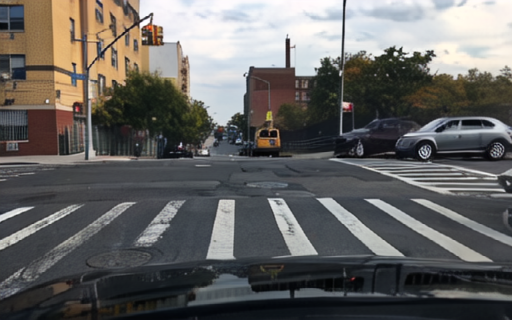}
      & \includegraphics[width=\hsize]{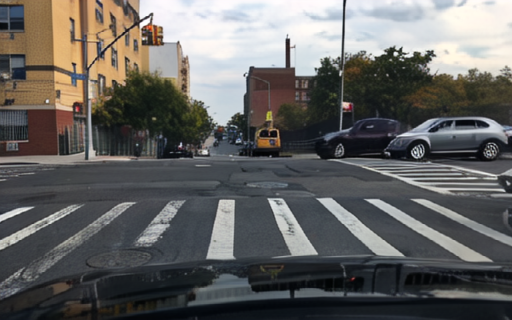} \\
       \includegraphics[width=\hsize]{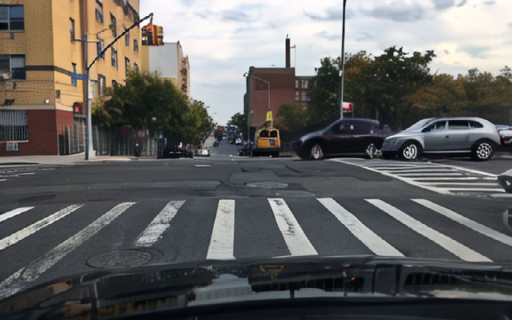}   
      & \includegraphics[width=\hsize]{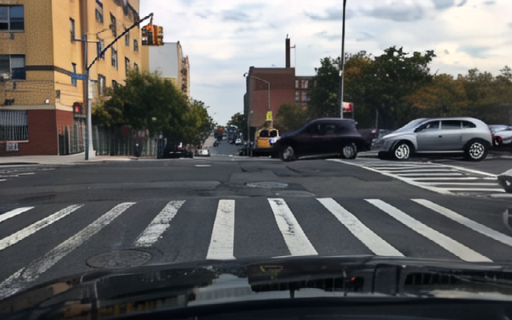} 
      & \includegraphics[width=\hsize]{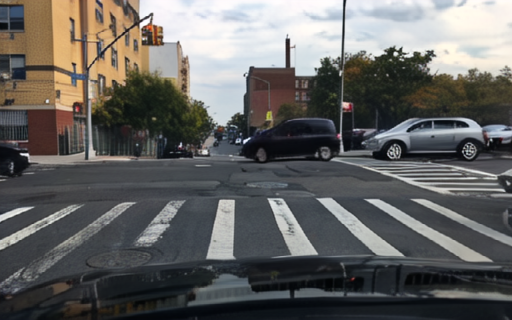}
      & \includegraphics[width=\hsize]{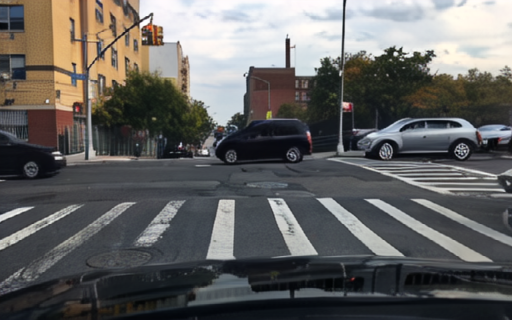}
      & \includegraphics[width=\hsize]{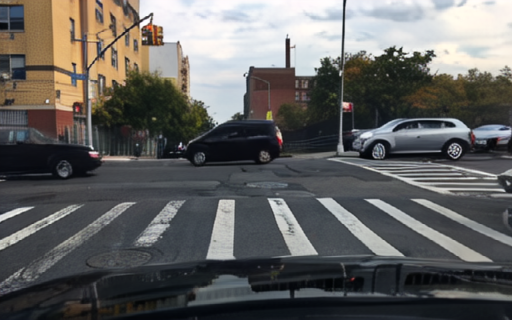} \\
      \includegraphics[width=\hsize]{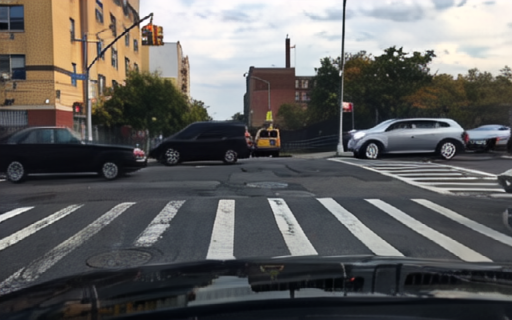}   
      & \includegraphics[width=\hsize]{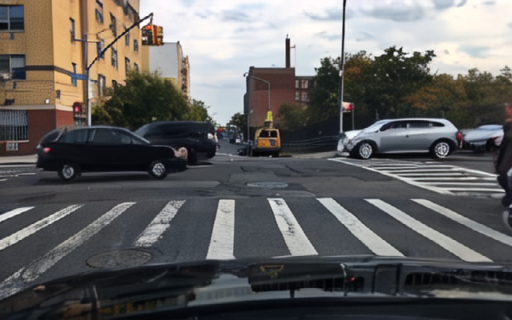} 
      & \includegraphics[width=\hsize]{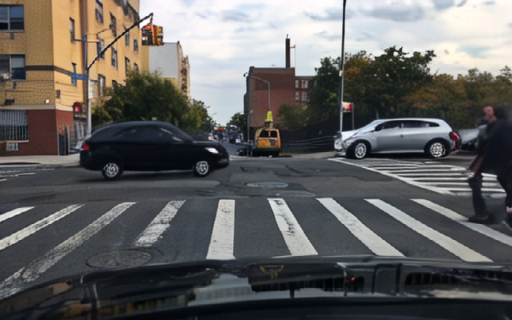}
      & \includegraphics[width=\hsize]{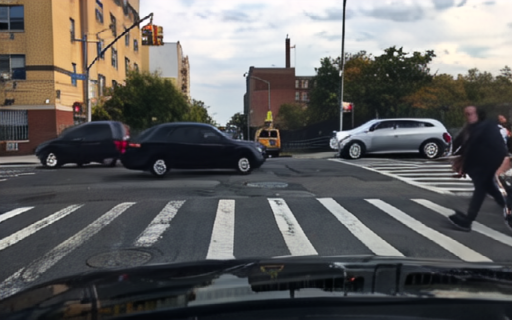}
      & \includegraphics[width=\hsize]{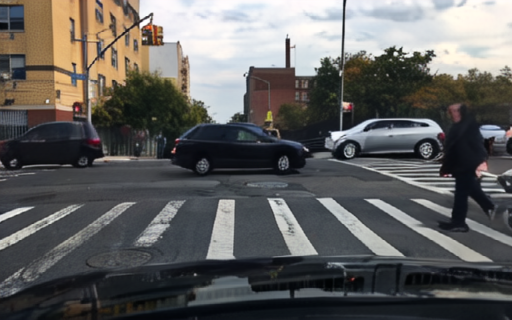} \\
      \includegraphics[width=\hsize]{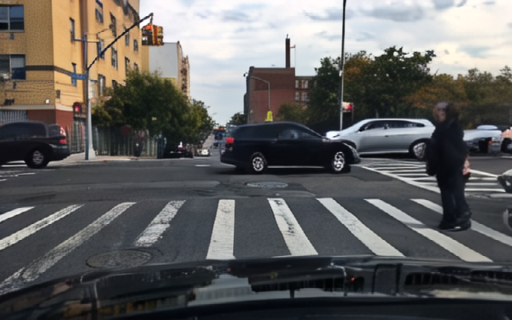}   
      & \includegraphics[width=\hsize]{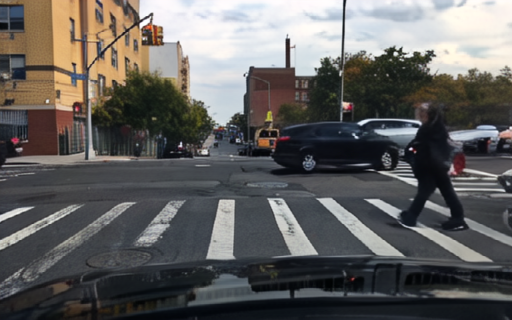} 
      & \includegraphics[width=\hsize]{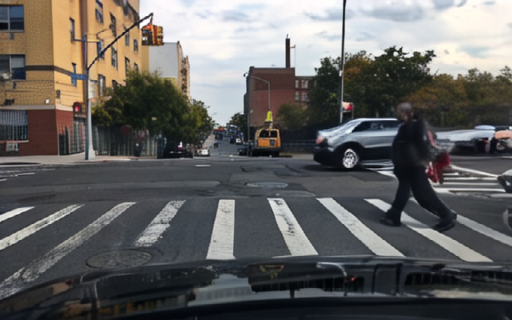}
      & \includegraphics[width=\hsize]{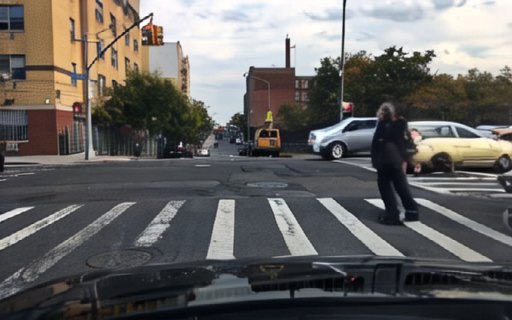}
      & \includegraphics[width=\hsize]{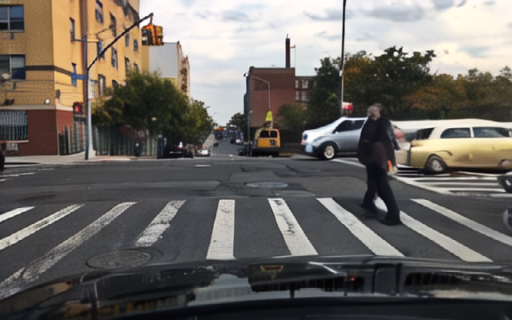}
\end{tabular}
    \caption{A frame-by-frame visualization of \modelvid~generation, conditioned on the ground-truth bbox frame sequence from the BDD dataset, demonstrating the model's capability of handling incoming objects.}
    \label{fig:proof_of_id}
\end{figure}

Our experiments have yielded findings suggesting that our \modelvid~model not only understands bounding box locations in scenes but also the encoded track IDs and 3D bounding box orientation information in the rendered bounding box frames. While these findings are not conclusive and require further investigation to be confirmed, they highlight potential directions for future projects. Here we share some of these findings to guide further research and development.
\begin{enumerate}
    \item 
During preprocessing, we assign a random color to each track ID and fill the bounding boxes with their corresponding track IDs' colors. When our \modelvid~ receives unreasonable generations, it still attempts to produce outputs that adhere to the given conditions, including the false bounding box IDs.
    \item 
    In our 3D bounding box plots, we encode the vehicle's orientation by marking the rear end with an ``X". Occasionally, our \modelbbox~incorrectly identifies the car's orientation, marking the front side with an ``X". Therefore, during the video generation phase, we sometimes observe that the \modelvid~model recognizes the wrong markings in the bounding box frames and generates a video of a car driving backwards.
    \item 
    We also observe that \modelvid~may recognize the encoded object information regarding its class in the bounding box frames. Specifically, when an object is not present in the initial frame but its bounding box label appears in the final frame, the model can decode its class information and generate a correct instance of that class in the clip. For example, observing Figure~\ref{fig:proof_of_id}, we see the model successfully handle new objects entering the scene, accurately assigning the incoming car and pedestrian to their corresponding bounding boxes by leveraging encoded object IDs. However, we suspect that the object's class information could also be inferred from the bounding box shapes. Further investigation is required.
    \item 
    Given the previous findings, we believe that plotting the bounding box outlines slightly thicker to make them more prominent could potentially improve our results.
\end{enumerate}

\subsection{Future Directions}
For future work, we believe it is worthwhile to develop a systematic approach to investigate the aforementioned observations. Additionally, it would be interesting to explore alternative methods for generating bounding boxes. Furthermore, it would be beneficial to create a systematic method to measure the likelihood of the bounding box frames, instead of directly computing their IoU, recall, and precision rate with ground truth.